\documentclass[oneside,a4paper,onecolumn,11pt]{article}

\usepackage{fullpage}
\usepackage[left=2cm,top=2cm,bottom=2cm,right=2cm,includehead,nomarginpar,headheight=16pt]{geometry}
\usepackage[ruled, linesnumbered, vlined, commentsnumbered]{algorithm2e}
\usepackage{graphicx,subfig} 
\usepackage{amsfonts,amssymb,amsmath,amsthm,amsopn,mathtools}	
\usepackage{booktabs,diagbox,colortbl,multirow,tabularx,threeparttable,hhline}
\usepackage[listings,skins,breakable]{tcolorbox}

\usepackage{fancyhdr,fancyheadings,nopageno,lastpage} 
\usepackage{url}
\usepackage{enumerate}
\usepackage[shortlabels]{enumitem}
\usepackage{csquotes}
\usepackage{authblk}
\usepackage{footnote}
\usepackage{hyperref}
\usepackage{prettyref}
\usepackage{cite}
\usepackage[numbers]{natbib}
\usepackage{setspace}
\usepackage{color}
\usepackage{xcolor}  
\usepackage{geometry}
\usepackage{academicons}
\usepackage{fontawesome5}
\usepackage{pifont,ifsym,marvosym,manfnt} 

\UseRawInputEncoding

\usepackage{tikz}
\usepackage{pgfplots}
\usetikzlibrary{positioning,shapes,shadows,arrows,calc}
\tikzstyle{component}=[rectangle, draw=black, rounded corners, fill=blue!40, drop shadow, text centered, anchor=north, text=white, minimum height=1cm]
\tikzstyle{arrow}=[->, thick]

\pgfplotsset{compat=1.12}
\usetikzlibrary{intersections}
\usetikzlibrary{pgfplots.statistics}
\usepgfplotslibrary{fillbetween}

\fancypagestyle{plain}{%
    \fancyfoot[C]{}
    \fancyfoot[R]{\faLeanpub\ \, \thepage\ / \pageref{LastPage}}
    \fancyfoot[L]{\faBraille\ \textsc{COLALab Report} \faSlackHash\ \footnotesize\textifsym{2024005}}
}

\hypersetup{
    colorlinks=true,
    linkcolor=ultramarine,
    filecolor=magenta,
    urlcolor=ultramarine,
    pdftitle={Overleaf Example},
    pdfpagemode=FullScreen,
}

\definecolor{red(munsell)}{rgb}{0.95, 0.0, 0.24}
\definecolor{navyblue}{RGB}{0, 0, 128}
\definecolor{myblue}{RGB}{34,31,217}
\definecolor{mycyan}{gray}{.7}
\definecolor{Gray}{gray}{0.9}
\definecolor{usccardinal}{rgb}{0.6, 0.0, 0.0}
\definecolor{ultramarine}{RGB}{0,32,96}
\definecolor{amber}{rgb}{1.0, 0.49, 0.0}

\newtcolorbox{quotebox}{colback=gray!10,boxrule=0.4pt,colframe=black,fonttitle=\bfseries,top=1pt,bottom=1pt}

\usepackage[ruled, linesnumbered, vlined, commentsnumbered]{algorithm2e}
\usepackage{prettyref}
\usepackage{amsfonts,amssymb,amsmath,amsthm,amsopn,mathrsfs,mathtools,wasysym}	
\usepackage{bm}
\usepackage{multirow}

\SetCommentSty{mycommfont}

\newcommand{\pref}{\prettyref}

\newrefformat{fig}{Fig.~\ref{#1}}
\newrefformat{tab}{Table~\ref{#1}}
\newrefformat{sec}{Section~\ref{#1}}
\newrefformat{alg}{Algorithm~\ref{#1}}
\newrefformat{property}{Property~\ref{#1}}
\newrefformat{theorem}{Theorem~\ref{#1}}
\newrefformat{definition}{Definition~\ref{#1}}
\newrefformat{corollary}{Corollary~\ref{#1}}
\newrefformat{lemma}{Lemma~\ref{#1}}
\newrefformat{conj}{Conjecture~\ref{#1}}
\newrefformat{def}{Definition~\ref{#1}}
\newrefformat{eq}{equation~(\ref{#1})}
\newrefformat{app}{Appendix~\ref{#1}}

\usepackage{lscape}

\newenvironment{code-example}
{
\vspace{0.15cm}
\noindent\begin{minipage}{\linewidth}
\begin{center}
\arrayrulecolor{black}
\color{black}
\begin{tabular}{|p{0.95\linewidth}|}
\hline%
\rowcolor{pink!20}%
}
{
\\\hline
\end{tabular}
\end{center}
\end{minipage}
\vspace{-0.2cm}
}

\begin{document}

\title{\vspace{-1ex}\LARGE\textbf{Multi-Fidelity Methods for Optimization: A Survey}~\footnote{This paper is submitted for potential publication. Reviewers can feel free to use this manuscript in peer review.}}

\author[1]{\normalsize Ke Li}
\author[1]{\normalsize Fan Li}
\affil[1]{\normalsize Department of Computer Science, University of Exeter, EX4 4RN, Exeter, UK}
\affil[\Faxmachine\ ]{\normalsize \texttt{\{k.li,f.li\}@exeter.ac.uk}}

\date{}
\maketitle

\vspace{-3ex}
{\normalsize\textbf{Abstract: }}Real-world black-box optimization often involves time-consuming or costly experiments and simulations. Multi-fidelity optimization (MFO) stands out as a cost-effective strategy that balances high-fidelity accuracy with computational efficiency through a hierarchical fidelity approach. This survey presents a systematic exploration of MFO, underpinned by a novel text mining framework based on a pre-trained language model. We delve deep into the foundational principles and methodologies of MFO, focusing on three core components---multi-fidelity surrogate models, fidelity management strategies, and optimization techniques. Additionally, this survey highlights the diverse applications of MFO across several key domains, including machine learning, engineering design optimization, and scientific discovery, showcasing the adaptability and effectiveness of MFO in tackling complex computational challenges. Furthermore, we also envision several emerging challenges and prospects in the MFO landscape, spanning scalability, the composition of lower fidelities, and the integration of human-in-the-loop approaches at the algorithmic level. We also address critical issues related to benchmarking and the advancement of open science within the MFO community. Overall, this survey aims to catalyze further research and foster collaborations in MFO, setting the stage for future innovations and breakthroughs in the field.



\section{Introduction}
\label{sec:introduction}

\begin{quote}
    \lq\lq \textit{Everything should be made as simple as possible, but no simpler.}\rq\rq\ --- \textsl{Albert Einstein}
\end{quote}

Complex system design, spanning aerospace~\cite{ShiLYWWL21}, automotive systems~\cite{WaschlKSR14}, and missile technology~\cite{WuYJZZFG22}, evolve through distinct phases---beginning with early design, advancing through detailed design, and culminating in manufacturing. Each phase is characterized by unique attributes, such as the number and complexity of design parameters and the required depth of analysis for various phenomena. Tailored physical modeling methodologies are essential to accurately represent relevant disciplines within these stages. In the early design phase, the primary goal is to explore a broad range of possibilities. Models chosen for this phase prioritize computational efficiency over detail. While these low-fidelity (LF) models provide rapid assessments of numerous design configurations, thereby identifying viable pathways with minimal resource use, their reduced precision often necessitates further refinement. As the process moves to the detailed design phase, the focus shifts to accuracy and specificity. High-fidelity (HF) models, acclaimed for their detailed representation of complex physical phenomena, become indispensable. These models, while offering enhanced predictive capabilities crucial for identifying and mitigating potential design flaws and optimizing performance parameters, bring significant computational challenges, requiring intensive computing resources and extended processing time. 

The strategic selection of modeling approaches at each phase reflects a balance between precision and efficiency, highlighting the evolving needs of complex system design. This dynamic interplay, known as multi-fidelity methods~\cite{PeherstorferWG18}, not only influences the design outcome but also fosters innovations in computational methodologies, potentially impacting broader fields such as system engineering and scientific discovery. In the realm of optimization, multi-fidelity optimization (MFO) utilizes LF information to strategically reduce reliance on HF analysis during optimization, ensuring convergence to the HF optimal design. As a result, MFO has emerged as an essential tool in complex system optimization, capable of tackling challenges ranging from resource constraints to the need for rapid design iterations.

The concepts of multi-fidelity design analysis and optimization have garnered significant attention, as evidenced by numerous contributions, including survey papers (e.g.,~\cite{GodinoPKH16,GisellePKH19,ParkHH17,BrevaultBH20,PeherstorferWG18}), a comprehensive book~\cite{ZhouZHM22}, and several Ph.D. theses (e.g.,~\cite{Gratiet13,Robinson07,March12,Haas12,Demange18,Santos21}). For instance, Godino et al.~\cite{GodinoPKH16} conducted a thorough analysis of methods for constructing LF models, focusing on four primary correction techniques---multiplicative, additive, comprehensive, and space mapping. Their study also explored the cost and time-efficiency advantages of multi-fidelity models in optimization tasks. Building on this, Giselle et al.~\cite{GisellePKH19} highlighted the significance of selecting appropriate LF data sources within the scope of MFO. In a related study, Park et al.~\cite{ParkHH17} and Brevault et al.~\cite{BrevaultBH20} examined the use of Gaussian processes in multi-fidelity surrogate modeling, comparing them to other regression models. Peherstorfer et al.~\cite{PeherstorferWG18} discussed fidelity management strategies, detailing their effectiveness in optimization, uncertainty propagation, and inference tasks. More recently, Zhou et al.~\cite{ZhouZHM22} compiled a comprehensive book on hierarchical, non-hierarchical, and sequential multi-fidelity surrogate modeling methods, including surrogates-assisted efficient global optimization. This book also provides practical insights into applying multi-fidelity techniques in various engineering design challenges, such as reliability and robust design optimization. Complementing the survey papers, numerous Ph.D. theses have delved into multi-fidelity methods across diverse domains. These studies cover a range of topics, including sensitivity analysis~\cite{Gratiet13}, aerothermal analysis~\cite{Santos21}, encoder-decoder networks~\cite{Partin22}, simulation of wax deposition~\cite{Farah23}, and uncertainty quantification in cardiovascular hemodynamics~\cite{Masuda22}. In addition, these theses have explored the application of MFO in practical engineering scenarios, such as the optimization of aircraft~\cite{Haas12}, high-lift devices~\cite{Demange18}, wing aerofoils in multidisciplinary optimization frameworks~\cite{Berci11}, and comprehensive multidisciplinary system designs~\cite{March12}.

The overarching goal of this paper is to offer a holistic survey on MFO, enabling researchers with a comprehensive understanding of this dynamic and expanding field. Supported by data-driven analytics, this survey provides insights into pivotal trends, methodologies, benchmark problems, real-world applications, limitations, and future directions in this field. The organization of this survey is as follows. In~\pref{sec:topic}, we start with a text mining by using curated meta-features extracted from over $1,200$ articles collected from various sources. Then, in Sections~\ref{sec:surrogate_models} to~\ref{sec:fidelity_management}, we conduct an in-depth analysis of the three core algorithmic components when designing MFO algorithms, including multi-fidelity surrogate models, optimizers, and fidelity management strategies. Subsequently, in Sections~\ref{sec:benchmark_problems} and~\ref{sec:applications}, we respectively overview the current benchmark test problems and applications of MFO according to the taxonomy of different sectors broadly spanning machine learning, engineering design and optimization, and scientific discovery. In~\pref{sec:challenges}, we plan to shed some lights on the challenges and future directions of MFO, before concluding this paper in~\pref{sec:conclusion}.


\section{Meta-analysis of multi-fidelity optimization literature: A text mining approach}
\label{sec:topic}

\begin{figure}[t!]
    \centering
    \includegraphics[width=\linewidth]{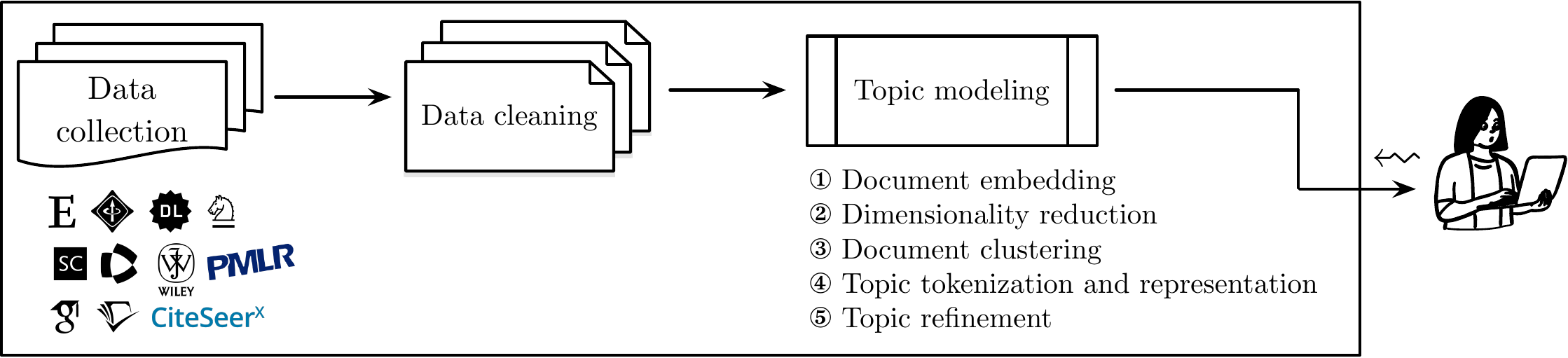}
    \caption{The flowchart of the text mining framework underpinning our survey. It is a closed-loop process where human experts' feedback will be fed back into data collection, data cleaning, and topic modeling steps to improve the quality of text mining results.}
    \label{fig:survey_flowchart}
\end{figure}

In this paper, we have developed a robust text mining framework\footnote{Our text mining framework is open source, and the source code can be found in our \href{https://github.com/COLA-Laboratory/Data_Driven_Survey}{Github repository}.} to underpin our survey. The following paragraphs will start by introducing four building blocks of this framework. Subsequently, we will discuss the key outcomes and insights derived from our text mining process.

\subsection{Text Mining Framework}
\label{sec:data_collection}

As illustrated in the flowchart presented in~\pref{fig:survey_flowchart}, our approach embodies a closed-loop process, wherein human feedback is continually integrated into the text mining process to refine and improve the analysis results.

\begin{enumerate}[Step 1.]
    \item\underline{\textit{Data collection}}: To identify relevant articles for this survey, we conducted a comprehensive search across nine divserse academic databases including \href{https://www.sciencedirect.com/}{\textsc{ScienceDirect}}, \href{https://ieeexplore.ieee.org/Xplore/home.jsp}{\textsc{IEEE Xplore}}, \href{https://dl.acm.org/}{\textsc{ACM Digital Library}}, \href{https://www.springer.com/gp}{\textsc{Springer}}, \href{https://www.wiley.com/en-gb}{\textsc{Wiley}}, \href{https://taylorandfrancis.com/}{\textsc{Taylor \& Francis}}, \href{https://www.scopus.com/home.uri}{\textsc{Scopus}}, \href{https://www.webofscience.com/wos/woscc/basic-search}{\textsc{Web of Science}}, as well as \href{https://proceedings.mlr.press/}{PMLR}. Additionally, we leveraged three major search engines: \href{https://scholar.google.com/}{\textsc{Google Scholar}}, \href{https://www.semanticscholar.org/product/api}{Semantic Scholar}, and \href{https://citeseerx.ist.psu.edu/Wiley}{\textsc{CiteSeerX}}. Due to copyright protection from the major publishers, instead of accessing PDFs, our information retrieval protocol primarily involved retrieving relevant abstracts containing a range of search terms in various combinations. These terms included $\langle$\textit{multi-fidelity optimization}, \textit{multi-level optimization}, \textit{multi-discipline optimization}, \textit{multi-source optimization}$\rangle$, $\langle$\textit{evolutionary algorithm}, \textit{genetic algorithm}, \textit{swarm intelligence}, \textit{meta-heuristics}$\rangle$, $\langle$\textit{Bayesian optimization}, \textit{bandit optimization}$\rangle$.

    \item\underline{\textit{Data cleaning}}: To curate the data collected in Step $1$, we undertook a meticulous data cleaning process. This involved the removal of invalid articles, specifically those lacking content or abstracts. Additionally, we excluded articles with extremely short abstracts, defined as those containing fewer than $10$ words.

    \item\underline{\textit{Topic modeling}}: To explore latent themes within the collected documents, we employed \texttt{BERTopic}~\cite{Grootendorst22}, a Transformer-based topic modeling tool, for creating dense clusters of topics. As shown in~\pref{fig:survey_flowchart}, the topic modeling process consists of five steps. Each document is first transformed into an embedded representation using the \texttt{Sentence-BERT} framework~\cite{ReimersG19}. The document embeddings then undergo dimensionality reduction through \texttt{UMAP}~\cite{McInnesH18}. The reduced dimensions facilitate document clustering via \texttt{HDBSCAN}~\cite{McInnesHA17}. Subsequently, topics are tokenized, and topic representations are extracted from the document clusters using a customized \texttt{TF-IDF} approach~\cite{Joachims97}. Lastly, these extracted topics are fine-tuned using maximal marginal relevance~\cite{CarbonellG98} and part-of-speech tagging in \texttt{KeyBERT}~\cite{Grootendorst20}.

    \item\underline{\textit{Human insepction and feedback}}: Last but not the least, the generated analysis will undergo a thorough inspection by human experts. Their valuable feedback will be fed back into each of the previous three steps, thereby fine-tuning the text mining outcomes.
\end{enumerate}

\subsection{Text Mining Results}
\label{sec:mining_results}

In this section, we present a detailed discussion of the analysis results derived from our text mining framework. We begin by summarizing key statistics and patterns observed in the MFO literature. This is followed by an exploration of the specific topics identified through our topic modeling process. The insights gained from these exploratory analyses have been instrumental in shaping the structure of this survey paper, and we aim to elucidate how they inform and guide the content in the subsequent sections.

\subsubsection{Text mining from publications}
\label{sec:analysis_publication}

\begin{figure}[t!]
    \centering
    \includegraphics[width=\linewidth]{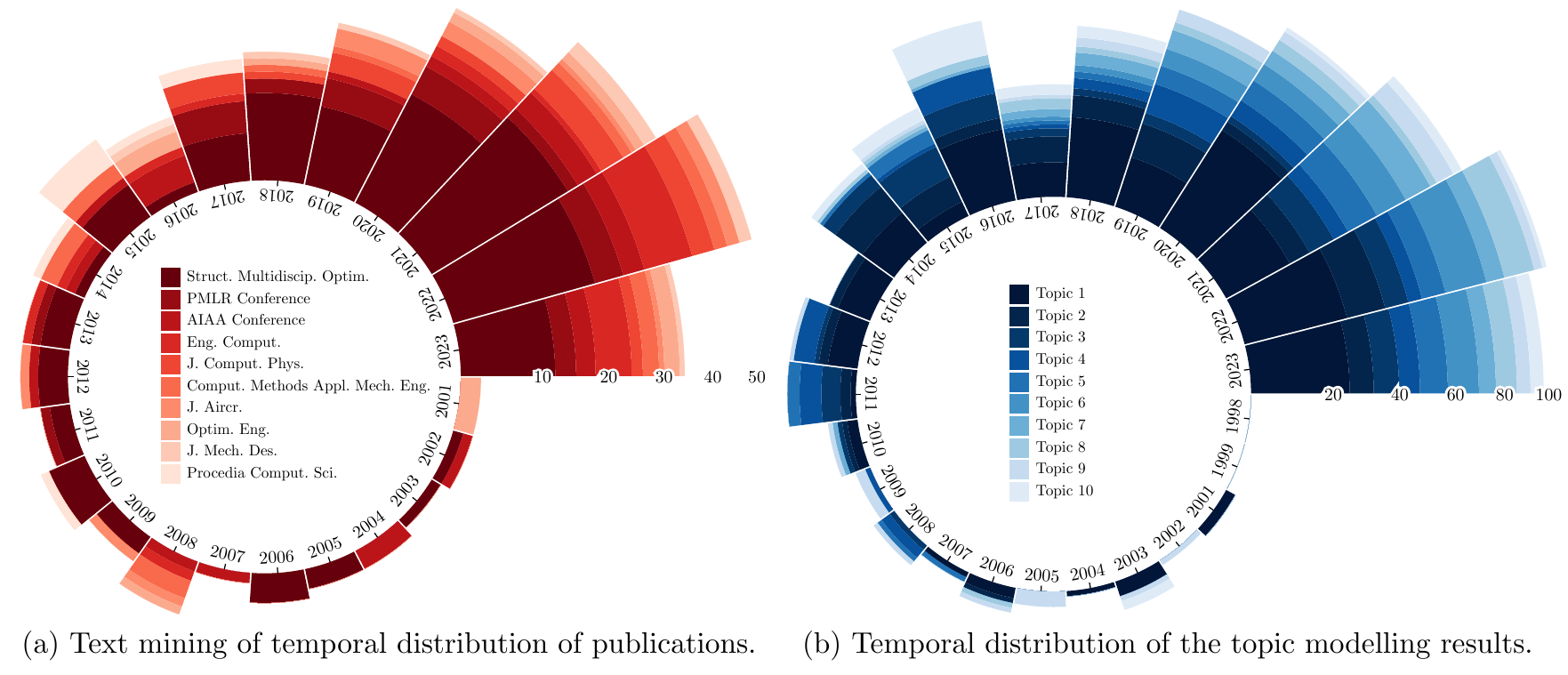}
    \caption{The temporal statistics from 1998 to 2023 regarding: (a) the publication trend of ten most active publication venues (represented in their acronyms) , and (b) the progression of top ten most popular topics learned from the topic modeling.}
    \label{fig:journal_trend}
\end{figure}

\begin{figure}[t!]
    \centering
    \includegraphics[width=\linewidth]{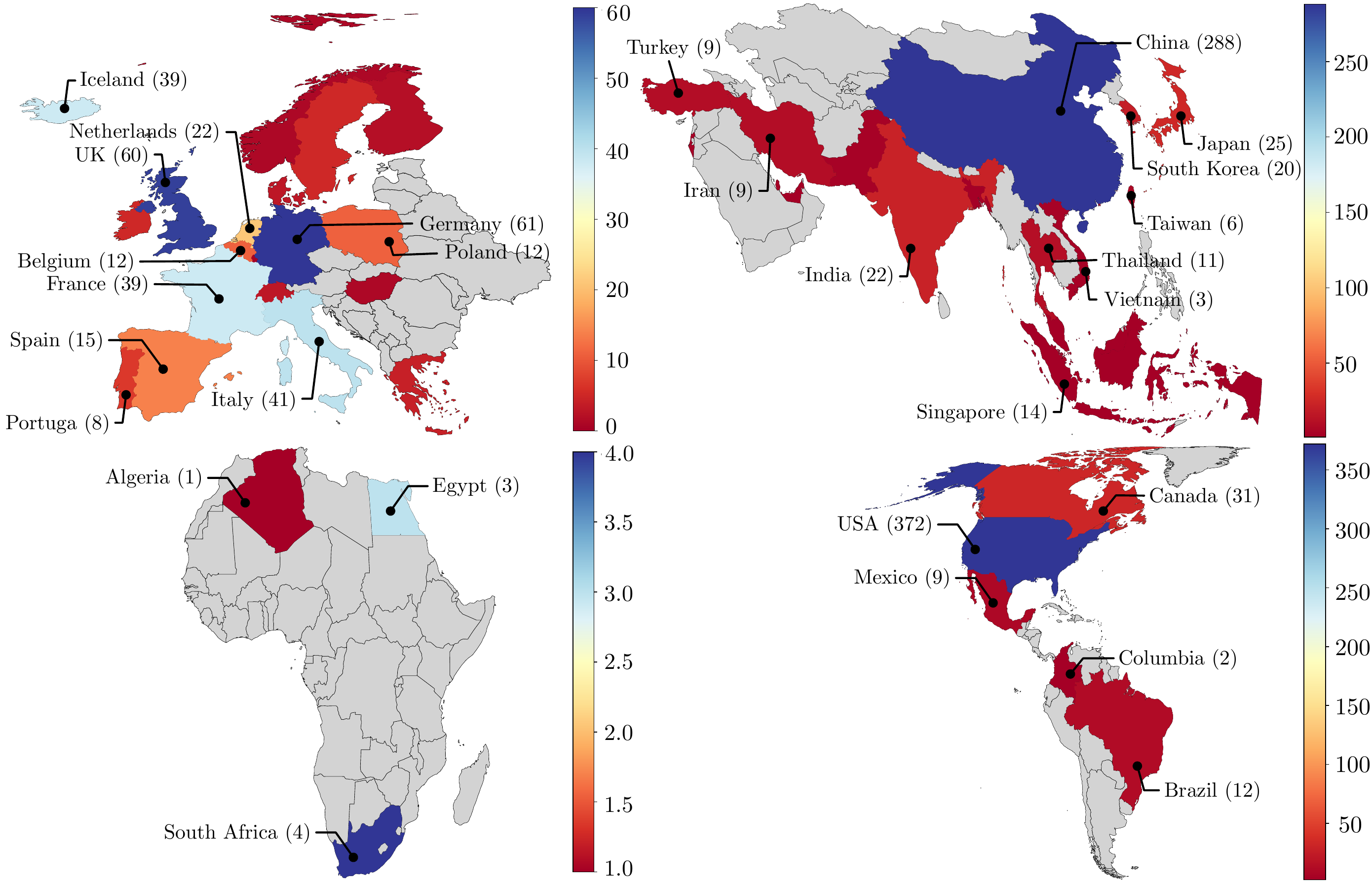}
    \caption{Geographical distribution of publications across different regions. To minimize information cluttering, we only list at most top ten most productive regions in each continent.}
    \label{fig:nation_trend}
\end{figure}

After completing the initial Steps $1$ and $2$ of our text mining framework, we compiled a comprehensive dataset of $1,242$ articles, spanning from 1998 to 2023. Analysis of publication trends, as illustrated in~\pref{fig:journal_trend}(a), reveals a significant evolution in journal preferences within the MFO field. In particular, \href{https://www.springer.com/journal/158}{\textsc{Structural and Multidisciplinary Optimization}} emerges as the foremost journal in this domain. Conference proceedings, particularly those in \href{https://proceedings.mlr.press/}{\textsc{PMLR}} and \href{https://www.aiaa.org/publications/Meeting-Papers}{\textsc{AIAA}}, rank closely behind this journal, though their publication volume is four times less. Further, it is also interesting to note that there is a steady increase in publications after 2018, suggesting a growing interest and advancements in MFO research. Furthermore, \pref{fig:nation_trend} extends our exploration beyond the venue of publications to the geographical distribution of contributing institutions. This global perspective highlights the USA, China, and Europe are the most productive regions contributing to the MFO research, underscoring their active involvement and possibly indicating robust research communities and funding opportunities in these countries.

\subsubsection{Text mining from topic modeling}
\label{sec:analysis_topics}

\begin{figure}[t!]
    \centering
    \includegraphics[width=\linewidth]{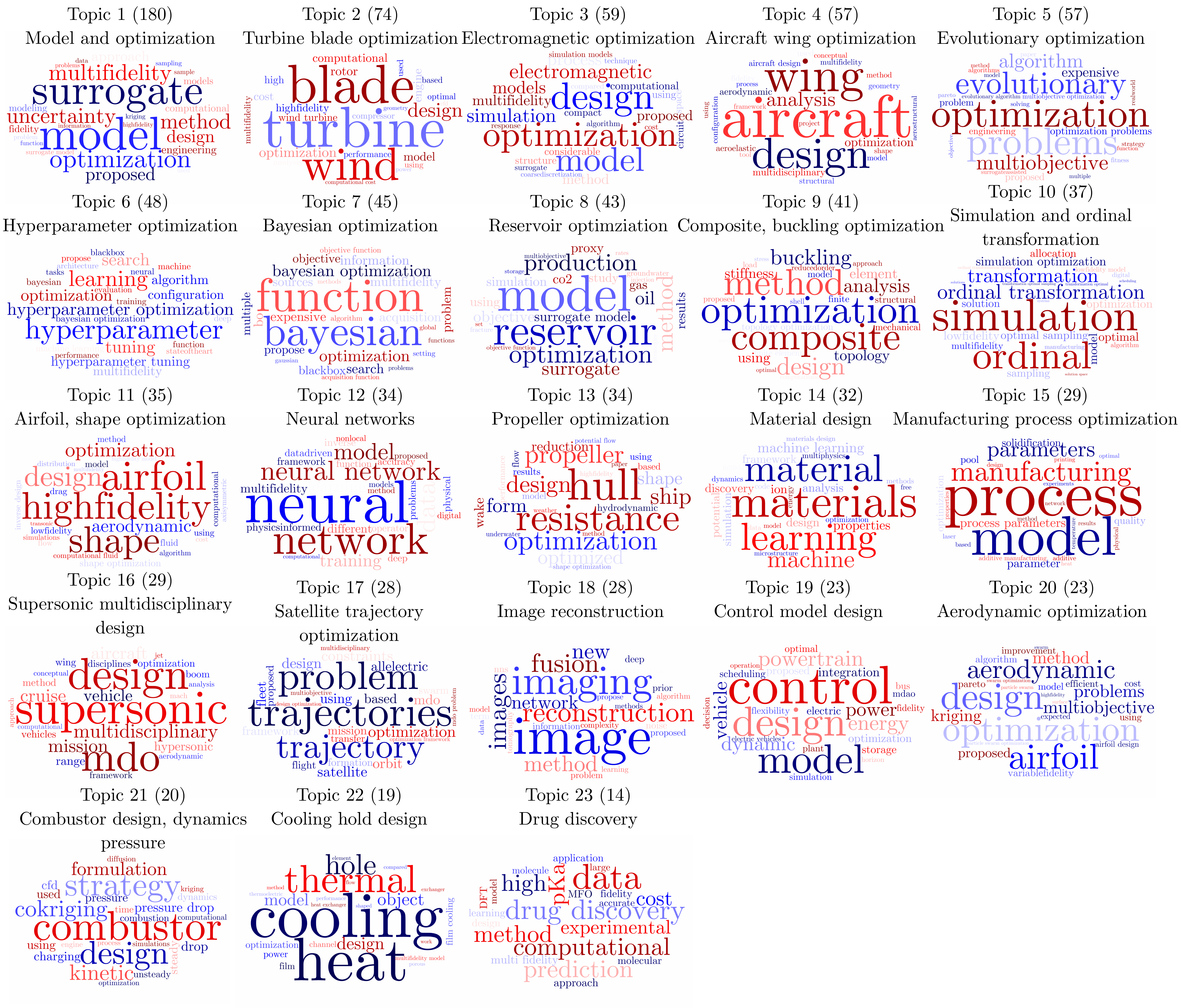}
    \caption{Word clouds of $23$ topics learned by our topic modeling pipeline from the collected articles. Note that the topics are indexed according to a descending order of the number of underpinned articles.}
    \label{fig:word_clouds}
\end{figure}

After completing the topic modeling in Step $3$ of our text mining framework, $23$ distinct topics have been identified. These are depicted as word clouds in~\pref{fig:word_clouds}, where the size of each keyword, along with the density of the corresponding color, are proportional to its frequency within the corresponding topic. Notably, only four topics focus on algorithmic aspects---surrogate modeling, evolutionary optimization, Bayesian optimization, and neural networks---while the remaining $19$ topics predominantly pertain to specific application domains.

In~\pref{fig:journal_trend}(b), we examined the temporal trends of these $23$ identified topics from 1998 to 2023. The analysis, considering both~\pref{fig:word_clouds} and~\pref{fig:journal_trend}(b), reveals a consistent emphasis on \lq model and optimization\rq\ as the most active topic over the past two decades. Interestingly, \lq bandit optimization\rq, despite being a search term, is underrepresented in the word clouds. A detailed examination of related articles indicates that approximately $10$ of them specifically address bandit optimization, focusing mainly on variants of Hyperband~\cite{LiJDRT17} for hyperparameter optimization in machine learning.

\begin{figure}[t!]
    \centering
    \includegraphics[width=0.8\linewidth]{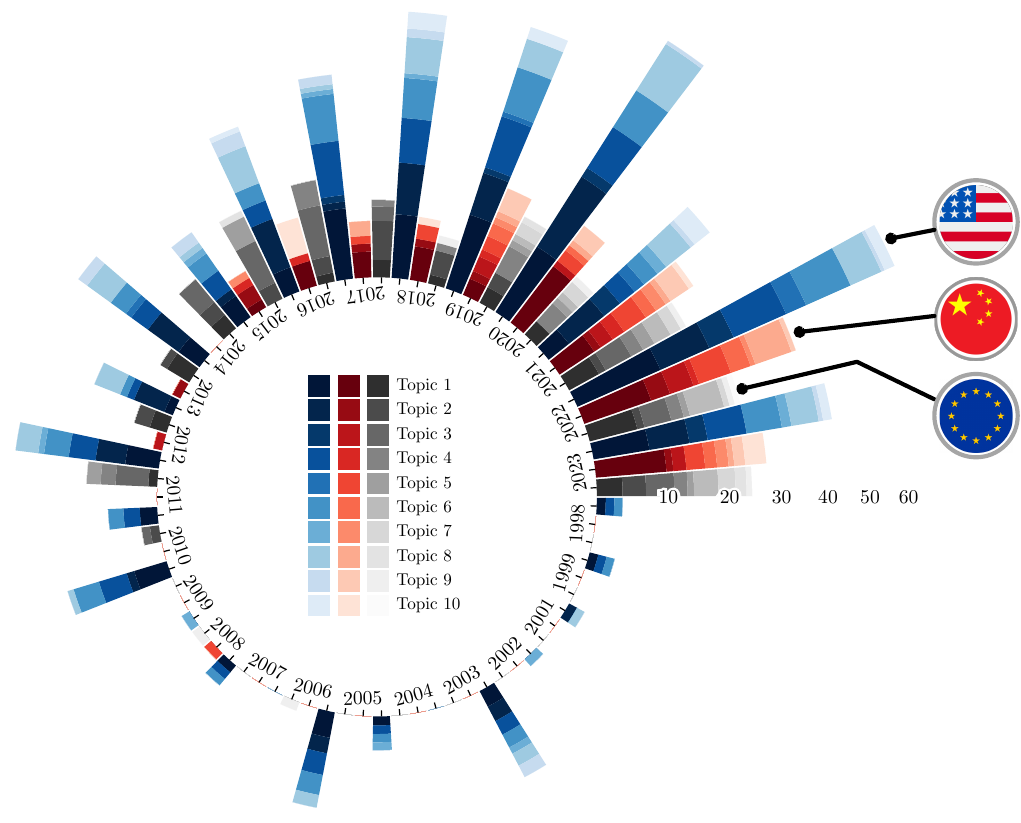}
    \caption{A comparison among USA, China, and EU (including UK and Switzerland) about the temporal statistics from 1998 to 2023 regarding the progression of topics. To mitigate information cluttering, we only display the top ten most popular topics.}
    \label{fig:topic_trend_nation}
\end{figure}

Regarding application-specific topics, our analysis categorizes them into three key domains---engineering design optimization, hyperparameter optimization in machine learning, and scientific discovery. In particular, we find that the topic of neural networks is highly intertwined with applications across domains. These domains, each with their unique challenges and advancements, will be further discussed in~\pref{sec:applications}. The geographical analysis in~\pref{fig:nation_trend} highlights the USA, China, and the EU, including the UK and Switzerland, are the leading regions of MFO research. An in-depth exploration of their topic trends, illustrated in~\pref{fig:topic_trend_nation}, reveals distinct research interests. While \lq model and optimization\rq\ remains a dominant theme in all regions, it garners more attention in China. Conversely, in the USA, topics such as \lq aircraft design\rq, \lq hyperparameter optimization\rq, and \lq Bayesian optimization\rq\ are more actively pursued. Meanwhile, researchers in China and EU show significant focus on \lq electromagnetic applications\rq, \lq reservoir optimization\rq, and \lq evolutionary optimization\rq.

\subsubsection{Taxonomy of MFO research}
\label{sec:skeleton}


Leveraging the insights derived from our text mining analysis, we have constructed a comprehensive taxonomy of the research components in MFO. This taxonomy is visually represented in the form of a mind map, shown in Fig. A$1$ of the \textsc{Appendix}\footnote{The \textsc{Appendix} document can be downloaded from \url{http://tinyurl.com/25y8h3s4}.}, forming the foundational structure for the remainder of this survey.
\begin{itemize}
    \item As depicted in the word clouds in~\pref{fig:word_clouds}, multi-fidelity surrogate modeling stands out as the most extensively researched topic within the MFO domain. In line with this trend, \pref{sec:surrogate_models} is dedicated to providing an in-depth overview of the seven most prevalent surrogate models in MFO, informed by the statistical analysis of the surrogate models extracted through our topic modeling pipeline, as shown in \pref{fig:mining_model_management}(a).

        \begin{figure}[t!]
            \centering
            \includegraphics[width=\linewidth]{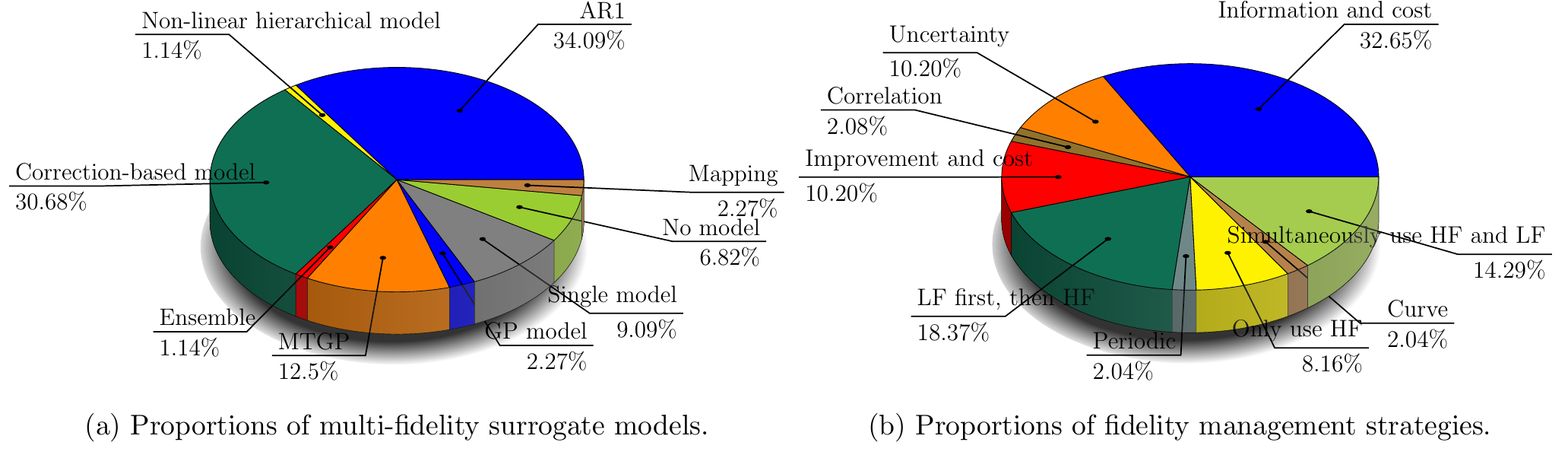}
            \caption{Pie charts of the proportion of number of articles regarding (a) different multi-fidelity surrogate modeling methods, and (b) fidelity management strategies considered in this survey.}
            \label{fig:mining_model_management}
        \end{figure}

    \item In~\pref{sec:optimizer}, we explore the current advancements in optimizers, an essential component of MFO. This section is anchored in Bayesian optimization (BO)~\cite{Garnett23} and surrogate-assisted evolutionary algorithms (SAEA)~\cite{HeZGJ23}, as highlighted by the word clouds shown in~\pref{fig:word_clouds}. Additionally, we extend our exploration to include bandit optimization, in view of its recognized efficacy in hyperparameter optimization, despite being underrepresented in our word cloud analysis.

    \item Fidelity management, an essential algorithmic component in MFO, orchestrates transitions between different fidelity levels, thus optimizing computational efficiency. The variety of fidelity management methods identified by our topic modeling pipeline is presented in~\pref{fig:mining_model_management}(b), offering a statistical overview of various approaches in the MFO field. In~\pref{sec:fidelity_management}, we categorize these methods into two primary types---fixed and adaptive.

    \item While benchmark test problems were not explicitly identified by our text mining framework, they are crucial for evaluating and benchmarking the performance of different MFO algorithms. Therefore, \pref{sec:benchmark_problems} is dedicated to discussing the development of benchmark test problems.

    \item As pointed out in~\pref{sec:analysis_topics}, a significant portion of MFO research is application-specific. In~\pref{sec:applications}, we categorize the diverse applications identified in \pref{fig:word_clouds} into three primary domains---hyperparameter optimization, engineering design and optimization, and scientific discovery, along with applications of neural networks.
\end{itemize}




\section{Multi-Fidelity Surrogate Modeling}
\label{sec:surrogate_models}

In this section, we offer a comprehensive overview of various strategies for constructing surrogate models from multi-fidelity data, with a particular focus on their application in MFO. The primary challenges in this area arise not only from modeling the black-box response surface at each fidelity level but also from accurately capturing the complex interrelations between these levels. To tackle these challenges, existing methodologies are broadly classified into seven categories---\textit{single model} methods, \textit{space-mapping} methods, \textit{correction-based} methods, \textit{autoregressive-based} methods, \textit{multi-task Gaussian process-based} methods, \textit{nonlinear hierarchical} and \textit{physics-informed neural networks} methods. For each category, we aim to highlight their unique features, benefits, and limitations. Note that this taxonomy is slightly different from the text mining outcomes shown in~\pref{fig:mining_model_management}(a), because we synthesize homogeneous surrogate models.

\subsection{Single Model Methods}
\label{sec:single_model}

As the name suggests, single-model based methods involve constructing independent surrogate models for each fidelity level, without assuming any correlation between the low- and high-fidelity data. Common regression techniques used in this approach include response surface models (RSMs)~\cite{GiuntaBKBGHMW97}, Gaussian process (GP)~\cite{LiuKZ16,KandasamyDOSP19,FolchLSWTWM23}, and radial basis function (RBF)~\cite{YiSS20a,LiuKZ16,KampolisZAG07}. RSMs have been shown to effectively address problems with low-order nonlinearity~\cite{ZhaoX10}, while GP and RBF are generally more suitable for handling high-order nonlinear problems~\cite{ZhaoX10}. In a different vein, \cite{HabibSR19} proposed an approach that concurrently trains multiple models, selecting only the model with the highest training accuracy for surrogate modeling.

\subsection{Space-Mapping Methods}
\label{sec:space_mapping}

The space-mapping method, pioneered in~\cite{BandlerBCGH94}, aims to link low- and high-fidelity models by aligning their input variables. Its key idea is to start with the optimization at the LF space, followed by an inversion of the mapping to pinpoint trial points in the HF space. However, this method has no convergence guarantee, even to a local HF optimum. To accelerate convergence, Bandler et al.~\cite{BandlerBCGHM95} introduced aggressive space-mapping, which, while faster, hinges on bijective mapping and equal dimensional design variables across LF and HF spaces. Despite these requirements, non-unique parameter extractions can impede convergence. To address this, the trust-region aggressive space-mapping algorithm was proposed in~\cite{BandlerBCGHM98} but it risks of being trapped by local optima due to misalignment between fidelity levels. In~\cite{BakrBGM99}, a novel hybrid strategy was proposed to combine space-mapping optimization with direct approaches, flexibly transitioning between the two to circumvent convergence difficulties. Beyond linear methods, neural networks~\cite{BakrBIRZ00} and manifold methods~\cite{RenLKT16} have offered alternative ways to integrate different fidelity levels.

\subsection{Correction-based Methods}
\label{sec:correction}

The basic idea here lies in refining HF models using an offset or scaling factor derived from LF models. This approach generates multi-fidelity surrogate models to approximate these additive and/or multiplicative corrections.

Early efforts in this domain often utilized first-order (e.g.,~\cite{Haftka91,ChangHGK93,LewisN00}) or second-order (e.g.,~\cite{EldreGCd04,GanoPRBS04,GanoRS05}) Taylor series to estimate the offset or scaling factor between LF and HF models. While robust, these methods typically require gradient information, which may not be available in black-box scenarios. First-order methods provide linear approximations, whereas second-order methods, more accurate but computationally intensive, can be less robust against input noise due to their reliance on higher-order polynomials~\cite{MarduelTT06}. Nonetheless, quadratic correction has been shown to be more effective than linear approximation in empirical studies~\cite{MarduelTT02}.

The correction-based methods have also incorporated machine learning approaches, such as neural networks (NNs) and support vector regression (SVR). Multi-layer feedforward NNs, with their universal approximation capabilities, have been used to model nonlinear relationships between low- and high-fidelity models in various applications (e.g.,~\cite{WarnerM96,KimKN07}). Due to the scalability and lightweight structure of RBF NNs, they have been widely applied in transonic~\cite{TyanNL15} and subsonic aerofoil optimization~\cite{NguyenTL15}, as well as in hybrid methods where the scaling factor and network parameters are learned either concurrently (e.g.,~\cite{DurantinRDG17,SongLSZ19,WangJYJ20}) or separately~\cite{YiGLSL19}. Two-stage modeling approaches, combining linear regression and least-square SVR, have been proposed to refine HF data approximations~\cite{ZhengSGJQ15,ZhouSJZS15}. For instance, Shi et al.~\cite{ShiLSS20} used $\epsilon$-SVR to model relationships between inputs and outputs, after projecting low- and high-fidelity data into a high-dimensional feature space. In~\cite{SunLL12}, an ensemble of regression models was trained as a correction model, optimizing draw-bead restraining forces for automobile panels using the compensated LF model.

In addition to these techniques, correction-based methods have also utilized regression models from experimental design. Given the global approximation capabilities of RSMs, they have been employed to estimate discrepancies between low- and high-fidelity models, such as stress intensity factors in composite panels~\cite{Vitali98HS,VitaliHS02} and aerodynamic solutions in high-speed transport design~\cite{KnillGBGMHWA99,SunLZXYL11}. Kriging, another widely-used interpolation model, has been applied to approximate differences or ratios between low- and high-fidelity models, facilitating correction and enabling the use of multi-fidelity surrogate models in sequential model-based optimization (e.g.,~\cite{GanoRS05,GanoRMS06,LiuC14,XiongCT08,ZhouWCJSH17,ZhuWC14,ZhouSJGWS16,ZhouWXJ21,LiTLSW21}).

\subsection{Autoregressive-based (AR1) Methods}
\label{sec:ar1}

Autoregressive methods, based on the assumption of a linear relationship between different fidelity levels~\cite{KennedyO00}, operate similarly to co-Kriging~\cite{JournelH89}. These methods enhance HF output prediction by incorporating LF data as supplementary variables. Typically, the first Kriging model manages the LF data, while the second model focuses on a multiplicative scaling factor that bridges the gap between low- and high-fidelity outputs. Studies such as~\cite{ParkHH17} and~\cite{HuangGZ13} have shown that AR1 methods outperform correction-based methods in terms of approximation capabilities and convergence speed, given the same amount of HF samples. Their robustness and effectiveness in modeling fidelity correlations have led to widespread applications in design optimization tasks like aerodynamic shape optimization~\cite{ForresterSK07,DingK18}, shaped film cooling hole optimization~\cite{ZhangLCSY19}, and engine optimization using medial meshes~\cite{YongWTKS19}.

Another line of efforts have been on refining co-Kriging methods. For instance, Han et al.~\cite{HanZG12,ZimmermannH10} suggested an augmented correlation matrix incorporating all observations to simplify co-Kriging training. They also explored integrating gradient information into co-Kriging, resulting in a hierarchical Kriging model that better aligns LF functions with HF data~\cite{HanZG12}. Although this approach shows promising results, such as in \textsc{RAE} $2822$ airfoil optimization, it comes with increased computational costs due to the aggregation of observations from various fidelity levels. To address this, Gratiet~\cite{Gratiet14} introduced a recursive method for constructing multi-level co-Kriging models, using independent Kriging models for each level. These techniques have been applied in diverse fields, ranging from uncertainty analysis in computational fluid dynamics~\cite{PalarS17} and aerodynamic shape optimization~\cite{ZhangHZ18,LiuHZSSGT19}, to the shape optimization of super-cavitating hydrofoils~\cite{BoPBK18}. Different from other methods, which are primarily applicable to problems with hierarchical LF models, Zhang et al.~\cite{ZhangWJCZ22} developed a non-hierarchical co-Kriging modeling method designed to combine HF data with numerous non-hierarchical LF data to construct a multi-fidelity surrogate model.

\subsection{Multi-task Gaussian Process (MTGP) Methods}
\label{sec:mtgp_method}

In contrast to AR1 methods, MTGP methods introduce enhanced expressiveness by modeling different fidelities as either linear or nonlinear combinations of latent GPs~\cite{AlvarezRL12,LiCY23,YangL23,ChenL23}. This approach effectively captures the correlations among various fidelities by learning weighted coefficients from samples across all fidelity levels~\cite{BrevaultBH20}. As the pioneer, Swersky et al.~\cite{SwerskySA13} proposed employing a linear model of coregionalization (LMC) to concurrently model HF problems with LF data, using subsets of validation data as LF representations. Subsequent works~\cite{TakenoFTKSTK20,MossLR20,TakenoFTKSTK22,LinQCZH22,FolchLSWTWM23} have further investigated MTGP variants to model correlations between different fidelities, with a focus on designing efficient acquisition functions for MFO. Mikkola et al.~\cite{MikkolaMFK23} proposed a dual-surrogate strategy to mitigate performance issues caused by unreliable LF data. This approach involves constructing a GP surrogate from HF data alongside an LMC surrogate using all available data. The LMC solution is utilized only if its predictive variance is below a predetermined threshold and the correlation between low- and high-fidelity data exceeds a certain level. Poloczek et al.~\cite{PoloczekWF17} introduced a model representing LF problems as a combination of HF data and an independent error term, following an independent GP. This model is particularly effective when LF data significantly diverge yet share a common goal and has been adapted to incorporate correlations between discrepancies across different fidelities. Complementing these approaches, Zhang et al.~\cite{ZhangHLK17} developed a convolved MTGP to simultaneously model low- and high-fidelity data, enabling an efficient approximation of multi-fidelity predictive entropy search.

\subsection{Nonlinear Hierarchical Methods}
\label{sec:deep_gp}

Nonlinear hierarchical methods operate under the premise that different fidelities can be hierarchically structured, with the composition function between them being either linear or nonlinear to reflect their interrelationship~\cite{CutajarPDLG19}. A notable development in this area is~\cite{PerdikarisRDLK17} that adapted the AR1 method for nonlinear relationships by employing a nonlinear mapping between fidelities. They proposed utilizing a disjoint hierarchical structure where each fidelity level is modeled by a distinct GP, trained on its variables and the posterior mean of the preceding LF level. Requeima et al.~\cite{RequeimaTBT19} adopted a similar structure but trained each GP with outputs from all previous fidelity levels. However, this model does not allow GPs at lower fidelities to be updated with HF training points, limiting the potential of benefiting from HF data~\cite{BrevaultBH20}.

Deep GP (DGP), characterized by its nested GP layers, uses functional compositions of GP to map inputs to outputs~\cite{DamianouL13}. This structure is effectively harnessed in multi-fidelity DGP (MF-DGP) in~\cite{CutajarPDLG19}, where each layer represents a different fidelity. To address the computational challenges posed by intractable marginal likelihood, variational inference is employed for collective training across all layers~\cite{SalimbeniD17}. Despite its increased expressiveness, MF-DGP introduces a substantial growth of hyperparameters and computational demands, compared to the methods in~\cite{PerdikarisRDLK17} and~\cite{RequeimaTBT19}.

Deep NNs (DNNs) have also been employed to model complex fidelity relationships. For instance, Li et al.~\cite{LiXKZ20} proposed a stacked configuration of individual NNs, with each representing a different fidelity level. Information is propagated through this structure by feeding both input and output from the previous fidelity into the subsequent NN. This setup, however, is restricted to capturing correlations between successive fidelity levels. To model more intricate correlations, including highly nonlinear and non-stationary relationships across all fidelities, Li et al.~\cite{LiKZ21} enhanced their method by incorporating data from all preceding fidelities into each layer. In a similar vein, Xu et al.~\cite{XuSZ21} proposed utilizing a hierarchical regression framework for multi-fidelity surrogate modeling.

\subsection{Multi-Fidelity Physics-Informed Neural Networks (PINNs)}
\label{sec:mf_pinn}

In addition to the previous data-driven methods, recent research has also focused on the development of multi-fidelity PINNs~\cite{RaissiPK17}, universal function approximators capable of incorporating physical laws, such as partial differential equations (PDEs), into the training process. Despite their expressive capability, PINNs often require a large amount of HF data to accurately identify unknown parameters in nonlinear PDEs. As in~\cite{MengK20} and~\cite{LuDKRKS20}, recent advancements can be categorized into transfer learning, simultaneous training, and consecutive training approaches. Specifically, transfer learning methods such as~\cite{Chakraborty21} and~\cite{HaradaRM22} involve initially training the network on LF data before applying a correction to align the LF output with HF data. This approach is particularly useful when the underlying physics is not precisely known or when generating LF data through solvers is prohibitively expensive. In simultaneous training, as explored in~\cite{MengWFTK21}, both linear and nonlinear correlations between low- and high-fidelity data are learned by training low- and high-fidelity networks concurrently. This method is adept at handling complex correlations between datasets and can even be effective without incorporating physical constraints~\cite{MengK20}. Lastly, the consecutive training approach, pioneered in~\cite{LiuW19}, involves training two separate physics-constrained neural networks for low- and high-fidelity data, followed by a third network that learns the correlation between low- and high-fidelity outputs. This method provides a robust framework for capturing and understanding the nuanced relationship between different fidelity levels.

\subsection{Discussions}
\label{sec:discussion_model}

There are four key takeaways from the multi-fidelity surrogate modeling methodologies reviewed in this section.
\begin{itemize}
    \item As outlined in Table A$4$ of the \textsc{Appendix}, except for the single model method, the other multi-fidelity surrogate modeling techniques can be represented in a unified form\footnote{While the MTGP method presents a distinct format, it can be transformed to align with~\pref{eq:model} by solving a system of linear equations. For a detailed derivation process, please refer to Section B.$2$ of the \textsc{Appendix}.}:
        \begin{equation}
            f_\mathrm{h}(\mathbf{x})=\rho(\mathbf{x})g(f_\mathrm{l}(\mathbf{x}))+\epsilon(\mathbf{x}),
            \label{eq:model}
        \end{equation}
        where different components of~\pref{eq:model} are utilized to varying degrees in the different methodologies:
        \begin{itemize}
            \item $g(\cdot)$ is a constant in correction-, autoregressive-, and MTGP-based methods, allowing them to capture linear relationships, as shown in the first line of Figs.~\ref{fig:cop_model}(b) to (d). However, these methods are less effective with nonlinear functions, as seen in the corresponding subfigures in the second line of Figs.~\ref{fig:cop_model}(a) to (e).

            \item Conversely, $g(\cdot)$ represents a nonlinear transformation function in the nonlinear hierarchical approach, enabling an effective capture of nonlinear relationships, as depicted in the second line of~\pref{fig:cop_model}(e). However, this method is less efficient when dealing with linear functions, as shown in the first line of~\pref{fig:cop_model}(e).
            
            \item Further, $g(\cdot)$ is encoded with physics meaning of the underlying system, such as PDEs, to align with the intrinsic system dynamics.

            \item The scaling term $\rho(\cdot)$, applied to the LF function, is set as a constant in the AR1 method. This term does not influence the MTGP and nonlinear hierarchical methods.
        \end{itemize}

\begin{figure}[t!]
    \centering
    \includegraphics[width=1.0\textwidth]{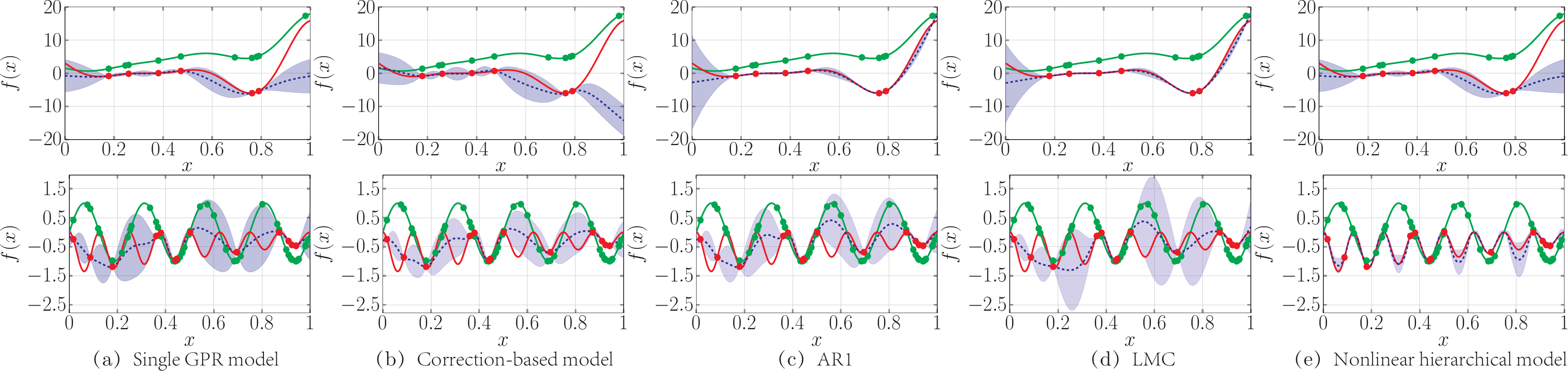}
    \caption{An illustration of surrogate models of an HF function created using single GPR, correction-based, AR1, LMC, and nonlinear hierarchical models. The HF function is depicted in red, the LF function in green, and the surrogate models, along with their estimated uncertainties, are represented in blue. The HF and LF functions of the first line are $f_\mathrm{h}(x) = (6x - 2)^2 \sin(12x - 4)$ and $f_\mathrm{l}(x) =0.5f_\mathrm{h}(x)-10x$, while the HF and LF functions of the second line are $f_\mathrm{h}(x) = (x - \sqrt{2}){f_\mathrm{l}(x)}^2$ and $f_\mathrm{l}(x) =\sin(8\pi x)$.}
    \label{fig:cop_model}
\end{figure}

    \item The complexity of the seven multi-fidelity surrogate modeling methods increases with their capacity to capture relationships between LF and HF models. Moving from simpler to more complex methods broadens the range of relationships that can be captured, at the cost of increased complexity. As summarized in Table A$4$ of the \textsc{Appendix}, each method has distinct features and drawbacks. As the pie chart shown in~\pref{fig:mining_model_management}(a), the correction-based method as the most extensively studied approach.


    \item There exists an inherent trade-off between the expressiveness of a multi-fidelity surrogate model and the simplicity of its training process. Simpler models, while easier to train, often have limited expressiveness. Conversely, complex models that capture intricate low- and high-fidelity relationships entail significant computational costs for training and inference. In accordance with the no-free-lunch theorem~\cite{DolpertM97}, an empirical study encompassing various benchmark test problems and aerospace design problems~\cite{BrevaultBH20} reveals that no single surrogate modeling method consistently excels across all scenarios. This underscores the importance of selecting a model based on specific use cases.

    \item Choosing an appropriate multi-fidelity surrogate model depends on the specifics of the problem and the available data. Factors such as data volume, the complexity of low- and high-fidelity relationships, computational constraints, and desired accuracy levels are crucial in determining the optimal method for a particular scenario. Advances in machine learning and optimization are continually opening new avenues for innovative modeling methods and enhancements to existing ones. Thus, this field remains ripe for further research. Additionally, with the growth of high-dimensional, multi-fidelity datasets in diverse domains, future research may need to focus on developing scalable and efficient multi-fidelity surrogate models capable of handling such complex data while maintaining peak performance.
\end{itemize}


\section{Optimizer for Multi-Fidelity Optimization}
\label{sec:optimizer}

Several multi-fidelity optimization strategies exist for optimizing a HF function using a LF surrogate. Early attempts, such as those by Alexandrov et al.~\cite{AlexandrovLGGN01,AlexandrovLDLT98,AlexandrovLGGN99}, primarily employed trust regions, known for their proven convergence to a local optimum of the HF function. This convergence is achievable when the LF function value and derivative at the center of the trust region align with those of the HF function. However, the inherent complexities of engineering design often make these approaches less effective. Situations involving non-smooth objectives, experimental results, black-box codes, or numerical methods that are sensitive to noise or susceptible to solution failure can make the gradient of the HF function unavailable or unreliable. Such challenging scenarios necessitate derivative-free methods like BO, SAEA, and bandit optimization. These methods have gained widespread acceptance due to their independence from a HF gradient estimation. The literature review of optimizers for MFO in this section is thus framed along these lines.

\subsection{Bayesian Optimization for Multi-Fidelity Optimization}
\label{sec:MFBO}

The effectiveness of MFO within the BO framework is influenced by several interconnected factors, including the choice of surrogate modeling techniques, the tailored design of acquisition functions, and the application of strategic fidelity management approaches. The selection of a specific modeling technique not only determines the development of acquisition functions but also influences the implementation of appropriate fidelity management strategies. This section reviews existing literature on multi-fidelity BO (MFBO), categorizing it based on the modeling methodologies employed. A dedicated discussion on the aspect of fidelity management is reserved for later in~\pref{sec:fidelity_management}.

\subsubsection{Single model based MFBO}
\label{sec:single_mfbo}

In the realm of MFO, a specific line of research has focused on developing discrete models for each fidelity level, as discussed in~\pref{sec:single_model}. For example, Kandasamy et al.~\cite{KandasamyDOSP19} constructed separate GP models for each fidelity, using an upper confidence bound (UCB) infill criterion to select novel solutions. These solutions are evaluated at the lowest fidelity that exceeds a predefined uncertainty threshold. However, this single-model approach mainly improves the surrogate model's accuracy at a specific fidelity level, which can lead to an inefficient use of resources on LF evaluations if the promising regions within the low- and high-fidelity models significantly diverge. This limitation underscores the necessity for more advanced modeling techniques that effectively capture the relationships across different fidelity levels, thereby achieving a more efficient balance between exploration and exploitation.

\subsubsection{Correction-based MFBO}
\label{sec:correction_mfbo}

In the correction-based MFBO domain, many studies utilize the trust-region framework. This strategy involves constructing an initial multi-fidelity surrogate model using both low- and high-fidelity samples. Optimal solutions within each trust region are evaluated at both fidelity levels, with these assessments serving to update the trust region and the multi-fidelity surrogate model. Model approximations can be local, as demonstrated by second-order Taylor series approaches~\cite{GanoPRBS04,GanoRS05,MarduelTT06}, or global, as seen with RBF NNs~\cite{TyanNL15, NguyenTL15} and Kriging methods~\cite{GanoRMS06}. A significant advantage of such trust-region-based optimization is its theoretical guarantee of converging to a local optimum~\cite{EldreGCd04}, though it may struggle when encountering multi-modal fitness landscapes.

Alternatively, some research adheres to the conventional rigor of BO. For instance, Durantin et al.~\cite{DurantinRDG17} proposed a method where the optimal solution of the multi-fidelity surrogate model is iteratively refined by exploring the entire variable space until a stopping criterion is met. Xiong et al.~\cite{XiongCT08} introduced a Kriging-based MFO that uses a lower bounding criterion with variable weights for iteratively sampling new solutions. These approaches update the multi-fidelity surrogate model by evaluating new samples at both low- and high-fidelity levels, eliminating the need to choose which fidelity level should be used for function evaluations. In contrast, Zhou et al.~\cite{ZhouWCJSH17} employed the predicted variance of a multi-fidelity surrogate model as the acquisition function. It leverages improvements in model accuracy to decide whether to conduct a single HF evaluation or multiple LF simulations, considering comparable computational costs during the update process. These methods exemplify how fidelity management strategies in correction-based MFBO focus on utilizing model accuracy to guide the updating of the MF model with new samples evaluated at varying fidelity levels.

\subsubsection{AR1-based MFBO}
\label{sec:ar1_mfbo}

AR1-based MFBO methods commonly use the surrogate model directly~\cite{HuangGZ13} or employ the expected improvement (EI) as the infill criterion, performing evaluations at both low- and high-fidelity levels~\cite{ForresterSK07,YongWTKS19,HanXLYKS20}. A limitation of EI is its focus solely on the uncertainty within the HF model. To overcome this, Zhang et al.~\cite{ZhangHZ18} introduced an uncertainty quantification technique that integrates spatial location and fidelity level, enabling EI to select samples from any fidelity level. In a similar vein, improved EI strategies were proposed in~\cite{HuangANM06,ReisenthelT14} that consider the correlation and cost ratio between different fidelity levels, achieving comparable effectiveness. Furthermore, Fiore et al.~\cite{FioreM22} developed a two-step lookahead multi-fidelity acquisition function for EI, aimed at maximizing long-term benefits in the optimization process. Fidelity management strategies in AR1-based MFBO typically focus on incorporating LF information into the EI function, thereby facilitating sampling across various fidelity levels.

\subsubsection{MTGP-based MFBO}
\label{sec:mtgp_mfbo}

In MTGP-based MFBO, significant efforts have been dedicated in developing efficient acquisition functions that can acquire new solutions across various fidelity levels. These efforts have led to the adaptation of traditional acquisition functions like entropy search (ES)~\cite{SwerskySA13,ZhangHLK17}, max-value ES (MES)~\cite{MossLGR21,TakenoFTKSTK20,MossLR20,LiKZ21,LiXKZ20}, and EI~\cite{LuongNGRV21,LeePAS20} to accommodate multi-fidelity data. For instance, Swersky et al.~\cite{SwerskySA13} introduced a cost-sensitive ES, normalizing information gain from uncertainty reduction against query cost. However, the scalability of this method is constrained by the requirement for sampling-based approximations of the optimal point distribution, which becomes increasingly complex with a higher dimensionality.

Takeno et al.~\cite{TakenoFTKSTK20} developed a multi-fidelity MES that requires only one-dimensional numerical integration, with a parallelization scheme suitable for asynchronous optimization settings. This approach was later improved to support synchronous parallelization and trace-aware querying~\cite{TakenoFTKSTK22}. Folch et al.~\cite{FolchLSWTWM23} expanded upon MES for asynchronous batch optimization, incorporating local penalization and Thompson sampling techniques~\cite{GonzalezDHL16,KandasamyKSP18}. Additionally, Moss et al.~\cite{MossLR20} adapted the multi-fidelity MES algorithm for both continuous and discrete fidelity levels. Later, Moss et al.~\cite{MossLGR21} proposed a computationally efficient, information-theoretic acquisition function for synchronous batch optimization, leveraging a lower bound of approximate information gain. Zhang et al.~\cite{ZhangHLK17} employed a convolved MTGP for joint modeling across fidelities, effectively sampling the HF optimum and utilizing MTGP's inherent properties for multi-fidelity surrogate modeling. These illustrate how fidelity management in MTGP-based MFBO primarily focuses on integrating cost considerations into the information gain, thereby balancing exploration across different fidelity levels.

\subsubsection{Nonlinear hierarchical model-based MFBO}
\label{sec:hierachical_mfbo}

Like its MTGP-based counterpart, this method focuses on designing efficient acquisition functions and developing intricate models to capture complex correlations across fidelity levels. Li et al.~\cite{LiXKZ20} demonstrated this by employing a DNN to represent the relationships between successive fidelities. They also implemented the multi-fidelity MES acquisition function~\cite{TakenoFTKSTK20} for jointly selecting the fidelity level and input location of new solutions. To overcome the analytical challenges posed by nonlinearly coupled outputs at different fidelities, they approximated each posterior distribution as Gaussian using fidelity-wise moment matching and Gauss-Hermite quadrature. In a further work~\cite{LiKZ21}, they utilized deep autoregressive networks to capture intricate relationships across all fidelity levels. They introduced a batch acquisition function based on MES to enable parallel queries, enhancing query quality and diversity. This function specifically penalizes highly correlated queries to encourage diversity. In a different vein, Savage et al.~\cite{SavageBMR22} used DGPs~\cite{CutajarPDLG19} to model multi-fidelity coiled-tube reactor simulations, employing an UCB function for the HF model to sample new solutions. The choice of fidelity in their method was based on the variance of the DGP. Thus, nonlinear hierarchical model-based MFBO shares fidelity management strategies with MTGP-based MFBO, emphasizing the integration of intricate modeling techniques and efficient acquisition functions.

\subsection{Surrogate-Assisted Evolutionary Algorithms for Multi-Fidelity Optimization}
\label{sec:mfsaea}

In contrast to BO, SAEAs are characterized by their population-based search dynamics. The existing SAEAs for MFO typically leverage either single model or correction-based methods for surrogate modeling.

In the single-model approach, LF information aid in identifying high-potential HF solutions. This involves an initial exploration in the LF space over several iterations, followed by a focus on HF exploration using local optima~\cite{YiSS20} or superior solutions from each cluster~\cite{LiuKZ16}. Although effective, this two-stage strategy may not fully exploit the dynamic nature of MFO, particularly regarding the fluid transition between fidelity levels during optimization.

Correction-based methods, on the other hand, enable a gradual exploration of the multi-fidelity model by the population, effectively locating optimal solutions over time. For instance, Kriging is used in~\cite{LiuC14} to create a multi-fidelity surrogate model, where the population initially explores the LF space before transitioning to the multi-fidelity surrogate model, with non-dominated solutions being chosen iteratively for HF evaluations. This approach is echoed in~\cite{ZhuWC14} and~\cite{ZhouWXJ21}. Li et al.~\cite{LiTLSW21} developed a correlation-based multi-fidelity surrogate model using GP, where the population explores the multi-fidelity surrogate model and top and middle-tier individuals are selected for low- and high-fidelity evaluations, respectively. Similarly, Yi et al.~\cite{YiGLSL19} utilized RBF NNs to construct a correction-based multi-fidelity surrogate model, evaluating non-dominated solutions initially at LF, followed by HF updates. Sun et al.~\cite{SunLL12} used a colony algorithm on a correlation model, with optimal solutions evaluated at both low- and high-fidelity levels.

Wang et al.~\cite{WangJYJ20} proposed a different approach, employing a multi-fidelity surrogate model comprising an ensemble model trained on LF data and RBF NNs on HF data. They identify promising and uncertain solutions from the population for HF evaluations in each iteration. Similarly, Kenny et al.~\cite{KennyRS23} formulated a method using current sample information to define a sampling neighborhood for LF evaluations, which are then used to construct a co-Kriging surrogate model. This model undergoes a global search to identify candidates for HF evaluation.

\subsection{Bandit-based Algorithms for Multi-Fidelity Optimization}
\label{sec:bandit}

Another line of research builds upon the concept of bandit algorithms~\cite{LattimoreS20}, dynamically orchestrating exploration across varying fidelity levels on the fly. These methods, predominantly deployed within the domain of hyperparameter optimization, strategically balance between the volume of configurations and the resources allocated to each fidelity.

As the pioneering work in this field, Jamieson et al.~\cite{JamiesonT16} reframed the hyperparameter optimization problem as an instance of non-stochastic best-arm identification problem, utilizing the successive halving (SH) algorithm~\cite{KarninKS13} as an effective tool for resource allocation. The SH algorithm favours promising solutions with larger resource allocation. Each iteration starts with a uniform sampling within the search space, evaluated at the lowest fidelity level. The top-performing half advances to the subsequent fidelity level, a process that continues until the highest fidelity level is achieved. This process, known as an SH bracket, repeats until a termination criterion is met. However, the SH algorithm can prematurely discard solutions that perform poorly at low fidelities but may excel at high fidelities.

To address this aforementioned challenge, Li et al.~\cite{LiJDRT17} proposed the seminal Hyperband method. It integrates different early stopping settings across multiple SH brackets, each starting at different fidelities with varying solution numbers. However, since the Hyperband samples solutions randomly, it does not incorporate information from previous samples~\cite{FalknerKH18}. To improve upon this, Falkner et al.~\cite{FalknerKH18} developed BOHB that utilizes the tree Parzen estimator~\cite{BergstraBBK11} to construct individual surrogate models for each fidelity level, similar to the approach in~\cite{WangXW18}. BOHB, while effective, struggles with discrete dimensions  and is not well scalable to high-dimensional problems. To address these limitations, Awad et al.~\cite{AwadMH21} enhanced BOHB with differential evolution for candidate sampling. Additionally, Li et al.~\cite{LiSJG0021} developed a multi-fidelity ensemble model that integrates information from all fidelity levels to approximate the highest fidelity.

Further, Zhu et al.~\cite{ZhuLGBL22} proposed an LCM model across all fidelity levels for guided sampling. In asynchronous MFO, such as ASHA~\cite{LiJRGBHRT20} introduced an asynchronous evaluation paradigm based on SH. Extending this, Bohdal et al.~\cite{BohdalBWEAZ23} proposed PASHA, which starts with minimal resource allocation and increases as needed. Last but not the least, Li et al.~\cite{LiSJZLLZC22} developed Hyper-Tune, an asynchronous multi-fidelity optimizer within the MFES-HB framework.

\subsection{Discussions}
\label{sec:discussion_optimizer}

\begin{figure}[t!]
    \centering
    \includegraphics[width=\linewidth]{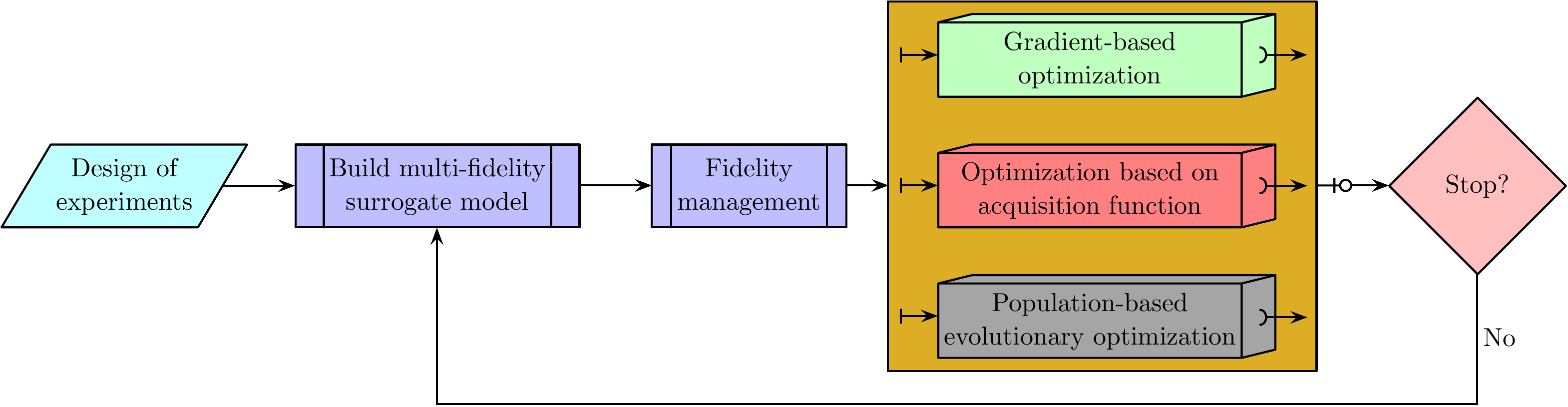}
    \caption{Schematic workflow of three multi-fidelity optimizer---traditional gradient-based optimization algorithms, Bayesian optimization and bandit-based algorithms, surrogate-assisted evolutionary algorithms.}
    \label{fig:optimizer}
\end{figure}

In the following discussion, we provide a high-level summary of the optimizers examined in this section.

\begin{itemize}
    \item As summarized in~\pref{fig:optimizer}, traditional gradient-based optimization, BO and bandit optimization, and SAEAs share common grounds in MFO. The key difference lies in the search for the next point(s) of merit for conducting function evaluation(s) and update the multi-fidelity surrogate models, as the yellow box highlighted in~\pref{fig:optimizer}.

    \item Despite significant advancements in MFBO, several challenges remain open for further exploration. These include accurately characterizing and modeling inter-fidelity relationships, effectively managing information from different fidelity levels, and balancing exploration with exploitation. Additionally, there is an urgent need for comprehensive benchmarking studies to evaluate, compare, and validate different algorithms. Such studies could provide valuable insights for practical applications, and offer a more definitive basis for assessments.

    \item Another imperative consideration is the development of robust theoretical foundations for MFBO. In conventional BO, the objective function is typically a scalar function drawn from a GP, facilitating rigorous mathematical analysis, including convergence properties. However, these analyses do not directly apply to the multi-fidelity context, where managing multiple interrelated objective functions of varying fidelity levels introduces extra complexity. Challenges include accurately modeling these interrelationships, determining the optimal fidelity level for evaluation, and confirming the convergence of such complex systems.

    \item Although SAEAs share common grounds with BO, they bring unique opportunities and challenges for future exploration, primarily due to their inherent population-based characteristics. In particular, we underscore the necessity of self-adaptive fidelity management strategies at the individual solution level. This is akin to the maintenance of diversity in EAs, which opens a new dimension of complexity and potential in SAEAs for MFO.
\end{itemize}


\section{Fidelity Management Strategies}
\label{sec:fidelity_management}

Contrasting with traditional optimization, which typically operates within a single fidelity level, MFO introduces an additional layer of complexity by requiring exploration across diverse fidelities throughout the optimization process. This shift elevates the importance of developing effective fidelity management strategies that can harmonize the search across multiple fidelity levels. In this context, existing methodologies summarized from our topic modeling results in~\pref{fig:mining_model_management}(b) predominantly fall into two categories---\textit{fixed} and \textit{adaptive} strategies.

\subsection{Fixed methods}
\label{sec:fixed_methods}

Fixed fidelity management strategies, as the name suggests, transition between different fidelity levels following a predetermined pattern. Typically, the optimization processes of these strategies start with a LF level and gradually progress to higher fidelity levels~\cite{LiuC14,YiSS20}. For instance, Li et al.~\cite{LiHLZZ20} introduced an approach that uses incremental curves to systematically increase fidelity levels throughout the optimization, thereby reducing the computational time compared to the methods that do not utilize LF information. Within a genetic algorithm framework, El-Beltagy et al.~\cite{ElBeltagyAA99} experimented with three fidelity selection techniques and observed that a sequential fidelity selection strategy yielded the best results. Another alternative fixed strategy involves transitioning between fidelity levels at predetermined intervals. In this vein, Zhu et al.~\cite{ZhuWC14} utilized fitness values corrected by a Kriging model as LF values within a genetic algorithm, with a selected subset of non-dominated solutions undergoing HF evaluations. In multi-objective optimization, the variation of Hypervolume between adjacent generations~\cite{LiuWY22,JiangZ0C19}, and the accuracy of the LF model~\cite{MamunSR22} have also been used to control the transition between different fidelities. Additionally, some methods also suggested using low- and high-fidelities simultaneously during the optimization process (e.g.,~\cite{YangHW22,YangW23,LiuXZ20,ShuJZSHM18}).

However, fixed fidelity management strategies face significant challenges. Their primary limitation is the difficulty in establishing a universally effective strategy, especially for black-box optimization problems. The inherent rigidity of these strategies, and their lack of adaptability to the search process's evolving dynamics, is a major drawback. Without specific knowledge of the problem landscape, optimally allocating resources and determining the most efficient frequency for transitioning between fidelity levels become challenging. This inflexibility and inability to adapt to changing search dynamics could potentially compromise the efficiency and overall performance.

\subsection{Adaptive methods}
\label{sec:adaptive_methods}

Unlike the fixed methods, adaptive methods dynamically adjust the fidelity level during the optimization process based on various indicators. One commonly used indicator is the uncertainty estimated by the surrogate model. For instance, Kandasamy et al.~\cite{KandasamyDOSP19} introduced a strategy that initially leverages lower evaluation costs at the lower fidelities. It escalates to higher-fidelity evaluations only when the uncertainty of a solution at the lowest fidelity exceeds a certain threshold. However, this method has a potential drawback---it may misallocate resources at the LF level if the promising regions in the LF landscape do not correspond with those in the HF landscape.

Another valuable indicator in adaptive methods is the correlation between low- and high-fidelity evaluations. Lim et al.~\cite{LimOJS08} developed a method that conducts a local search at the LF level to reduce the number of HF evaluations. The fidelity level for this local search is determined based on the correlation between low- and high-fidelity levels, assessed using neighboring training data. In a similar vein, Habib et al.~\cite{HabibSR19} proposed using Kendall's rank correlation coefficient to gauge this correlation. Other approaches focus on evaluating the consistency between low- and high-fidelity solution rankings as a means to estimate their correlation~\cite{BrysonR16,WangJD18}.

Furthermore, many approaches incorporate the cost associated with each fidelity level into the information gain derived from acquisition functions, such as ES~\cite{SwerskySA13,ZhangHLK17}, MES~\cite{MossLGR21,TakenoFTKSTK20,MossLR20,LiKZ21,LiXKZ20}, and EI~\cite{LuongNGRV21,LeePAS20}. These methods typically prioritize less expensive LF solutions over their costlier HF counterparts. By doing so, they effectively identify the most cost-efficient fidelity level to use throughout the optimization process.

\subsection{Discussions}
\label{sec:discussion_fidelity_management}

Here we provide a summative discussion on the fidelity management strategies reviewed in this section.

\begin{itemize}
    \item As the pie charts shown in~\pref{fig:mining_model_management}(b), the nine fidelity management strategies demonstrate varying degrees of flexibility. Except the approaches which do not use any dedicated fidelity management strategies, the method that integrates information gain with the associated cost of each fidelity level is the most commonly investigated approach in the literature. Table A$6$ of the \textsc{Appendix} summarizes the key characteristics and their underpinned disadvantages of the current fidelity management strategies.

    \item While adaptive methods offer a greater flexibility and the potential for better solutions, they also demand more computational resources to determine the optimal fidelity level at each iteration. This additional computation might offset the benefits of using multiple fidelities. In contrast, fixed methods, though potentially less effective in utilizing fidelity levels, can be more computationally efficient.

    \item The suitability of both fixed and adaptive methods often depends on the specific problem. For instance, a fixed method may be more appropriate for problems with consistently high correlation between low- and high-fidelity models, whereas adaptive methods may be advantageous when the correlation varies across the search space.

    \item It is important to note that, despite the increased flexibility of adaptive methods compared to fixed methods, they often require careful parameter tuning (e.g., the threshold for switching fidelities or the weight of cost in the acquisition function). Over-tuning based on a limited set of benchmark problems can lead to overfitting, which may limit their applicability to a broader range of problems.
\end{itemize}


\section{Benchmark Test Problems in multi-fidelity Optimization}
\label{sec:benchmark_problems}

Benchmark problems play an essential role in evaluating the performance of different algorithms, facilitating understanding of their strengths and weaknesses. However, benchmark problem design specific to MFO has received limited attention in the current literature. Generally speaking, the existing approach involves constructing a LF function $f_\mathrm{l}(\mathbf{x})$ as a transformation of the HF function $f_\mathrm{h}(\mathbf{x})$. This can be typically formulated as:
\begin{equation}
    f_\mathrm{l}(\mathbf{x})=a(\mathbf{x})g(f_\mathrm{h}(A\mathbf{x}+b))+\epsilon(\mathbf{x}),
    \label{eq:mf_problem}
\end{equation}
where $g(\cdot)$ represents a transformation function, $a(\mathbf{x})$ is a multiplicative factor, and $\epsilon(\mathbf{x})$ is an error term. 

A common method for designing multi-fidelity benchmark problems involves setting all components of~\eqref{eq:mf_problem} as constants, leading to a linear transformation~\cite{DurantinRDG17}. Other studies have adopted polynomial~\cite{LiWD18} or exponential functions~\cite{SePWQVD19} for $\epsilon(\mathbf{x})$. However, these approaches tend to be limited to fixed problem dimensions. To address this limitation, \cite{RumpfkelB20} and~\cite{BrysonR16} introduced scalable functions suitable for varied dimensions. Additionally, some research has focused on nonlinear transformations of HF problems, such as squaring the HF functions and multiplying them by polynomials~\cite{RijnS20,CutajarPDLG19}. However, these benchmark problems are typically formulated with only a single variable.

Predominant approaches in developing multi-fidelity benchmark suites involve using a collection of problems with diverse properties. For instance, Toal~\cite{Toal15} introduced four multi-fidelity benchmark test problems where the correlation between low- and high-fidelity models can be controlled by specific parameters. However, the dimensionality of these benchmark test problems is fixed. To address this problem, Mainini et al.~\cite{MaininiSRMQPYFPFBNDB22} proposed using six foundational functions featured by multi-modality, discontinuity, and unusual local behaviors to constitute the HF problems. Then, the LF problems are designed by adding noise and introducing linear or nonlinear discrepancies regarding the HF problems. While these LF problems are scalable in dimension, they lack controllable parameters to adjust the correlation between low- and high-fidelity levels, limiting their usefulness in benchmarking purposes. To tackle this issue, Wang et al.~\cite{WangJD18} proposed three distinct error functions to construct LF problems, each serving a specific purpose. In particular, the resolution error introduces divergences in global and local errors, the stochastic error causes fitness value variations for the same solution across different simulations, and the instability error results in failed simulations. However, since the low- and high-fidelity models within this test suite have a considerable correlation, their practicality is thus limited. In~\cite{RijnS20}, a suite of multi-fidelity benchmark problems from various sources, including a significant portion from Toal's work~\cite{Toal15}, are compiled into a Python package. Note that these aforementioned benchmark test problems are limited to single-objective optimization. To bridge this gap, more recently, Kenny et al.~\cite{KennyRSL23} developed a test suite for multi-objective, multi-fidelity optimization by combining existing unconstrained, multi-objective design optimization problems featured with resolution, stochastic, and instability errors, common in multi-fidelity simulations.

\subsection{Discussions}
\label{sec:discussion_benchmark}

Due to the stringent page limit, we leave more detailed information about the benchmark problems in Section B.$5$ and B.$6$ of the \textsc{Appendix}, including the mathematical formulations and $2$D contour plots. In summary, there are four takeaways learned from benchmark problems in this section.

\begin{itemize}
    \item Most existing multi-fidelity benchmark problems are limited to linear transformations, which restricts their applicability in evaluating algorithm performance on problems involving non-linear transformations, common in many real-world scenarios.
    \item The dimensionality of most current multi-fidelity benchmark problems is fixed. This limitation hampers the ability to assess how well algorithms scale across problems of varying dimensions.
    \item Many of the current multi-fidelity benchmark problems do not allow for modifications to fidelity correlations. This restricts its capability to investigate how varying LF functions impact algorithm performance.
    \item The characteristics of HF problems used to construct multi-fidelity benchmarks lack sufficient diversity. As a result, there is a concern for benchmarking the capability performance of existing MFO algorithms.
\end{itemize}


\section{Applications of multi-fidelity optimization}
\label{sec:applications}

Beyond the reviewed methodological advancements, MFO has profound implications in a diverse array of real-world applications. Based on the topic modeling analysis in~\pref{sec:analysis_topics}, this section delves into three prominent domains including hyperparameter optimization, engineering design and optimization, and scientific discovery. Additionally, given the emergence of PINNs, we also investigate the corresponding applications in various domains. In Fig. A$2$ of the \textsc{Appendix}, we use the mind map to visualize the taxonomy of the applications of MFO.

\subsection{Applications on Hyperparameter Optimization (HPO)}
\label{sec:application_hpo}

Machine learning technologies, particularly DNNs, have significantly transformed many aspects of modern society~\cite{LeCunBH15}. A crucial factor in the performance of many machine learning approaches is the selection of design choices, leading to renewed interest in HPO~\cite{FeurerH19}. A major challenge in contemporary HPO is the growing computational cost associated with evaluating a single hyperparameter configuration, due to increasing dataset sizes and model complexity. Multi-fidelity methods, which use LF approximations of the actual loss function to speed up the process while compromising approximation accuracy, have gained popularity in addressing this challenge~\cite{FeurerH19}. Predominantly, bandit-based algorithms and MFBO are the two primary methods employed in multi-fidelity HPO. In the following paragraphs, we will explore selected works in HPO, focusing on three key design components of MFO.

\subsubsection{Surrogate modeling in HPO}
\label{sec:hpo_surrogate}

Given the mixed variable characteristics of HPO, tree-based models, such as the tree Parzen estimator~\cite{FalknerKH18} and random forest~\cite{LiSJG0021,LiSJZLLZC22}, are popular choices for surrogate modeling, especially in bandit-based algorithms. For MFBO, researchers have focused on developing composite kernels to address the limitations of GP in modeling mixed variables~\cite{KleinFBHH17,KandasamyDSP17}. These methods jointly model both the parameters designated for fidelity settings (as introduced in~\pref{sec:hpo_fidelity_setting}) and the hyperparameters of the underlying machine learning algorithm.

\subsubsection{Fidelity settings in HPO}
\label{sec:hpo_fidelity_setting}

In HPO, various parameters can set LF approximations to expedite the performance evaluation of a hyperparameter configuration. Examples include the size of the training dataset, the number of iterations or epochs, the number of features, the number of folds for cross-validation, and even down-sampling images in computer vision. Most existing works use either the number of epochs (e.g.,~\cite{LiJDRT17} and~\cite{FalknerKH18}) or the size of the training dataset (e.g.,~\cite{KleinFBHH17} and~\cite{WangXW18}) as key parameters to set the fidelity levels. In bandit-based algorithms, the number of fidelity levels is determined by a reduction factor that controls the exponential reduction of computational resources~\cite{LiJDRT17}. Conversely, the fidelity levels in MFBO are often predetermined based on the prior experience of machine learning experts~\cite{KandasamyDOSP19}. There have also been some efforts to establish continuous fidelity levels~\cite{KleinFBHH17,KandasamyDSP17}.

\subsubsection{Fidelity management in HPO}
\label{sec:hpo_fidelity_management}

Most multi-fidelity HPO methods follow the fidelity management strategy of Hyperband, as introduced in~\pref{sec:bandit}. They use LF approximations to identify promising candidates for higher fidelities, gradually increasing fidelity levels until reaching the highest fidelity (e.g.,~\cite{BertrandAPB17,FalknerKH18,WangXW18,LiJRGBHRT20,AwadMH21}). Conversely, Bohdal et al.~\cite{BohdalBWEAZ23} proposed progressively increasing fidelity levels when the ranking of configurations within the top two fidelities remains unstable. Similarly, Li et al.~\cite{LiSJZLLZC22} considered both the consistency of ranks and the cost of LF in fidelity selection. Unlike bandit-based algorithms, MFBO methods often determine the fidelity level of a solution based on the corresponding estimated uncertainty (e.g.,~\cite{KleinFBHH17} and~\cite{WistubaKG22}).

\subsection{Applications on Engineering Design and Optimization}
\label{sec:application_engineering}

In the realm of engineering, optimization challenges are often characterized by their inherent complexity and the significant resources they demand, both economically and computationally. This challenge is particularly pronounced in sectors where precision and efficiency are paramount, and traditional optimization methods might struggle to meet the intensive computational requirements. In this context, MFO emerges as an invaluable tool. It empowers engineers to utilize LF models for rapid insights, reducing the reliance on more expensive HF simulations or experiments. By adeptly navigating through various levels of fidelity, MFO enables more efficient, cost-effective engineering design and optimization. In this subsection, we overview the application of MFO across seven key engineering sectors, focusing on the design objectives typically targeted in these sectors, the fidelity settings employed to streamline optimization processes, and the surrogate modeling methods integral to the success of MFO.

\subsubsection{Design objectives of MFO for engineering design and optimization}
\label{sec:engineering_objectives}

In this section, we briefly overview seven popular engineering sectors identified by the topic modeling analysis in~\pref{sec:analysis_topics}.

\begin{itemize}
\item\textit{Aerospace vehicles}: This sector focuses on optimizing airflow characteristics and design features, such as airfoils, wings, and body shapes. These optimization tasks, which can significantly improve performance and efficiency, often require intricate simulations. Specifically, in airfoil design, MFO algorithms target objectives like maximizing lift~\cite{LeifssonK10}, minimizing drag coefficient~\cite{BenamaraBL16,BarrettBK06,MarchW2012}, and optimizing lift-to-drag ratios~\cite{AyeWKTBYP23,DemangeSK16,TaoS19}. Some studies also explore optimizing multiple conflicting objectives, such as balancing aerodynamic drag with airfoil thickness at the trailing edge~\cite{AriyaritK17} or enhancing lift coefficient and endurance factor~\cite{PriyankaS21}. For wing design, the goals include maximizing lift coefficient~\cite{TaoACGPG19} and reducing wing drag~\cite{Elham15}. MFO in this domain extends to optimizing entire aircraft and flight vehicles, with objectives like drag minimization~\cite{BrahmacharyNS17,RajnarayanHK08,TancredR15} and gross weight optimization~\cite{NguyenTL15}.

    \item \textit{Shipbuilding}: Optimization challenges in shipbuilding, including turbine and ship hull design, often revolve around fluid dynamics interactions. Turbine blade design research, for instance, aims at optimizing velocity profiles~\cite{BahramiTDVG16}, blade length~\cite{BahramiTVVF14}, isentropic efficiency~\cite{GuoSPLH18}, emissions reduction~\cite{ToalZKLZ21}, and stress management~\cite{NacharBNR20}. Engine design optimization focuses on fuel consumption efficiency~\cite{ToalKBDYPRRK2014} and mass minimization~\cite{YongWTKS19}. Compressor design aims to enhance pressure ratios~\cite{JolyVP14} and adiabatic efficiency~\cite{MondalJS19}. Ship hull optimization targets mass reduction~\cite{TezzeleFSS23}, drag optimization~\cite{LiuWL23}, and displacement optimization for twin hulls~\cite{PellegriniSHD17}. Combustor design focuses on NOx emission reduction~\cite{WankhedeBK2013} and combustion performance enhancement~\cite{WankhedeBK11}. Symmetric diffusers aim for maximum pressure recovery~\cite{MadsenL01}, while rotor and glider shape optimization focus on efficiency~\cite{BuSHZ22} and lift-drag ratio improvement~\cite{ZhangZWZ21}, respectively. Cyclone separator research addresses gas cyclone pressure drop minimization for more efficient separation processes~\cite{SinghCEDD17}.

    \item\textit{Reservoir optimization}: This domain mainly focuses on two main areas---reservoir production optimization and well spacing optimization. The prior one aims to determine the most effective operational scheme to achieve a set of predefined objectives. Common objectives include optimizing production for improved oil and gas recovery, maximizing net present value for economic gains, and maintaining reservoir pressure to ensure continuous and stable production~\cite{WangYZAZL22a,WangYZAZL22b,WangYTAZL22c,CardosoD10}. For the well spacing optimization, research focuses on the optimal design of horizontal well spacing and fracture stage placement. The goal is to balance gas production with economic benefits, navigating the complex trade-offs inherent in these decisions~\cite{WangYWDZL22,zhaoYWAFW22}.

    \item\textit{Thermal management and optimization}: Research in this field is dedicated to a variety of applications, ranging from turbomachinery cooling to the thermal management of electronic devices and batteries. Turbomachinery cooling design, for instance, aims to optimize the placement of film cooling holes in gas turbine end walls to enhance cooling performance~\cite{BuYSL22}, designing film-cooling hole arrays on high-pressure turbine nozzles to reduce nozzle surface temperature~\cite{KimLY18}, and shaping film cooling holes in centrifugal compressors for improved cooling effectiveness~\cite{ZhangLCSY19}. In pump cooling subsystems, the focus here is on minimizing the temperature of heat sources and the mass of filled channels in these systems~\cite{HartlGMB16}. Optimization efforts in pump thermal transport systems target reducing thermal transport losses, minimizing driving current, and improving overall efficiency~\cite{HartlFB17}. For battery thermal management, the goal is to extend battery life and equalize temperature across cells~\cite{WangLSSZ18}. For electro-thermal optimization of multiport converters, the research involves optimizing static converter losses, reducing inductor weight, and controlling ripple current~\cite{TranCMVLBH19}.
    
    \item \textit{Electromagnetics}: In electromagnetic engineering, research spans from antenna design, focusing on minimizing reflection coefficients~\cite{JacobsKO13, LiuKA17}, to optimizing dielectric resonator antennas for efficient radiation and reception~\cite{KozielOL11}. The design of miniaturized couplers targets bandwidth maximization and size minimization~\cite{KozielB16}. Multi-level microwave design optimization, covering various components, employs automated model fidelity adjustments to enhance performance~\cite{KozielO14}.

    \item \textit{Manufacturing processes}: This domain aims to optimize production methods for cost and efficiency gains. In manufacturing, objectives include optimal heating in warm-forming processes for aluminum alloys~\cite{KimKN0}, parameter optimization in self-piercing riveting~\cite{LiLHCZHZ23}, and laser welding process optimization for depth penetration and bead width control~\cite{ZhouYJSCHGW17, YangGC18}. Additive manufacturing focuses on tensile strength optimization with varying infill topologies~\cite{ZhouHW19}, using surrogate-based multi-fidelity models to accelerate these optimization processes.

    \item\textit{Control optimization}: Research in this field spans several areas including solar-powered building systems, aircraft design, and vehicle control. For instance, \cite{SorourifarCP23} delves into the integrated design and control of solar-powered heating/cooling systems, specifically addressing uncertainties in the operational environment. In aircraft design optimization, the primary goals are to minimize weight and drag while adhering to stability and control requirements in both longitudinal and lateral directions~\cite{ReistZRPB19}. In vehicle optimization, a critical area of research is the near-optimal online control of parallel hybrid electric vehicle powertrains~\cite{AnselmaBRBE19}. In~\cite{KeskinPKW20}, pro-social control strategies for connected automated vehicles are developed to enhance both traffic efficiency and driving comfort. 
\end{itemize}

\subsubsection{Fidelity settings in engineering design and optimization}
\label{sec:engineering_fidelity}

The performance evaluation of the design objectives outlined in~\pref{sec:engineering_objectives} often relies on the approximation of complex partial differential equations by computational fluid dynamics (CFD) simulations. The intricacy of setting fidelity levels in engineering design and optimization lies in balancing the precision of these CFD simulations against the associated computational cost. In the following paragraphs, we outline nine commonly used fidelity setting strategies.

\begin{itemize}
    \item\textit{Simplified fluid dynamics}: For both aerodynamics and hydrodynamics, $2$D models are frequently used to capture essential flow characteristics, offering a significant reduction in computational load while preserving critical fluid dynamics features (e.g.,~\cite{AriyaritK17,PriyankaS21,BenamaraBL16,AyeWKTBYP23,DemangeSK16}).

    \item\textit{Coarser mesh}: Employing a coarser mesh in CFD simulations strikes a balance between computational speed and resolution. This approach is beneficial when detailed geometrical fidelity is less critical to the overall analysis (e.g.,~\cite{AlexandrovLGGN00,LeifssonK10,NagawkarRDLK21,MadsenL01,BuYSL22, KimLY18}).

    \item\textit{Reduced-order model}: In early design stages, where broad performance trends are more crucial than fine details, reduced-order models like the Euler flow equations, quasi-three-dimensional analysis~\cite{Elham15}, or small-disturbance equations~\cite{LeifssonK10} can be utilized to achieve considerable computational savings.

    \item\textit{Alternate fidelity levels in CFD}: In some scenarios, CFD simulations themselves serve as lower-fidelity approximations of real-world experiments, particularly in aerospace engineering, to balance realism with computational feasibility (e.g.,~\cite{GoldfeldVAK05,ECCOMAS,TaoS19}).

    \item\textit{Fidelity adjustment in electromagnetics}: In electromagnetics engineering, fidelity levels are often adjusted via mesh refinement and model simplification techniques. Strategies include reducing the number of frequency points or employing early stopping in simulations to streamline the process (e.g.,~\cite{KozielOL11,KozielO14,KozielB16,LiuKA17,JacobsKO13}).

    \item\textit{Manufacturing process optimization}: For manufacturing processes, fidelity levels are often set by early termination of simulations and contrasting simulation results with physical experiments, thereby facilitating a practical balance between theoretical models and real-world experiments (e.g.,~\cite{KimKN0,ZhouYJSCHGW17,YangGC18,ZhouHW19,LiLHCZHZ23}).
    
    \item\textit{Reservoir optimization}: In this domain, the LF model mainly includes reduced order model~\cite{CardosoD10} and streamline model~\cite{WangYZAZL22a,WangYZAZL22b,WangYTAZL22c}.
    
    \item\textit{Thermal management and optimization}: For thermal management, LF models are developed using several methods, such as data from existing studies~\cite{ZhangLCSY19}, coarse mesh techniques~\cite{BuYSL22,KimLY18}, mathematical approximations~\cite{HartlGMB16,HartlFB17}, reduced order models~\cite{CardosoD10}, and simplified physical models~\cite{YajiYF22,TranCMVLBH19,WangLSSZ18}.
    
    \item\textit{Control optimization}: In control engineering, LF models can be developed as simplified decision rules from simplified model predictive control~\cite{SorourifarCP23}, numerical simulations~\cite{AnselmaBRBE19}, reduced-order models~\cite{MannarinoM14}, and coarse mesh techniques~\cite{ReistZRPB19}.
\end{itemize}

\subsubsection{Surrogate modeling in engineering design and optimization}
\label{sec:modeling_engineering}

\begin{figure}[t!]
    \centering
    \includegraphics[width=0.5\linewidth]{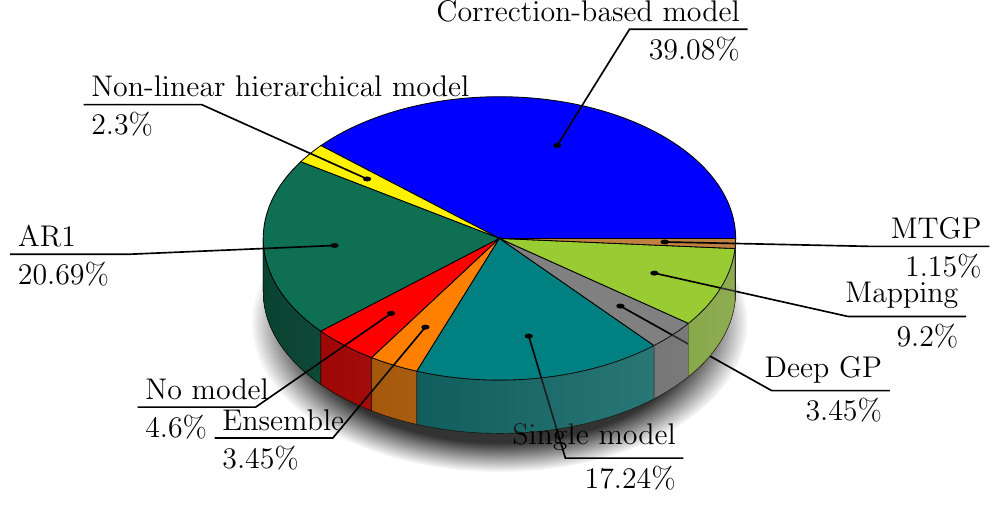}
    \caption{Pie chart of the proportion of a number of publications about different surrogate modeling methods in engineering design and optimization.}
    \label{fig:engineering_pie_chart}
\end{figure}

Given the continuous property of engineering design and optimization scenarios, all surrogate modeling approaches in~\pref{sec:surrogate_models} can serve our purpose. \pref{fig:engineering_pie_chart} shows a breakdown of different multi-fidelity surrogate modeling approaches used in engineering design and optimization, where correction-based methods and AR1 models emerge as the predominant techniques. While most studies typically employ two levels of fidelity, there has been a growing interest in exploring models with more than two fidelity levels, as evidenced in works like~\cite{JacobsKO13,KozielO14,NguyenTL15}. The use of multiple fidelities allows for a more nuanced approach to surrogate modeling, where each additional fidelity level can provide deeper insights or more accurate predictions, contributing significantly to the optimization process.

\subsection{Applications in Scientific Discovery}
\label{sec:scientific_application}

Scientific discovery is an emerging area that often faces challenges such as high computational costs, the need for accurate predictions from limited experimental data, and the complexity of modeling intricate in-vivo systems. MFO has emerged as a powerful approach to address these challenges, leveraging a range of data sources with varying levels of accuracy and computational expense. According to the topic modeling analysis in~\pref{sec:analysis_topics}, here we mainly overview relevant developments in material science and drug discovery.

\subsubsection{Applications of MFO in material science}
\label{sec:material_application}

In the vibrant field of materials science and engineering, MFO has shown its transformative potential in advancing our understanding and capabilities in material design and optimization. A diverse array of research endeavors leverages MFO to refine various aspects of materials and their properties, demonstrating its wide-ranging applications and impact.

One notable area is the optimization of architected materials, specifically designed for optimal energy absorption during compression. In~\cite{MoPR23}, MFBO has shown its potential to enhance material resilience. In parallel, efforts have also concentrated on reducing the weight of laminated conical shells~\cite{GoldfeldVAK05}, optimizing their buckling behavior~\cite{PilaniaGL17}, and increasing the strength-normalized strain hardening rate of dual-phase steel~\cite{GhoreishiMSAA18}. These studies not only showcase the usefulness of MFO in optimizing mechanical properties but also highlight the intricate interplay between material composition, processing parameters, and performance outcomes, as discussed in~\cite{KhatamsazMCJAAS21}.

In addition to these applications, MFO has been instrumental in the optimization of aircraft structural components, focusing on minimizing tension and compression stresses~\cite{VianaSBL09}, and in the development of variable-stiffness composite laminates tailored for improved buckling behavior~\cite{AnYK23}. The optimization of Young's modulus in PVA-treated buckypaper by Pourhabib et al.~\cite{PourhabibHWZWD15} further underscores the versatility of MFO in addressing diverse material properties. Further, the innovative design of metamaterial vibration isolators with honeycomb structures in~\cite{QianCZZZ21} demonstrated the application of MFO for minimizing natural frequency differences, a key factor in material stability and performance.

Similar to engineering design and optimization, the success of MFO in material science also hinges on solving complex PDEs to accurately determine material performance. The fidelity settings in these studies reveal a spectrum of approaches tailored to specific material science challenges. Commonly used methods include the use of coarse mesh for computational efficiency~\cite{VitaliHS02,GuoHWHX20,QianCZZZ21,AnYK23} and the strategy of early stopping to balance accuracy with computational budgets~\cite{NacharBNR20}. Further, there have been attempts on model simplification through the selection of variables versus constant material properties~\cite{GoldfeldVAK05} and the exploration of different functionals, such as Heyd-Scuseria-Ernzerhof and Perdew-Burke-Ernzerhoff, to suit specific material characteristics~\cite{PilaniaGL17}. Additionally, reduced-order models~\cite{GhoreishiMSAA18,KhatamsazMCJAAS21,KhatamsazMCJASA21} and the consideration of trade-offs between linear static and non-linear finite element analysis methods~\cite{VianaSBL09} reflect a nuanced understanding of balancing computational demands against accurate modeling.

\subsubsection{Applications of MFO in drug discovery}
\label{sec:drug_application}

The development of new drug compounds is often constrained by the high costs and time demands associated with laboratory synthesis and characterization, necessitating accurate predictions of molecular properties. MFO offers a promising solution by efficiently incorporating multiple approximations of the target function, thereby reducing costs and mitigating risks associated with physical experiments~\cite{FareFBVP22}. Density functional theory (DFT), for instance, provides a computational approach to approximate quantum mechanics simulations across a broad range of compounds. Despite its computational intensity, DFT offers a viable alternative to costly and potentially hazardous laboratory experiments~\cite{Lewis10}. In~\cite{Greenman21}, experiments showed the successful integration of computational and experimental data in predicting the optical spectra of molecules. Additionally, to reduce the costs associated with DFT, a multi-fidelity approach was applied to bridge the gap between LF crystal structure-lattice energy pairs data and the expensive hybrid functional DFT calculations in crystal structure prediction~\cite{EgorovaHWD20}.

To tackle the scarcity of accurate pKa data, \cite{MayrWWL22} and \cite{WuWWZCHH22} proposed using transfer learning approaches to augment and refine predictions based on experimental pKa. Similar methods were employed for predicting $\log{D_{7.4}}$~\cite{DuanFZLHLLDHHC23,WangXXZCRNTQZWCLZ23}. Such approaches illustrate how MFO can bridge the gap between limited high-quality data and more abundant but less accurate simulation data. In high-throughput screening, Buterez et al.~\cite{ButerezJKL23} highlighted the benefits of multi-fidelity data, which includes both primary, noisy screening data for a large number of compounds and higher-quality confirmatory data for a smaller subset. Their work underscores the value of MFO in enhancing drug performance prediction by effectively integrating data across fidelity levels. Additionally, \cite{HernSJLB23} adopted a multi-fidelity approach, which optimizes both electron affinity and ionization potential, to identify effective molecules. This research exemplifies the innovative application of MFO in navigating the complex chemical space for drug discovery. Likewise, in molecular dynamics, cost-effective surrogate models were employed to approximate physical properties, thus accelerating the MFO of force field parameters~\cite{MadinS23}. This method facilitates rapid assessments and streamlines the search within parameter space.

\subsection{Applications of Multi-Fidelity Physics-Informed Neural Networks}
\label{sec:pinn}

Last but not the least, multi-fidelity PINNs reviewed in~\pref{sec:mf_pinn} have been widely application in complex inverse problems across diverse domains. For instance, in the realm of environmental and geophysical sciences, multi-fidelity PINNs have been applied to model critical processes like unsaturated flows and reactive transport in porous media, using numerical simulations for both low- and high-fidelity data~\cite{MengK20} . In~\cite{JagtapMK22}, both experimental and synthetic data were utilized in PINNs to solve the inverse water wave problem, demonstrating their potential in designing structures like oil platforms and wind turbines. In the medical field, \cite{BasirS21} used multi-fidelity data to identify tumor cells beyond the threshold visible to magnetic resonance images, demonstrating the ability of PINNs to handle both low- and high-fidelity medical imaging data. In~\cite{RegazzoniPCLQ21}, multi-fidelity PINNs were applied in cardiac cell ionic dynamics that optimize parameters of a nonlinear model with various simulation times controlling data fidelity. In thermal sciences, \cite{BasirS21} investigated boundary heat flux identification in inverse heat conduction problems, underscoring the importance of multi-fidelity data in learning nonlinear unsaturated flows. Ramezankhani et al.~\cite{RamezankhaniNNV22} proposed exploring heat transfer in composite materials, with their multi-fidelity framework outperforming traditional models in predictive performance. In computational chemistry, multi-fidelity PINNs were applied to predict molecular dynamics properties, reducing the reliance on extensive simulation runs~\cite{IslamTMN21}. This work effectively bridges the gap between LF data from larger time-step simulations and HF results from finer simulations. In~\cite{LeiLW22,LeiW23}, multi-fidelity approaches were applied to tackle the accuracy and noise challenges in electrical capacitance tomography.


\section{Challenges and Future Opportunities}
\label{sec:challenges}

While MFO has been extensively studied across different disciplines in the past decades, fundamental challenges still remain. In addition to the discussion at the end of Sections~\ref{sec:surrogate_models} to~\ref{sec:benchmark_problems}, we outline five promising future research directions.

\subsection{Scalability of MFO}
\label{sec:challenge_scalability}

Real-world optimization usually involves a large number of decision variables (e.g., the control paramaters in aerodynamic shape design~\cite{GiuntaBKBGHMW97,ThelenBSB22} and the combinatorial space of molecule design~\cite{HernSJLB23}), along with a multitude of functional and/or non-functional performance attributes to accommodate a range of stakeholders' requirements. The curse of dimensionality has always been the Achilles’ heel of machine learning and optimization. In the context of MFO, scalability transcends the well-known constraints of surrogate models like Kriging and GPs in high-dimensional space~\cite{LiuCOW19,PadidarZHGB21}. One critical aspect is computational efficiency and resource constraints. As the dimensionality of the optimization problem increases, the computational resources required for accurate model predictions and algorithm convergence escalate significantly, often outpacing available resources. This is exacerbated in MFO, where different fidelity levels demand varied computational investments. Another challenge is data sparsity in high-dimensional spaces, where limited data samples become insufficient for surrogate models to capture the complex landscape of the optimization problem effectively. Furthermore, integrating multi-objective optimization in MFO introduces an additional layer of complexity. This involves navigating through conflicting objectives, often stemming from diverse stakeholder requirements, and necessitates sophisticated trade-off strategies to balance these competing goals. Lastly, the dynamic nature of real-world environments demands MFO algorithms to be both robust and flexible, capable of adapting to evolving optimization parameters or constraints. Addressing these multifaceted scalability challenges is critical for the advancement and practical application of MFO in complex, high-dimensional problem-solving. It is also worth noting that evolutionary computation~\cite{LiFK11,LiKWTM13,SunL20} and multi-objective optimization~\cite{LiZLZL09,LiKWCR12,LiZKLW14,LiFKZ14,LiKD15,LiDZK15,LiKZD15,WuKZLWL15,WuKJLZ17,WuLKZZ17,LiCMY18,WuLKZZ19,WuLKZ20} have been successfully applied to solve real-world problems, e.g., natural language processing~\cite{YangL23}, neural architecture search~\cite{ChenL23,LyuYWHL23,LyuHYCYLWH23,LyuLHWYL23}, water engineering~\cite{TangLDWZF20}, robustness of neural networks~\cite{ZhouLM22a,ZhouLM22b,WilliamsLM23a,WilliamsLM23b,WilliamsLM22,WilliamsL23c}, software engineering~\cite{LiXT19,LiXCWT20,LiuLC20,LiXCT20,LiYV23}, smart grid management~\cite{XuLAZ21,XuLA21,XuLA22}, communication networks~\cite{BillingsleyLMMG19,BillingsleyLMMG20,BillingsleyMLMG20,BillingsleyLMMG21}, machine learning~\cite{CaoKWL12,LiWKC13,LiK14,CaoKWLLK15,WangYLK21}, and visualization~\cite{GaoNL19}.

\subsection{Balancing Cost and Accuracy in Lower Fidelity Models}
\label{sec:challenge_fidelity}

In MFO, LF models are essential for reducing computational time and resource consumption. However, the main challenge lies in ensuring that these models are sufficiently accurate to guide the optimization process towards the global optimum. If the fidelity is too low, the model may lead to misleading optimization directions, causing inefficiencies or suboptimal solutions. On the other hand, a model with slightly higher fidelity might offer better guidance but at the cost of increased computational resources. The challenge is to develop methodologies or criteria for selecting and calibrating these lower fidelity models in a way that they provide the most beneficial trade-off between computational efficiency and accuracy. This involves researching and implementing methods to assess the predictive quality of lower fidelity models and their alignment with higher fidelity models. It may also include exploring adaptive strategies where the fidelity level is dynamically adjusted based on the current state of the optimization process. Addressing this challenge is crucial for enhancing the efficiency and effectiveness of MFO in practical applications, especially in scenarios with limited computational resources or time constraints.

\subsection{Human-in-the-loop Optimization}
\label{sec:challenge_human}

The ultimate goal of optimization is to assist human decision-makers in identifying solution(s), potentially reconciling multiple conflicting objectives according to their preferences. This is particularly important in fields like architectural design, fashion, and product development, where human intuition and creativity are indispensable. While the importance and efficacy of incorporating the decision-makers' preferences into the evolutionary multi-objective optimization process have been empirically validated in~\cite{LiDY18,LiLDMY20,LiNGY22,TanabeL23}, involving human in the loop has yet been studied in the MFO literature. Incorporating human expertise into multi-fidelity optimization presents distinct challenges. A primary challenge is the interpretation of different fidelity levels, where experts must understand varying approximations and simulations with differing accuracy and computational costs. This complexity is further compounded by the need to optimize human-algorithm interaction, necessitating an interface that allows effective decision-making based on multi-fidelity models. Additionally, managing uncertainty across these fidelity levels is crucial. Experts are tasked with balancing the use of LF models, which offer lower computational cost but higher uncertainty, against HF models that are more accurate but computationally intensive. Developing a human-computer interaction platform and mechanism is essential for enhancing the effectiveness of interactive MORL. Furthermore, we assume that the DM's preferences remain consistent throughout the MORL process. Proactively detecting and adapting to changes in the DM's preferences in dynamic and uncertain environments pose a significant challenge~\cite{ChenLY18,Li19,LiCSY19,FanLT20,ChenL21a,LiCY23,LiLL23}. Another missing, yet important issue, is how to handle constraints in the context of multiple objectives~\cite{LiCFY19,ShanL21}. Things become even more challenging when the constraints are (partially) unobservable~\cite{LiLL22,WangL24}. These challenges underscore the intricate nature of human-in-the-loop of MFO, requiring a sophisticated understanding of the interplay between various fidelity levels, algorithmic processes, and the inherent uncertainties within these systems.

\subsection{Near Real-World Benchmarks}
\label{sec:challenge_benchmarks}

As discussed in~\pref{sec:benchmark_problems}, benchmark problems are not only important for benchmarking performance, but also facilitate the understanding of algorithmic bottlenecks. The synthetic benchmark test problems outlined in~\pref{sec:benchmark_problems}, while mathematically well-defined, often build on unrealistic assumptions and lack sufficient consideration of realistic user requirements, potentially leading to biased evaluations~\cite{IshibuchiSMN17}. While there have been some efforts in HPO (e.g.,~\cite{YingKCR0H19,DongY20,EggenspergerMMF21,PfistererSMBB22}), there is an urgent need establish a new standard in benchmark suite design that mirrors real-world challenges. This effort will involve developing cross-disciplinary benchmarks, incorporating realistic user requirements, and employing robust objective evaluation methods.

\subsection{Reproducible Research}
\label{sec:challenge_open}

Last but not the least, reproducibility has been the cornerstone of open science, yet is unfortunately under development in the current MFO literature. While MFO has a diverse range of applications across various disciplines, only $26$ out of $1,242$ articles examined in this survey provided their source code. Relevant information is summarized in Section B.$4$ of \textsc{Appendix}. Note that this not only includes the algorithms developed in the paper, but also the benchmark problems and environment used in the corresponding experiments, leading to a significant gap in reproducibility and open science. In parallel to benchmarks, although there has been an attempt in HPO~\cite{PfistererSMBB22}, there is an urgent need for open-source software package for the MFO community that allows for $1)$ a well-documented source code and user-friendly examples; $2)$ user-friendly interface and API for adaptation and extention; $3)$ a robust and objective benchmark environment for performance evaluation.

\section{Conlusion}
\label{sec:conclusion}

In this survey paper, we systematically explore the multifaceted realm of MFO. We commence with a text mining framework based on a pre-trained language model. Specifically, the outcomes of text mining constitute the foundation of our literature understanding, topic modeling, and the organization of this survey. In the main body of our survey, we delve into an in-depth analysis of the fundamental principles and methodologies of three key building blocks for an effective implementation of MFO---a spectrum of surrogate models, fidelity management strategies, and optimization techniques. Each section presents significant findings and insights, contributing to a comprehensive understanding of MFO's current state and potential. Furthermore, while MFO finds diverse applications across various disciplines, this survey particularly focuses on machine learning, engineering design and optimization, and scientific discovery. These areas exemplify the versatility and impact of MFO in addressing complex computational challenges. Finally, this survey sheds light on emerging challenges and opportunities in MFO. These range from scalability and the composition of lower fidelities to the integration of human-in-the-loop approaches at the algorithmic side. Additionally, we highlight issues related to benchmarking and the promotion of open science in the field of MFO. This survey aims to inspire further research and collaboration in MFO, paving the way for new innovations and advancements.

\section*{Acknowledgment}
This work was supported in part by the UKRI Future Leaders Fellowship under Grant MR/S017062/1 and MR/X011135/1; in part by NSFC under Grant 62376056 and 62076056; in part by the Royal Society under Grant IES/R2/212077; in part by the EPSRC under Grant 2404317; in part by the Kan Tong Po Fellowship (KTP\textbackslash R1\textbackslash 231017); and in part by the Amazon Research Award and Alan Turing Fellowship.

\bibliographystyle{IEEEtran}
\bibliography{IEEEabrv,mfo}

\begin{thebibliography}{100}
\providecommand{\url}[1]{#1}
\csname url@samestyle\endcsname
\providecommand{\newblock}{\relax}
\providecommand{\bibinfo}[2]{#2}
\providecommand{\BIBentrySTDinterwordspacing}{\spaceskip=0pt\relax}
\providecommand{\BIBentryALTinterwordstretchfactor}{4}
\providecommand{\BIBentryALTinterwordspacing}{\spaceskip=\fontdimen2\font plus
\BIBentryALTinterwordstretchfactor\fontdimen3\font minus
  \fontdimen4\font\relax}
\providecommand{\BIBforeignlanguage}[2]{{%
\expandafter\ifx\csname l@#1\endcsname\relax
\typeout{** WARNING: IEEEtran.bst: No hyphenation pattern has been}%
\typeout{** loaded for the language `#1'. Using the pattern for}%
\typeout{** the default language instead.}%
\else
\language=\csname l@#1\endcsname
\fi
#2}}
\providecommand{\BIBdecl}{\relax}
\BIBdecl

\bibitem{ShiLYWWL21}
R.~Shi, T.~Long, N.~Ye, Y.~Wu, Z.~Wei, and Z.~Liu, ``Metamodel-based
  multidisciplinary design optimization methods for aerospace system,''
  \emph{Astrodynamics}, vol.~5, pp. 185--215, 2021.

\bibitem{WaschlKSR14}
H.~Waschl, I.~Kolmanovsky, M.~Steinbuch, and L.~D. Re, \emph{Optimization and
  optimal control in automotive systems}.\hskip 1em plus 0.5em minus
  0.4em\relax Springer, 2014, vol. 455.

\bibitem{WuYJZZFG22}
P.~Wu, W.~Yuan, L.~Ji, L.~Zhou, Z.~Zhou, W.~Feng, and Y.~Guo, ``Missile
  aerodynamic shape optimization design using deep neural networks,''
  \emph{Aerosp. Sci. Technol.}, vol. 126, p. 107640, 2022.

\bibitem{PeherstorferWG18}
B.~Peherstorfer, K.~Willcox, and M.~D. Gunzburger, ``Survey of multifidelity
  methods in uncertainty propagation, inference, and optimization,''
  \emph{{SIAM} Rev.}, vol.~60, no.~3, pp. 550--591, 2018.

\bibitem{GodinoPKH16}
M.~G. Fern{\'a}ndez-Godino, C.~Park, N.~H. Kim, and R.~T. Haftka, ``Review of
  multi-fidelity models,'' \emph{AIAA J.}, vol.~57, pp. 351--400, 2016.

\bibitem{GisellePKH19}
M.~F.-G. Giselle, C.~Park, N.~H. Kim, and R.~T. Haftka, ``Issues in deciding
  whether to use multifidelity surrogates,'' \emph{AIAA J.}, vol.~57, no.~5,
  pp. 2039--2054, 2019.

\bibitem{ParkHH17}
C.~Park, R.~T. Haftka, and N.~H. Kim, ``Remarks on multi-fidelity surrogates,''
  \emph{Struct. Multidiscipl. Optim.}, vol.~55, no.~3, pp. 1029--1050, 2017.

\bibitem{BrevaultBH20}
L.~Brevault, M.~Balesdent, and A.~Hebbal, ``Overview of {Gaussian} process
  based multi-fidelity techniques with variable relationship between
  fidelities,'' \emph{Aerosp. Sci. Technol.}, vol. 107, p. 106339, 2020.

\bibitem{ZhouZHM22}
Q.~Zhou, M.~Zhao, J.~Hu, and M.~Ma, \emph{Multi-fidelity surrogates modeling,
  optimization and applications}.\hskip 1em plus 0.5em minus 0.4em\relax
  Singapore: Springer, 2022.

\bibitem{Gratiet13}
L.~L. Gratiet, ``Multi-fidelity {Gaussian} process regression for computer
  experiments,'' Ph.D. dissertation, Universit{\'e} Paris-Diderot-Paris VII,
  2013.

\bibitem{Robinson07}
T.~D. Robinson, ``Surrogate-based optimization using multifidelity models with
  variable parameterization,'' Ph.D. dissertation, Massachusetts Institute of
  Technology, 2007.

\bibitem{March12}
A.~I. March, ``Multidelity methods for multidisciplinary system design,'' Ph.D.
  dissertation, Massachusetts Institute of Technology, 2012.

\bibitem{Haas12}
A.~Haas, ``Multifidelity optimization for supersonic aircraft design,'' Ph.D.
  dissertation, Stanford University, 2012.

\bibitem{Demange18}
J.~Demange, ``Multifidelity multiobjective trust-region-based optimisation for
  high-lift devices.'' Ph.D. dissertation, Cranfield University, 2018.

\bibitem{Santos21}
M.~Santos, ``Multi-fidelity modeling for aerothermal analysis of deployable
  planetary entry technologies,'' Ph.D. dissertation, Missouri University of
  Science and Technology, 2021.

\bibitem{Partin22}
L.~Partin, ``Multitask and multifidelity convolutional encoder-decoder
  networks,'' Ph.D. dissertation, University of Notre Dame, 2022.

\bibitem{Farah23}
N.~G.~G. Farah, ``Multifidelity methods for simulation of wax deposition in
  single-phase and two-phase flow,'' Ph.D. dissertation, Imperial College
  London, 2023.

\bibitem{Masuda22}
C.~M.~F. Masuda, ``Multifidelity methods for uncertainty quantification in
  cardiovascular hemodynamics,'' Ph.D. dissertation, Stanford University, 2022.

\bibitem{Berci11}
M.~Berci, ``Multidisciplinary multifidelity optimisation of flexible wing
  aerofoils by passive adaptivity,'' Ph.D. dissertation, University of Leeds,
  2011.

\bibitem{Grootendorst22}
M.~Grootendorst, ``{BERTopic}: {Neural} topic modeling with a class-based
  {TF-IDF} procedure,'' \emph{arXiv preprint arXiv:2203.05794}, 2022.

\bibitem{ReimersG19}
N.~Reimers and I.~Gurevych, ``{Sentence-BERT}: Sentence embeddings using
  siamese {BERT}-networks,'' in \emph{Conference on Empirical Methods in
  Natural Language Processing}, 2019.

\bibitem{McInnesH18}
L.~McInnes and J.~Healy, ``{UMAP:} uniform manifold approximation and
  projection for dimension reduction,'' \emph{CoRR}, vol. abs/1802.03426, 2018.

\bibitem{McInnesHA17}
L.~McInnes, J.~Healy, and S.~Astels, ``{HDBSCAN}: {Hierarchical} density based
  clustering,'' \emph{J. Open Source Softw.}, vol.~2, no.~11, p. 205, 2017.

\bibitem{Joachims97}
T.~Joachims, ``A probabilistic analysis of the rocchio algorithm with {TFIDF}
  for text categorization,'' in \emph{ICML'97: Proc. of the Fourteenth
  International Conference on Machine Learning}.\hskip 1em plus 0.5em minus
  0.4em\relax Morgan Kaufmann, 1997, pp. 143--151.

\bibitem{CarbonellG98}
J.~G. Carbonell and J.~Goldstein, ``The use of mmr, diversity-based reranking
  for reordering documents and producing summaries,'' in \emph{{SIGIR}'98:
  Proc. of the 21st Annual International {ACM} {SIGIR} Conference on Research
  and Development in Information Retrieval}.\hskip 1em plus 0.5em minus
  0.4em\relax {ACM}, 1998, pp. 335--336.

\bibitem{Grootendorst20}
\BIBentryALTinterwordspacing
M.~Grootendorst, ``Keybert: Minimal keyword extraction with bert.'' 2020.
  [Online]. Available: \url{https://doi.org/10.5281/zenodo.4461265}
\BIBentrySTDinterwordspacing

\bibitem{LiJDRT17}
L.~Li, K.~G. Jamieson, G.~DeSalvo, A.~Rostamizadeh, and A.~Talwalkar,
  ``Hyperband: {A} novel bandit-based approach to hyperparameter
  optimization,'' \emph{J. Mach. Learn. Res.}, vol.~18, pp. 185:1--185:52,
  2017.

\bibitem{Garnett23}
R.~Garnett, \emph{{Bayesian Optimization}}.\hskip 1em plus 0.5em minus
  0.4em\relax Cambridge University Press, 2023.

\bibitem{HeZGJ23}
C.~He, Y.~Zhang, D.~Gong, and X.~Ji, ``A review of surrogate-assisted
  evolutionary algorithms for expensive optimization problems,'' \emph{Expert
  Syst. Appl.}, vol. 217, p. 119495, 2023.

\bibitem{GiuntaBKBGHMW97}
A.~A. Giunta, V.~Balabanov, M.~Kaufman, S.~Burgee, B.~Grossman, R.~T. Haftka,
  W.~H. Mason, and L.~T. Watson, ``Variable-complexity response surface design
  of an {HSCT} configuration,'' \emph{Multidisciplinary Design Optimization,
  Alexandrov NM and Hussaini MY eds}, pp. 53--69, 1997.

\bibitem{LiuKZ16}
B.~Liu, S.~Koziel, and Q.~Zhang, ``A multi-fidelity surrogate-model-assisted
  evolutionary algorithm for computationally expensive optimization problems,''
  \emph{J. Comput. Sci.}, vol.~12, pp. 28--37, 2016.

\bibitem{KandasamyDOSP19}
K.~Kandasamy, G.~Dasarathy, J.~B. Oliva, J.~G. Schneider, and B.~P{\'{o}}czos,
  ``Multi-fidelity {Gaussian} process bandit optimisation,'' \emph{J. Artif.
  Intell. Res.}, vol.~66, pp. 151--196, 2019.

\bibitem{FolchLSWTWM23}
J.~P. Folch, R.~M. Lee, B.~Shafei, D.~Walz, C.~Tsay, M.~van~der Wilk, and
  R.~Misener, ``Combining multi-fidelity modelling and asynchronous batch
  bayesian optimization,'' \emph{Comput. Chem. Eng.}, vol. 172, p. 108194,
  2023.

\bibitem{YiSS20a}
J.~Yi, Y.~Shen, and C.~AShoemaker, ``A multi-fidelity {RBF} surrogate-based
  optimization framework for computationally expensive multi-modal problems
  with application to capacity planning of manufacturing systems,''
  \emph{Struct. Multidiscip. Optim.}, vol.~62, no.~4, pp. 1787--1807, 2020.

\bibitem{KampolisZAG07}
I.~C. Kampolis, A.~S. Zymaris, V.~G. Asouti, and K.~C. Giannakoglou,
  ``Multilevel optimization strategies based on metamodel-assisted evolutionary
  algorithms, for computationally expensive problems,'' in \emph{CEC'07: Proc.
  of the {IEEE} Congress on Evolutionary Computation}, 2007, pp. 4116--4123.

\bibitem{ZhaoX10}
D.~Zhao and D.~Xue, ``A comparative study of metamodeling methods considering
  sample quality merits,'' \emph{Struct. Multidiscip. Optim.}, vol.~42, no.~6,
  pp. 923--938, 2010.

\bibitem{HabibSR19}
A.~Habib, H.~K. Singh, and T.~Ray, ``A multiple surrogate assisted
  multi/many-objective multi-fidelity evolutionary algorithm,'' \emph{Inf.
  Sci.}, vol. 502, pp. 537--557, 2019.

\bibitem{BandlerBCGH94}
J.~W. Bandler, R.~M. Biernacki, S.~H. Chen, P.~A. Grobelny, and R.~H. Hemmers,
  ``Space mapping technique for electromagnetic optimization,'' \emph{IEEE
  Trans. Microw. Theory Tech.}, vol.~42, no.~12, pp. 2536--2544, 1994.

\bibitem{BandlerBCGHM95}
J.~W. Bandler, R.~M. Biernacki, S.~H. Chen, P.~A. Grobelny, R.~H. Hemmers, and
  K.~Madsen, ``Electromagnetic optimization exploiting aggressive space
  mapping,'' \emph{IEEE Trans. Microw. Theory Tech.}, vol.~43, no.~12, pp.
  2874--2882, 1995.

\bibitem{BandlerBCGHM98}
------, ``A trust region aggressive space mapping algorithm for em
  optimization,'' \emph{IEEE Trans. Microw. Theory Tech.}, vol.~46, no.~12, pp.
  2412--2425, 1998.

\bibitem{BakrBGM99}
M.~H. Bakr, J.~W. Bandler, N.~Georgieva, and K.~Madsen, ``A hybrid aggressive
  space-mapping algorithm for em optimization,'' \emph{IEEE Trans. Microw.
  Theory Tech.}, vol.~47, no.~12, pp. 2440--2449, 1999.

\bibitem{BakrBIRZ00}
M.~H. Bakr, J.~W. Bandler, M.~A. Ismail, J.~E. Rayas-S{\'a}nchez, and
  Q.~j.~Zhang, ``Neural space mapping em optimization of microwave
  structures,'' \emph{2000 IEEE MTT-S International Microwave Symposium
  Digest}, vol.~2, pp. 879--882 vol.2, 2000.

\bibitem{RenLKT16}
J.~Ren, L.~Ã. Leifsson, S.~Koziel, and Y.~A. Tesfahunegn, ``Multi-fidelity
  aerodynamic shape optimization using manifold mapping,'' in \emph{Proc. of
  the 57th AIAA/ASCE/AHS/ASC Structures, Structural Dynamics, and Materials
  Conference}, 2016, p. 0419.

\bibitem{Haftka91}
R.~T. Haftka, ``Combining global and local approximations,'' \emph{AIAA J.},
  vol.~29, no.~9, pp. 1523--1525, 1991.

\bibitem{ChangHGK93}
K.~J. Chang, R.~T. Haftka, G.~L. Giles, and P.-J. Kao, ``Sensitivity-based
  scaling for approximating structural response,'' \emph{J. Aircr.}, vol.~30,
  no.~2, pp. 283--288, 1993.

\bibitem{LewisN00}
R.~Lewis and S.~Nash, ``A multigrid approach to the optimization of systems
  governed by differential equations,'' in \emph{Proc. of the 8th Symposium on
  Multidisciplinary Analysis and Optimization}, 2000, p. 4890.

\bibitem{EldreGCd04}
M.~Eldred, A.~Giunta, and S.~Collis, ``Second-order corrections for
  surrogate-based optimization with model hierarchies,'' in \emph{Proc. of the
  10th AIAA/ISSMO Multidisciplinary Analysis and Optimization Conference},
  2004, p. 4457.

\bibitem{GanoPRBS04}
S.~Gano, V.~Perez, J.~Renaud, S.~Batill, and B.~Sanders, ``Multilevel variable
  fidelity optimization of a morphing unmanned aerial vehicle,'' in \emph{Proc.
  of the 45th AIAA/ASME/ASCE/AHS/ASC Structures, Structural Dynamics \&
  Materials Conference}, 2004, p. 1763.

\bibitem{GanoRS05}
S.~E. Gano, J.~E. Renaud, and B.~Sanders, ``Hybrid variable fidelity
  optimization by using a {Kriging}-based scaling function,'' \emph{AIAA J.},
  vol.~43, no.~11, pp. 2422--2433, 2005.

\bibitem{MarduelTT06}
X.~Marduel, C.~Tribes, and J.-Y. Tr{\'e}panier, ``Variable-fidelity
  optimization: efficiency and robustness,'' \emph{Optim. Eng.}, vol.~7, no.~4,
  pp. 479--500, 2006.

\bibitem{MarduelTT02}
------, ``Optimization using variable fidelity solvers: exploration of an
  approximation management framework for aerodynamic shape optimization,'' in
  \emph{Proc. of the 9th AIAA/ISSMO symposium on multidisciplinary analysis and
  optimization}, 2002, p. 5595.

\bibitem{WarnerM96}
B.~Warner and M.~Misra, ``Understanding neural networks as statistical tools,''
  \emph{Am. Stat.}, vol.~50, no.~4, pp. 284--293, 1996.

\bibitem{KimKN07}
H.~S. Kim, M.~Koc, and J.~Ni, ``A hybrid multi-fidelity approach to the optimal
  design of warm forming processes using a knowledge-based artificial neural
  network,'' \emph{Int. J. Mach. Tools Manuf.}, vol.~47, no.~2, pp. 211--222,
  2007.

\bibitem{TyanNL15}
M.~Tyan, N.~V. Nguyen, and J.-W. Lee, ``Improving variable-fidelity modelling
  by exploring global design space and radial basis function networks for
  aerofoil design,'' \emph{Eng. Optim.}, vol.~47, no.~7, pp. 885--908, 2015.

\bibitem{NguyenTL15}
N.~V. N.~M. Tyan and J.-W. Lee, ``A modified variable complexity modeling for
  efficient multidisciplinary aircraft conceptual design,'' \emph{Optim. Eng.},
  vol.~16, no.~2, pp. 483--505, 2015.

\bibitem{DurantinRDG17}
C.~Durantin, J.~Rouxel, J.-A. D{\'e}sid{\'e}ri, and A.~Gli{\`e}re,
  ``Multifidelity surrogate modeling based on radial basis functions,''
  \emph{Struct. Multidiscipl. Optim.}, vol.~56, no.~5, pp. 1061--1075, 2017.

\bibitem{SongLSZ19}
X.~Song, L.~Lv, W.~Sun, and J.~Zhang, ``A radial basis function-based
  multi-fidelity surrogate model: exploring correlation between high-fidelity
  and low-fidelity models,'' \emph{Struct. Multidiscipl. Optim.}, vol.~60,
  no.~3, pp. 965--981, 2019.

\bibitem{WangJYJ20}
H.~Wang, Y.~Jin, C.~Yang, and L.~Jiao, ``Transfer stacking from low-to
  high-fidelity: {A} surrogate-assisted bi-fidelity evolutionary algorithm,''
  \emph{Appl. Soft Comput.}, vol.~92, p. 106276, 2020.

\bibitem{YiGLSL19}
J.~Yi, L.~Gao, X.~Li, C.~A. Shoemaker, and C.~Lu, ``An on-line
  variable-fidelity surrogate-assisted harmony search algorithm with
  multi-level screening strategy for expensive engineering design
  optimization,'' \emph{Knowl. Based Syst.}, vol. 170, pp. 1--19, 2019.

\bibitem{ZhengSGJQ15}
J.~Zheng, X.~Shao, L.~Gao, P.~Jiang, and H.~Qiu, ``Difference mapping method
  using least square support vector regression for variable-fidelity
  metamodelling,'' \emph{Eng. Optim.}, vol.~47, no.~6, pp. 719--736, 2015.

\bibitem{ZhouSJZS15}
Q.~Zhou, X.~Shao, P.~Jiang, H.~Zhou, and L.~Shu, ``An adaptive global variable
  fidelity metamodeling strategy using a support vector regression based
  scaling function,'' \emph{Simul. Model. Pract. Theory}, vol.~59, pp. 18--35,
  2015.

\bibitem{ShiLSS20}
M.~Shi, L.~Lv, W.~Sun, and X.~Song, ``A multi-fidelity surrogate model based on
  support vector regression,'' \emph{Struct. Multidiscip. Optim.}, vol.~61,
  no.~6, pp. 2363--2375, 2020.

\bibitem{SunLL12}
G.~Sun, G.~Li, and Q.~Li, ``Variable fidelity design based surrogate and
  artificial bee colony algorithm for sheet metal forming process,''
  \emph{Finite Elem. Anal. Des.}, vol.~59, pp. 76--90, 2012.

\bibitem{Vitali98HS}
R.~Vitali, R.~Haftka, and B.~Sankar, ``Correction response surface
  approximations for stress intensity factors of a composite stiffened plate,''
  in \emph{Proc. of the 39th AIAA/ASME/ASCE/AHS/ASC Structures, Structural
  Dynamics, and Materials Conference and Exhibit}, 1998, p. 2047.

\bibitem{VitaliHS02}
R.~Vitali, R.~T. Haftka, and B.~V. Sankar, ``Multi-fidelity design of stiffened
  composite panel with a crack,'' \emph{Struct. Multidisc. Optim.}, vol.~23,
  pp. 347--356, 2002.

\bibitem{KnillGBGMHWA99}
D.~L. Knill, A.~A. Giunta, C.~A. Baker, B.~Grossman, W.~H. Mason, R.~T. Haftka,
  and L.~T. Watson, ``Response surface models combining linear and euler
  aerodynamics for supersonic transport design,'' \emph{J. Aircr.}, vol.~36,
  no.~1, pp. 75--86, 1999.

\bibitem{SunLZXYL11}
G.~Sun, G.~Li, S.~Zhou, W.~Xu, X.~Yang, and Q.~Li, ``Multi-fidelity
  optimization for sheet metal forming process,'' \emph{Struct. Multidiscipl.
  Optim.}, vol.~44, no.~1, pp. 111--124, 2011.

\bibitem{GanoRMS06}
S.~E. Gano, J.~E. Renaud, J.~D. Martin, and T.~W. Simpson, ``Update strategies
  for {Kriging} models used in variable fidelity optimization,'' \emph{Struct.
  Multidiscip. Optim.}, vol.~32, no.~4, pp. 287--298, 2006.

\bibitem{LiuC14}
Y.~Liu and M.~Collette, ``Improving surrogate-assisted variable fidelity
  multi-objective optimization using a clustering algorithm,'' \emph{Appl. Soft
  Comput.}, vol.~24, pp. 482--493, 2014.

\bibitem{XiongCT08}
Y.~Xiong, W.~Chen, and K.-L. Tsui, ``A new variable-fidelity optimization
  framework based on model fusion and objective-oriented sequential sampling,''
  \emph{J. Mech. Des.}, vol. 130, no.~11, pp. 1\,114\,011--1\,114\,019, 2008.

\bibitem{ZhouWCJSH17}
Q.~Zhou, Y.~Wang, S.~Choi, P.~Jiang, X.~Shao, and J.~Hu, ``A sequential
  multi-fidelity metamodeling approach for data regression,'' \emph{Knowl.
  Based Syst.}, vol. 134, pp. 199--212, 2017.

\bibitem{ZhuWC14}
J.~Zhu, Y.-J. Wang, and M.~Collette, ``A multi-objective variable-fidelity
  optimization method for genetic algorithms,'' \emph{Eng. Optim.}, vol.~46,
  no.~4, pp. 521--542, 2014.

\bibitem{ZhouSJGWS16}
Q.~Zhou, X.~Shao, P.~Jiang, Z.~Gao, C.~Wang, and L.~Shu, ``An active learning
  metamodeling approach by sequentially exploiting difference information from
  variable-fidelity models,'' \emph{Adv. Eng. Informatics}, vol.~30, no.~3, pp.
  283--297, 2016.

\bibitem{ZhouWXJ21}
Q.~Zhou, J.~Wu, T.~Xue, and P.~Jin, ``A two-stage adaptive multi-fidelity
  surrogate model-assisted multi-objective genetic algorithm for
  computationally expensive problems,'' \emph{Eng. Comput.}, vol.~37, no.~1,
  pp. 623--639, 2021.

\bibitem{LiTLSW21}
Z.~Li, K.~Tian, H.~Li, Y.~Shi, and B.~Wang, ``A competitive variable-fidelity
  surrogate-assisted {CMA-ES} algorithm using data mining techniques,''
  \emph{Aerosp. Sci. Technol.}, vol. 119, p. 107084, 2021.

\bibitem{KennedyO00}
M.~Kennedy and A.~O'Hagan, ``Predicting the output from a complex computer code
  when fast approximations are available,'' \emph{Biometrika}, vol.~87, no.~1,
  pp. 1--13, 2000.

\bibitem{JournelH89}
A.~G. Journel and C.~J. Huijbregts, ``Mining geostatistics,'' 1989.

\bibitem{HuangGZ13}
L.~Huang, Z.~Gao, and D.~Zhang, ``Research on multi-fidelity aerodynamic
  optimization methods,'' \emph{Chinese J. Aeronaut.}, vol.~26, no.~2, pp.
  279--286, 2013.

\bibitem{ForresterSK07}
A.~I.~J. Forrester, A.~S{\'o}bester, and A.~J. Keane, ``Multi-fidelity
  optimization via surrogate modelling,'' \emph{Proc. Math. Phys. Eng. Sci. P
  ROY SOC A-MATH PHY}, vol. 463, no. 2088, pp. 3251--3269, 2007.

\bibitem{DingK18}
F.~Ding and A.~Kareem, ``A multi-fidelity shape optimization via surrogate
  modeling for civil structures,'' \emph{J. Wind Eng. Ind. Aerodyn.}, vol. 178,
  pp. 49--56, 2018.

\bibitem{ZhangLCSY19}
H.~Zhang, Y.~Li, Z.~Chen, X.~Su, and X.~Yuan, ``Multi-fidelity model based
  optimization of shaped film cooling hole and experimental validation,''
  \emph{Int. J. Heat Mass Transf.}, vol. 132, pp. 118--129, 2019.

\bibitem{YongWTKS19}
H.~K. Yong, L.~Wang, D.~J. Toal, A.~J. Keane, and F.~Stanley, ``Multi-fidelity
  {K}riging-assisted structural optimization of whole engine models employing
  medial meshes,'' \emph{Struct. Multidiscipl. Optim.}, vol.~60, no.~3, pp.
  1209--1226, 2019.

\bibitem{HanZG12}
Z.-H. Han, R.~Zimmermann, and S.~G{\"o}rtz, ``Alternative co-{Kriging} method
  for variable-fidelity surrogate modeling,'' \emph{AIAA J.}, vol.~50, no.~5,
  pp. 1205--1210, 2012.

\bibitem{ZimmermannH10}
R.~Zimmermannand and Z.-H. Han, ``Simplified cross-correlation estimation for
  multi-fidelity surrogate co-{Kriging} models,'' \emph{Adv. Appl. Math. Sci.},
  vol.~7, no.~2, pp. 181--202, 2010.

\bibitem{Gratiet14}
L.~L. Gratiet, ``Recursive co-{Kriging} model for design of computer
  experiments with multiple levels of fidelity,'' \emph{Int. J. Uncertain.
  Quantif.}, vol.~4, no.~5, 2014.

\bibitem{PalarS17}
P.~S. Palar and K.~Shimoyama, ``Multi-fidelity uncertainty analysis in cfd
  using hierarchical {Kriging},'' in \emph{Proc. of the 35th AIAA Applied
  Aerodynamics Conference}, 2017, p. 3261.

\bibitem{ZhangHZ18}
Y.~Zhang, Z.-H. Han, and K.-S. Zhang, ``Variable-fidelity expected improvement
  method for efficient global optimization of expensive functions,''
  \emph{Struct. Multidiscipl. Optim.}, vol.~58, no.~4, pp. 1431--1451, 2018.

\bibitem{LiuHZSSGT19}
F.~Liu, Z.-H. Han, Y.~Zhang, K.~Song, W.-P. Song, F.~Gui, and J.-B. Tang,
  ``Surrogate-based aerodynamic shape optimization of hypersonic flows
  considering transonic performance,'' \emph{Aerosp. Sci. Technol.}, vol.~93,
  p. 105345, 2019.

\bibitem{BoPBK18}
L.~Bonfiglio, P.~Perdikaris, S.~Brizzolara, and G.~Karniadakis,
  ``Multi-fidelity optimization of super-cavitating hydrofoils,'' \emph{Comput.
  Methods Appl. Mech. Eng.}, vol. 332, pp. 63--85, 2018.

\bibitem{ZhangWJCZ22}
L.~Zhang, Y.~Wu, P.~Jiang, S.~Choi, and Q.~Zhou, ``A multi-fidelity surrogate
  modeling approach for incorporating multiple non-hierarchical low-fidelity
  data,'' \emph{Adv. Eng. Informatics}, vol.~51, p. 101430, 2022.

\bibitem{AlvarezRL12}
M.~A. {\'{A}}lvarez, L.~Rosasco, and N.~D. Lawrence, ``Kernels for
  vector-valued functions: {A} review,'' \emph{Found. Trends Mach. Learn.},
  vol.~4, no.~3, pp. 195--266, 2012.

\bibitem{LiCY23}
K.~Li, R.~Chen, and X.~Yao, ``A data-driven evolutionary transfer optimization
  for expensive problems in dynamic environments,'' \emph{IEEE Trans. Evol.
  Comput.}, 2023.

\bibitem{YangL23}
H.~Yang and K.~Li, ``Instoptima: Evolutionary multi-objective instruction
  optimization via large language model-based instruction operators,'' in
  \emph{EMNLP'23: Findings of the Association for Computational Linguistics:
  {EMNLP} 2023}.\hskip 1em plus 0.5em minus 0.4em\relax Association for
  Computational Linguistics, 2023, pp. 13\,593--13\,602.

\bibitem{ChenL23}
R.~Chen and K.~Li, ``Data-driven evolutionary multi-objective optimization
  based on multiple-gradient descent for disconnected pareto fronts,'' in
  \emph{EMO'23: Proc. of the 12th International Conference on Evolutionary
  Multi-Criterion Optimization}, ser. Lecture Notes in Computer Science, vol.
  13970.\hskip 1em plus 0.5em minus 0.4em\relax Springer, 2023, pp. 56--70.

\bibitem{SwerskySA13}
K.~Swersky, J.~Snoek, and R.~P. Adams, ``Multi-task {Bayesian} optimization,''
  in \emph{NIPS'13: Proc. of the Annual Conference on Neural Information
  Processing Systems 2013}, 2013, pp. 2004--2012.

\bibitem{TakenoFTKSTK20}
S.~Takeno, H.~Fukuoka, Y.~Tsukada, T.~Koyama, M.~Shiga, I.~Takeuchi, and
  M.~Karasuyama, ``Multi-fidelity {Bayesian} optimization with max-value
  entropy search and its parallelization,'' in \emph{ICML'J20: Proc. of the
  37th International Conference on Machine Learning}, ser. Proceedings of
  Machine Learning Research, vol. 119.\hskip 1em plus 0.5em minus 0.4em\relax
  {PMLR}, 2020, pp. 9334--9345.

\bibitem{MossLR20}
H.~B. Moss, D.~S. Leslie, and P.~Rayson, ``{MUMBO:} multi-task max-value
  {Bayesian} optimization,'' in \emph{ECML/PKDD'20: Proc. of the 2020 European
  Conference on Machine Learning and Knowledge Discovery in Databases}, ser.
  Lecture Notes in Computer Science, vol. 12459.\hskip 1em plus 0.5em minus
  0.4em\relax Springer, 2020, pp. 447--462.

\bibitem{TakenoFTKSTK22}
S.~Takeno, H.~Fukuoka, Y.~Tsukada, T.~Koyama, M.~Shiga, I.~Takeuchi, and
  M.~Karasuyama, ``A generalized framework of multifidelity max-value entropy
  search through joint entropy,'' \emph{Neural Comput.}, vol.~34, no.~10, pp.
  2145--2203, 2022.

\bibitem{LinQCZH22}
Q.~Lin, J.~Qian, Y.~Cheng, Q.~Zhou, and J.~Hu, ``A multi-output multi-fidelity
  {Gaussian} process model for non-hierarchical low-fidelity data fusion,''
  \emph{Knowl. Based Syst.}, vol. 254, p. 109645, 2022.

\bibitem{MikkolaMFK23}
P.~Mikkola, J.~Martinelli, L.~Filstroff, and S.~Kaski, ``Multi-fidelity
  {Bayesian} optimization with unreliable information sources,'' in
  \emph{AISTATS'23: Proc. of the 2023 International Conference on Artificial
  Intelligence and Statistics}, vol. 206.\hskip 1em plus 0.5em minus
  0.4em\relax {PMLR}, 2023, pp. 7425--7454.

\bibitem{PoloczekWF17}
M.~Poloczek, J.~Wang, and P.~I. Frazier, ``Multi-information source
  optimization,'' in \emph{NeurIPS'17: Annual Conference on Neural Information
  Processing Systems 2017}, 2017, pp. 4288--4298.

\bibitem{ZhangHLK17}
Y.~Zhang, T.~N. Hoang, B.~K.~H. Low, and M.~Kankanhalli, ``Information-based
  multi-fidelity {Bayesian} optimization,'' in \emph{NIPS Workshop on
  {Bayesian} Optimization}, 2017.

\bibitem{CutajarPDLG19}
K.~Cutajar, M.~Pullin, A.~C. Damianou, N.~D. Lawrence, and J.~Gonz{\'{a}}lez,
  ``Deep {Gaussian} processes for multi-fidelity modeling,'' \emph{CoRR}, vol.
  abs/1903.07320, 2019.

\bibitem{PerdikarisRDLK17}
P.~Perdikaris, M.~Raissi, A.~Damianou, N.~D. Lawrence, and G.~E. Karniadakis,
  ``Nonlinear information fusion algorithms for data-efficient multi-fidelity
  modelling,'' \emph{Proc. R. Soc. A: Math. Phys. Eng. Sci.}, vol. 473, no.
  2198, p. 20160751, 2017.

\bibitem{RequeimaTBT19}
J.~Requeima, W.~Tebbutt, W.~Bruinsma, and R.~E. Turner, ``The {Gaussian}
  process autoregressive regression model {(GPAR)},'' in \emph{AISTAS'19: Proc.
  of the 22nd International Conference on Artificial Intelligence and
  Statistics}, vol.~89, 2019, pp. 1860--1869.

\bibitem{DamianouL13}
A.~C. Damianou and N.~D. Lawrence, ``Deep {Gaussian} processes,'' in
  \emph{{AISTATS}'13: Proc. of the Sixteenth International Conference on
  Artificial Intelligence and Statistics}, ser. {JMLR} Workshop and Conference
  Proceedings, vol.~31.\hskip 1em plus 0.5em minus 0.4em\relax JMLR.org, 2013,
  pp. 207--215.

\bibitem{SalimbeniD17}
H.~Salimbeni and M.~P. Deisenroth, ``Doubly stochastic variational inference
  for deep {Gaussian} processes,'' in \emph{NeurIPS'17: Annual Conference on
  Neural Information Processing Systems 2017}, 2017, pp. 4588--4599.

\bibitem{LiXKZ20}
S.~Li, W.~Xing, R.~M. Kirby, and S.~Zhe, ``Multi-fidelity {Bayesian}
  optimization via deep neural networks,'' in \emph{NeurIPS'20: Proc. of the
  Annual Conference on Neural Information Processing Systems 2020}, 2020.

\bibitem{LiKZ21}
S.~Li, R.~M. Kirby, and S.~Zhe, ``Batch multi-fidelity {Bayesian} optimization
  with deep auto-regressive networks,'' in \emph{NeurIPS'21: Proc. of the
  Annual Conference on Neural Information Processing Systems 2021}, 2021, pp.
  25\,463--25\,475.

\bibitem{XuSZ21}
Y.~Xu, X.~Song, and C.~Zhang, ``Hierarchical regression framework for
  multi-fidelity modeling,'' \emph{Knowl. Based Syst.}, vol. 212, p. 106587,
  2021.

\bibitem{RaissiPK17}
\BIBentryALTinterwordspacing
M.~Raissi, P.~Perdikaris, and G.~E. Karniadakis, ``Physics informed deep
  learning (part {I):} data-driven solutions of nonlinear partial differential
  equations,'' \emph{CoRR}, vol. abs/1711.10561, 2017. [Online]. Available:
  \url{http://arxiv.org/abs/1711.10561}
\BIBentrySTDinterwordspacing

\bibitem{MengK20}
X.~Meng and G.~E. Karniadakis, ``A composite neural network that learns from
  multi-fidelity data: Application to function approximation and inverse {PDE}
  problems,'' \emph{J. Comput. Phys.}, vol. 401, 2020.

\bibitem{LuDKRKS20}
L.~Lu, M.~Dao, P.~Kumar, U.~Ramamurty, G.~E. Karniadakis, and S.~Suresh,
  ``Extraction of mechanical properties of materials through deep learning from
  instrumented indentation,'' \emph{Proc. Natl. Acad. Sci. U. S. A.}, vol. 117,
  pp. 7052 -- 7062, 2020.

\bibitem{Chakraborty21}
S.~Chakraborty, ``Transfer learning based multi-fidelity physics informed deep
  neural network,'' \emph{J. Comput. Phys.}, vol. 426, p. 109942, 2021.

\bibitem{HaradaRM22}
K.~Harada, D.~Rajaram, and D.~N. Mavris, ``Application of multi-fidelity
  physics-informed neural network on transonic airfoil using wind tunnel
  measurements,'' \emph{AIAA SCITECH 2022 Forum}, 2022.

\bibitem{MengWFTK21}
X.~Meng, Z.~Wang, D.~Fan, M.~S. Triantafyllou, and G.~E. Karniadakis, ``A fast
  multi-fidelity method with uncertainty quantification for complex data
  correlations: Application to vortex-induced vibrations of marine risers,''
  \emph{Comput. Methods Appl. Mech. Eng.}, vol. 386, p. 114212, 2021.

\bibitem{LiuW19}
D.~Liu and Y.~Wang, ``Multi-fidelity physics-constrained neural network and its
  application in materials modeling,'' \emph{J. Mech. Des.}, vol. 141, no.~12,
  p. 121403, 2019.

\bibitem{DolpertM97}
D.~H. Wolpert and W.~G. Macready, ``No free lunch theorems for optimization,''
  \emph{{IEEE} Trans. Evol. Comput.}, vol.~1, no.~1, pp. 67--82, 1997.

\bibitem{AlexandrovLGGN01}
N.~M. Alexandrov, R.~M. Lewis, C.~R. Gumbert, L.~L. Green, and P.~A. Newman,
  ``Approximation and model management in aerodynamic optimization with
  variable-fidelity models,'' \emph{J. Aircr.}, vol.~38, pp. 1093--1101, 2001.

\bibitem{AlexandrovLDLT98}
N.~M. Alexandrov, J.~E.~D. Jr., R.~M. Lewis, and V.~Torczon, ``A trust-region
  framework for managing the use of approximation models in optimization,''
  \emph{Struct. Multidiscipl. Optim.}, vol.~15, pp. 16--23, 1998.

\bibitem{AlexandrovLGGN99}
N.~M. Alexandrov, R.~M. Lewis, C.~R. Gumbert, L.~L. Green, and P.~A. Newman,
  ``Optimization with variable-fidelity models applied to wing design,'' NASA,
  Tech. Rep. CR-209826, 1999.

\bibitem{HanXLYKS20}
Z.~Han, C.~Xu, Z.~Liang, Z.~Yu, K.~Zhang, and W.~Song, ``Efficient aerodynamic
  shape optimization using variable-fidelity surrogate models and multilevel
  computational grids,'' \emph{Chinese J. Aeronaut.}, vol.~33, no.~1, pp.
  31--47, 2020.

\bibitem{HuangANM06}
D.~Huang, T.~T. Allen, W.~I. Notz, and R.~A. Miller, ``Sequential kriging
  optimization using multiple-fidelity evaluations,'' \emph{Struct.
  Multidiscipl. Optim.}, vol.~32, pp. 369--382, 2006.

\bibitem{ReisenthelT14}
P.~H. Reisenthel and T.~T. Allen, ``Application of multifidelity expected
  improvement algorithms to aeroelastic design optimization,'' in
  \emph{AIAA'14: Proc. of the 10th AIAA Multidisciplinary Design Optimization
  Conference}, 2014, p. 1490.

\bibitem{FioreM22}
F.~D. Fiore and L.~Mainini, ``Non-myopic multifidelity {Bayesian}
  optimization,'' \emph{CoRR}, vol. abs/2207.06325, 2022.

\bibitem{MossLGR21}
H.~B. Moss, D.~S. Leslie, J.~Gonzalez, and P.~Rayson, ``{GIBBON:}
  general-purpose information-based {Bayesian} optimisation,'' \emph{J. Mach.
  Learn. Res.}, vol.~22, pp. 235:1--235:49, 2021.

\bibitem{LuongNGRV21}
P.~Luong, D.~Nguyen, S.~Gupta, S.~Rana, and S.~Venkatesh, ``Adaptive cost-aware
  {Bayesian} optimization,'' \emph{Knowl. Based Syst.}, vol. 232, p. 107481,
  2021.

\bibitem{LeePAS20}
E.~H. Lee, V.~Perrone, C.~Archambeau, and M.~W. Seeger, ``Cost-aware {Bayesian}
  optimization,'' \emph{CoRR}, vol. abs/2003.10870, 2020.

\bibitem{GonzalezDHL16}
J.~Gonz{\'{a}}lez, Z.~Dai, P.~Hennig, and N.~D. Lawrence, ``Batch bayesian
  optimization via local penalization,'' in \emph{{AISTATS}'16: Proc. of the
  19th International Conference on Artificial Intelligence and Statistics},
  vol.~51.\hskip 1em plus 0.5em minus 0.4em\relax JMLR.org, 2016, pp. 648--657.

\bibitem{KandasamyKSP18}
K.~Kandasamy, A.~Krishnamurthy, J.~Schneider, and B.~P{\'{o}}czos,
  ``Parallelised bayesian optimisation via thompson sampling,'' in
  \emph{{AISTATS}'18: Proc. of the 21st International Conference on Artificial
  Intelligence and Statistics}, vol.~84.\hskip 1em plus 0.5em minus 0.4em\relax
  {PMLR}, 2018, pp. 133--142.

\bibitem{SavageBMR22}
T.~Savage, N.~Basha, O.~K. Matar, and E.~A. del Rio{-}Chanona, ``Deep
  {Gaussian} process-based multi-fidelity bayesian optimization for simulated
  chemical reactors,'' \emph{CoRR}, vol. abs/2210.17213, 2022.

\bibitem{YiSS20}
J.~Yi, Y.~Shen, and C.~A. Shoemaker, ``A multi-fidelity {RBF} surrogate-based
  optimization framework for computationally expensive multi-modal problems
  with application to capacity planning of manufacturing systems,''
  \emph{Struct. Multidiscipl. Optim.}, vol.~62, no.~4, pp. 1787--1807, 2020.

\bibitem{KennyRS23}
A.~Kenny, T.~Ray, and H.~K. Singh, ``An iterative two-stage multifidelity
  optimization algorithm for computationally expensive problems,'' \emph{{IEEE}
  Trans. Evol. Comput.}, vol.~27, no.~3, pp. 520--534, 2023.

\bibitem{LattimoreS20}
T.~Lattimore and C.~Szepesv\'ari, \emph{Bandit Algorithms}.\hskip 1em plus
  0.5em minus 0.4em\relax Cambridge University Press, 2020.

\bibitem{JamiesonT16}
K.~G. Jamieson and A.~Talwalkar, ``Non-stochastic best arm identification and
  hyperparameter optimization,'' in \emph{AISTATS'16: Proc. of the 19th
  International Conference on Artificial Intelligence and Statistics}, vol.~51,
  2016, pp. 240--248.

\bibitem{KarninKS13}
Z.~S. Karnin, T.~Koren, and O.~Somekh, ``Almost optimal exploration in
  multi-armed bandits,'' in \emph{ICML'13: Proc. of the 30th International
  Conference on Machine Learning}, vol.~28, 2013, pp. 1238--1246.

\bibitem{FalknerKH18}
S.~Falkner, A.~Klein, and F.~Hutter, ``{BOHB:} robust and efficient
  hyperparameter optimization at scale,'' in \emph{ICML'18: Proc. of the 35th
  International Conference on Machine Learning}, ser. Proceedings of Machine
  Learning Research, vol.~80.\hskip 1em plus 0.5em minus 0.4em\relax {PMLR},
  2018, pp. 1436--1445.

\bibitem{BergstraBBK11}
J.~Bergstra, R.~Bardenet, Y.~Bengio, and B.~K{\'{e}}gl, ``Algorithms for
  hyper-parameter optimization,'' in \emph{NIPS'11: Proc. of 25th Annual
  Conference on Neural Information Processing Systems 2011}, 2011, pp.
  2546--2554.

\bibitem{WangXW18}
J.~Wang, J.~Xu, and X.~Wang, ``Combination of hyperband and {Bayesian}
  optimization for hyperparameter optimization in deep learning,'' \emph{CoRR},
  vol. abs/1801.01596, 2018.

\bibitem{AwadMH21}
N.~H. Awad, N.~Mallik, and F.~Hutter, ``{DEHB:} evolutionary hyberband for
  scalable, robust and efficient hyperparameter optimization,'' in
  \emph{{IJCAI}'21: Proc. of the Thirtieth International Joint Conference on
  Artificial Intelligence}.\hskip 1em plus 0.5em minus 0.4em\relax ijcai.org,
  2021, pp. 2147--2153.

\bibitem{LiSJG0021}
Y.~Li, Y.~Shen, J.~Jiang, J.~Gao, C.~Zhang, and B.~Cui, ``{MFES-HB:} efficient
  hyperband with multi-fidelity quality measurements,'' in \emph{{AAAI}'21:
  Proc. of the AAAI Conference on Artificial Intelligence}.\hskip 1em plus
  0.5em minus 0.4em\relax {AAAI} Press, 2021, pp. 8491--8500.

\bibitem{ZhuLGBL22}
X.~Zhu, Y.~Liu, P.~Ghysels, D.~Bindel, and X.~S. Li, ``{GPTuneBand}: Multi-task
  and multi-fidelity autotuning for large-scale high performance computing
  applications,'' in \emph{PPSC'22: Proc. of the 2022 {SIAM} Conference on
  Parallel Processing for Scientific Computing}, 2022, pp. 1--13.

\bibitem{LiJRGBHRT20}
L.~Li, K.~G. Jamieson, A.~Rostamizadeh, E.~Gonina, J.~Ben{-}tzur, M.~Hardt,
  B.~Recht, and A.~Talwalkar, ``A system for massively parallel hyperparameter
  tuning,'' in \emph{MLSys'20: Proc. of Machine Learning and Systems
  2020}.\hskip 1em plus 0.5em minus 0.4em\relax mlsys.org, 2020.

\bibitem{BohdalBWEAZ23}
O.~Bohdal, L.~Balles, M.~Wistuba, B.~Ermis, C.~Archambeau, and G.~Zappella,
  ``{PASHA:} efficient {HPO} and {NAS} with progressive resource allocation,''
  in \emph{ICLR'23: Proc. of The Eleventh International Conference on Learning
  Representations}, 2023.

\bibitem{LiSJZLLZC22}
Y.~Li, Y.~Shen, H.~Jiang, W.~Zhang, J.~Li, J.~Liu, C.~Zhang, and B.~Cui,
  ``{Hyper-Tune}: Towards efficient hyper-parameter tuning at scale,''
  \emph{Proc. {VLDB} Endow.}, vol.~15, no.~6, pp. 1256--1265, 2022.

\bibitem{LiHLZZ20}
H.~Li, Z.~Huang, X.~Liu, C.~Zeng, and P.~Zou, ``Multi-fidelity
  meta-optimization for nature inspired optimization algorithms,'' \emph{Appl.
  Soft Comput.}, vol.~96, p. 106619, 2020.

\bibitem{ElBeltagyAA99}
M.~A. El-Beltagy and A.~J. Keane, ``A comparison of various optimization
  algorithms on a multilevel problem,'' \emph{Eng. Appl. Artif. Intell.},
  vol.~12, no.~5, pp. 639--654, 1999.

\bibitem{LiuWY22}
S.~Liu, H.~Wang, and W.~Yao, ``A surrogate-assisted evolutionary algorithm with
  hypervolume triggered fidelity adjustment for noisy multiobjective integer
  programming,'' \emph{Appl. Soft Comput.}, vol. 126, p. 109263, 2022.

\bibitem{JiangZ0C19}
P.~Jiang, Q.~Zhou, J.~Liu, and Y.~Cheng, ``A three-stage surrogate model
  assisted multi-objective genetic algorithm for computationally expensive
  problems,'' in \emph{CEC'19: Proc. of 2019 {IEEE} Congress on Evolutionary
  Computation}, 2019, pp. 1680--1687.

\bibitem{MamunSR22}
M.~M. Mamun, H.~K. Singh, and T.~Ray, ``A multifidelity approach for bilevel
  optimization with limited computing budget,'' \emph{{IEEE} Trans. Evol.
  Comput.}, vol.~26, no.~2, pp. 392--399, 2022.

\bibitem{YangHW22}
H.~Yang, S.~H. Hong, and Y.~Wang, ``A sequential multi-fidelity surrogate-based
  optimization methodology based on expected improvement reduction,''
  \emph{Struct. Multidiscipl. Optim.}, vol.~65, no.~5, p. 153, 2022.

\bibitem{YangW23}
H.~Yang and Y.~Wang, ``A sparse multi-fidelity surrogate-based optimization
  method with computational awareness,'' \emph{Eng. Comput.}, vol.~39, no.~5,
  pp. 3473--3489, 2023.

\bibitem{LiuXZ20}
Z.~Liu, H.~Xu, and P.~Zhu, ``An adaptive multi-fidelity approach for design
  optimization of mesostructure-structure systems,'' \emph{Struct.
  Multidiscipl. Optim.}, vol.~62, pp. 375--386, 2020.

\bibitem{ShuJZSHM18}
L.~Shu, P.~Jiang, Q.~Zhou, X.~Shao, J.~Hu, and X.~Meng, ``An on-line variable
  fidelity metamodel assisted multi-objective genetic algorithm for engineering
  design optimization,'' \emph{Appl. Soft Comput.}, vol.~66, pp. 438--448,
  2018.

\bibitem{LimOJS08}
D.~Lim, Y.~Ong, Y.~Jin, and B.~Sendhoff, ``Evolutionary optimization with
  dynamic fidelity computational models,'' in \emph{ICIC'08: Proc. of the 4th
  International Conference on Intelligent Computing}, ser. Lecture Notes in
  Computer Science, vol. 5227.\hskip 1em plus 0.5em minus 0.4em\relax Springer,
  2008, pp. 235--242.

\bibitem{BrysonR16}
D.~E. Bryson and M.~P. Rumpfkeil, ``Variable-fidelity surrogate modeling of
  lambda wing transonic aerodynamic performance,'' in \emph{Proc. of the 54th
  AIAA Aerospace Sciences Meeting}.\hskip 1em plus 0.5em minus 0.4em\relax
  AIAA, 2016, p. 0294.

\bibitem{WangJD18}
H.~Wang, Y.~Jin, and J.~Doherty, ``A generic test suite for evolutionary
  multifidelity optimization,'' \emph{{IEEE} Trans. Evol. Comput.}, vol.~22,
  no.~6, pp. 836--850, 2018.

\bibitem{LiWD18}
C.~Li, P.~Wang, and H.~Dong, ``Kriging-based multi-fidelity optimization via
  information fusion with uncertainty,'' \emph{J. Mech. Sci. Technol.},
  vol.~32, no.~1, pp. 245--259, 2018.

\bibitem{SePWQVD19}
A.~Serani, R.~Pellegrini, J.~Wackers, C.-E. Jeanson, P.~Queutey, M.~Visonneau,
  and M.~Diez, ``Adaptive multi-fidelity sampling for {CFD}-based optimisation
  via radial basis function metamodels,'' \emph{Int. J. Comut. Fluid Dyn.},
  vol.~33, no. 6-7, pp. 237--255, 2019.

\bibitem{RumpfkelB20}
M.~P. Rumpfkeil and P.~S. Beran, ``Multi-fidelity, gradient-enhanced, and
  locally optimized sparse polynomial chaos and {Kriging} surrogate models
  applied to benchmark problems,'' in \emph{AIAA Scitech 2020 Forum}, 2020, p.
  0677.

\bibitem{RijnS20}
S.~van Rijn and S.~Schmitt, ``{MF2:} {A} collection of multi-fidelity benchmark
  functions in python,'' \emph{J. Open Source Softw.}, vol.~5, no.~52, p. 2049,
  2020.

\bibitem{Toal15}
D.~J. Toal, ``Some considerations regarding the use of multi-fidelity {Kriging}
  in the construction of surrogate models,'' \emph{Struct. Multidiscipl.
  Optim.}, vol.~51, no.~6, pp. 1223--1245, 2015.

\bibitem{MaininiSRMQPYFPFBNDB22}
L.~Mainini, A.~Serani, M.~P. Rumpfkeil, E.~A. Minisci, D.~Quagliarella,
  H.~Pehlivan, S.~Yildiz, S.~Ficini, R.~Pellegrini, F.~D. Fiore, D.~E. Bryson,
  M.~Nikbay, M.~Diez, and P.~Beran, ``Analytical benchmark problems for
  multifidelity optimization methods,'' in \emph{Research Workshop AVT-354 on
  Multi-fidelity Methods for Military Vehicle Design}, 2022.

\bibitem{KennyRSL23}
A.~Kenny, T.~Ray, H.~K. Singh, and X.~Li, ``A test suite for multi-objective
  multi-fidelity optimization,'' in \emph{EMO'23: Proc. of the 12th
  International Conference on Evolutionary Multi-Criterion Optimization}.\hskip
  1em plus 0.5em minus 0.4em\relax Springer, 2023, pp. 361--373.

\bibitem{LeCunBH15}
Y.~LeCun, Y.~Bengio, and G.~E. Hinton, ``Deep learning,'' \emph{Nature}, vol.
  521, no. 7553, pp. 436--444, 2015.

\bibitem{FeurerH19}
M.~Feurer and F.~Hutter, ``Hyperparameter optimization,'' in \emph{Automated
  Machine Learning - Methods, Systems, Challenges}, ser. The Springer Series on
  Challenges in Machine Learning, F.~Hutter, L.~Kotthoff, and J.~Vanschoren,
  Eds.\hskip 1em plus 0.5em minus 0.4em\relax Springer, 2019, pp. 3--33.

\bibitem{KleinFBHH17}
A.~Klein, S.~Falkner, S.~Bartels, P.~Hennig, and F.~Hutter, ``Fast {Bayesian}
  optimization of machine learning hyperparameters on large datasets,'' in
  \emph{{AISTATS}'17: Proc. of the 20th International Conference on Artificial
  Intelligence and Statistics}, vol.~54, 2017, pp. 528--536.

\bibitem{KandasamyDSP17}
K.~Kandasamy, G.~Dasarathy, J.~G. Schneider, and B.~P{\'{o}}czos,
  ``Multi-fidelity bayesian optimisation with continuous approximations,'' in
  \emph{ICML'17: Proc. of the 34th International Conference on Machine
  Learning}, vol.~70.\hskip 1em plus 0.5em minus 0.4em\relax {PMLR}, 2017, pp.
  1799--1808.

\bibitem{BertrandAPB17}
H.~Bertrand, R.~Ardon, M.~Perrot, and I.~Bloch, ``Hyperparameter optimization
  of deep neural networks: combining hperband with bayesian model selection,''
  in \emph{Conf{\'{e}}rence francophone sur l'apprentissage automatique}, 2017.

\bibitem{WistubaKG22}
M.~Wistuba, A.~Kadra, and J.~Grabocka, ``Supervising the multi-fidelity race of
  hyperparameter configurations,'' in \emph{NeurIPS'22: Proc. of the Annual
  Conference on Neural Information Processing Systems 2022}, 2022.

\bibitem{LeifssonK10}
L.~Leifsson and S.~Koziel, ``Multi-fidelity design optimization of transonic
  airfoils using shape-preserving response prediction,'' \emph{Procedia Comput.
  Sci.}, vol.~1, no.~1, pp. 1311--1320, 2010.

\bibitem{BenamaraBL16}
T.~Benamara, P.~Breitkopf, I.~Lepot, and C.~Sainvitu, ``Adaptive infill
  sampling criterion for multi-fidelity optimization based on {Gappy-POD}:
  {Application} to the flight domain study of a transonic airfoil,''
  \emph{Struct. Multidisc. Optim.}, vol.~54, pp. 843--855, 2016.

\bibitem{BarrettBK06}
T.~R. Barrett, N.~W. Bressloff, and A.~J. Keane, ``Airfoil shape design and
  optimization using multifidelity analysis and embedded inverse design,''
  \emph{AIAA J.}, vol.~44, no.~9, pp. 2051--2060, 2006.

\bibitem{MarchW2012}
A.~March and K.~Willcox, ``Constrained multifidelity optimization using model
  calibration,'' \emph{Struct. Multidisc. Optim.}, vol.~46, pp. 93--109, 2012.

\bibitem{AyeWKTBYP23}
C.~Aye, K.~Wansaseub, S.~Kumar, G.~Tejani, S.~Bureerat, A.~Yildiz, and
  N.~Pholdee, ``Airfoil shape optimisation using a multi-fidelity
  surrogate-assisted metaheuristic with a new multi-objective infill sampling
  technique,'' \emph{CMES-Computer Modeling in Engineering \& Sciences}, 2023.

\bibitem{DemangeSK16}
J.~Demange, A.~M. Savill, and T.~Kipouros, ``Multifidelity optimization for
  high-lift airfoils,'' in \emph{The Proc. of the 54th AIAA Aerospace Sciences
  Meeting}, 2016, p. 0557.

\bibitem{TaoS19}
J.~Tao and G.~Sun, ``Application of deep learning based multi-fidelity
  surrogate model to robust aerodynamic design optimization,'' \emph{Aerosp.
  Sci. Technol.}, vol.~92, pp. 722--737, 2019.

\bibitem{AriyaritK17}
A.~Ariyarit and M.~Kanazaki, ``Multi-fidelity multi-objective efficient global
  optimization applied to airfoil design problems,'' \emph{Appl. Sci.}, vol.~7,
  no.~12, p. 1318, 2017.

\bibitem{PriyankaS21}
R.~Priyanka and M.~Sivapragasam, ``Multi-fidelity surrogate model-based airfoil
  optimization at a transitional low {Reynolds} number,''
  \emph{S{\=a}dhan{\=a}}, vol.~46, no.~1, p.~58, 2021.

\bibitem{TaoACGPG19}
S.~Tao, D.~W. Apley, W.~Chen, A.~Garbo, D.~J. Pate, and B.~J. German, ``Input
  mapping for model calibration with application to wing aerodynamics,''
  \emph{AIAA J.}, vol.~57, no.~7, pp. 2734--2745, 2019.

\bibitem{Elham15}
A.~Elham, ``Adjoint quasi-three-dimensional aerodynamic solver for
  multi-fidelity wing aerodynamic shape optimization,'' \emph{Aerosp. Sci.
  Technol.}, vol.~41, pp. 241--249, 2015.

\bibitem{BrahmacharyNS17}
S.~Brahmachary, G.~Natarajan, and N.~Sahoo, ``A hybrid aerodynamic shape
  optimization approach for axisymmetric body in hypersonic flow,'' in
  \emph{Fluid Mechanics and Fluid Power--Contemporary Research: Proceedings of
  the 5th International and 41st National Conference on FMFP 2014}.\hskip 1em
  plus 0.5em minus 0.4em\relax Springer, 2017, pp. 301--311.

\bibitem{RajnarayanHK08}
D.~Rajnarayan, A.~Haas, and I.~Kroo, ``A multifidelity gradient-free
  optimization method and application to aerodynamic design,'' in \emph{Proc.
  of the 12th AIAA/ISSMO multidisciplinary analysis and optimization
  conference}, 2008, p. 6020.

\bibitem{TancredR15}
J.~Tancred and M.~P. Rumpfkeil, ``Aerodynamic response quantification of
  complex hypersonic configurations using variable fidelity surrogate
  modeling,'' in \emph{Proc. of the 53rd AIAA Aerospace Sciences Meeting},
  2015, p. 1703.

\bibitem{BahramiTDVG16}
S.~Bahrami, C.~Tribes, C.~Devals, T.~C. Vu, and F.~Guibault, ``Multi-fidelity
  shape optimization of hydraulic turbine runner blades using a multi-objective
  mesh adaptive direct search algorithm,'' \emph{Appl. Math. Model.}, vol.~40,
  no.~2, pp. 1650--1668, 2016.

\bibitem{BahramiTVVF14}
S.~Bahrami, C.~Tribes, S.~V. F. T.~C. Vu, and F.~Guibault, ``Multi-fidelity
  design optimization of francis turbine runner blades,'' in \emph{IOP
  Conference Series: Earth and Environmental Science}, vol.~22, no.~1.\hskip
  1em plus 0.5em minus 0.4em\relax IOP Publishing, 2014, p. 012029.

\bibitem{GuoSPLH18}
Z.~Guo, L.~Song, C.~Park, J.~Li, and R.~T. Haftka, ``Analysis of dataset
  selection for multi-fidelity surrogates for a turbine problem,''
  \emph{Struct. Multidiscip. Optim.}, vol.~57, pp. 2127--2142, 2018.

\bibitem{ToalZKLZ21}
D.~J.~J. Toal, X.~Zhang, A.~J. Keane, C.~Y. Lee, and M.~Zedda, ``The potential
  of a multifidelity approach to gas turbine combustor design optimization,''
  \emph{J. Eng. Gas Turbine. Power.}, vol. 143, no.~5, p. 051002, 2021.

\bibitem{NacharBNR20}
S.~Nachar, P.-A. Boucard, D.~N{\'e}ron, and C.~Rey, ``Multi-fidelity {Bayesian}
  optimization using model-order reduction for viscoplastic structures,''
  \emph{Finite Elem. Anal. Des.}, vol. 176, p. 103400, 2020.

\bibitem{ToalKBDYPRRK2014}
D.~J.~J. Toal, A.~J. Keane, D.~Benito, J.~A. Dixon, J.~Yang, M.~Price,
  T.~Robinson, A.~Remouchamps, and N.~Kill, ``Multifidelity multidisciplinary
  whole-engine thermomechanical design optimization,'' \emph{J. Propuls.
  Power}, vol.~30, no.~6, pp. 1654--1666, 2014.

\bibitem{JolyVP14}
M.~M. Joly, T.~Verstraete, and G.~Paniagua, ``Integrated multifidelity,
  multidisciplinary evolutionary design optimization of counterrotating
  compressors,'' \emph{Integr. Comput. Aided Eng.}, vol.~21, no.~3, pp.
  249--261, 2014.

\bibitem{MondalJS19}
S.~Mondal, M.~M. Joly, and S.~Sarkar, ``Multi-fidelity global-local
  optimization of a transonic compressor rotor,'' in \emph{Turbo Expo: Power
  for Land, Sea, and Air}, vol. 58585.\hskip 1em plus 0.5em minus 0.4em\relax
  American Society of Mechanical Engineers, 2019, p. V02DT46A020.

\bibitem{TezzeleFSS23}
M.~Tezzele, L.~Fabris, M.~Sidari, M.~Sicchiero, and G.~Rozza, ``A multifidelity
  approach coupling parameter space reduction and nonintrusive {POD} with
  application to structural optimization of passenger ship hulls,''
  \emph{International Journal for Numerical Methods in Engineering}, vol. 124,
  no.~5, pp. 1193--1210, 2023.

\bibitem{LiuWL23}
X.~Liu, D.~Wan, and L.~Lei, ``Multi-fidelity model and reduced-order method for
  comprehensive hydrodynamic performance optimization and prediction of {JBC}
  ship,'' \emph{Ocean Eng.}, vol. 267, p. 113321, 2023.

\bibitem{PellegriniSHD17}
R.~Pellegrini, A.~Serani, S.~Harries, and M.~Diez, ``Multi-objective hull-form
  optimization of a swath configuration via design-space dimensionality
  reduction, multi-fidelity metamodels, and swarm intelligence,'' in
  \emph{MARINE'17: Proc. of the the 7th International Conference on
  Computational Methods in Marine Engineering}.\hskip 1em plus 0.5em minus
  0.4em\relax CIMNE, 2017, pp. 95--106.

\bibitem{WankhedeBK2013}
M.~J. Wankhede, N.~W. Bressloff, and A.~J. Keane, ``Efficient strategy for low
  {NOx} combustor design in the spatial domain using multi-fidelity
  solutions,'' in \emph{Turbo Expo: Power for Land, Sea, and Air}, vol.
  55102.\hskip 1em plus 0.5em minus 0.4em\relax American Society of Mechanical
  Engineers, 2013, p. V01AT04A019.

\bibitem{WankhedeBK11}
------, ``Combustor design optimization using co-{Kriging} of steady and
  unsteady turbulent combustion,'' \emph{J. Eng. Gas Turbine. Power.}, vol.
  133, p. 121504, 2011.

\bibitem{MadsenL01}
J.~I. Madsen and M.~Langthjem, ``Multifidelity response surface approximations
  for the optimum design of diffuser flows,'' \emph{Optim. Eng.}, vol.~2, pp.
  453--468, 2001.

\bibitem{BuSHZ22}
Y.-P. Bu, W.-P. Song, Z.-H. Han, and Y.~Zhang, ``Efficient aerostructural
  optimization of helicopter rotors toward aeroacoustic noise reduction using
  multilevel hierarchical kriging model,'' \emph{Aerosp. Sci. Technol.}, vol.
  127, p. 107683, 2022.

\bibitem{ZhangZWZ21}
D.~Zhang, B.~Zhang, Z.~Wang, and X.~Zhu, ``An efficient surrogate-based
  optimization method for {BWBUG} based on multifidelity model and geometric
  constraint gradients,'' \emph{Math. Probl. Eng.}, vol. 2021, pp. 1--13, 2021.

\bibitem{SinghCEDD17}
P.~Singh, I.~Couckuyt, K.~Elsayed, D.~Deschrijver, and T.~Dhaene,
  ``Multi-objective geometry optimization of a gas cyclone using
  triple-fidelity co-{Kriging} surrogate models,'' \emph{J. Optim. Theory
  Appl.}, vol. 175, no.~1, pp. 172--193, 2017.

\bibitem{WangYZAZL22a}
L.~Wang, Y.~Yao, L.~Zhang, C.~D. Adenutsi, G.~Zhao, and F.~Lai, ``An
  intelligent multi-fidelity surrogate-assisted multi-objective reservoir
  production optimization method based on transfer stacking,'' \emph{Comput.
  Geosci.}, vol.~26, pp. 1279 -- 1295, 2022.

\bibitem{WangYZAZL22b}
L.~Wang, Y.~Yao, T.~Zhang, C.~D. Adenutsi, G.~Zhao, and F.~Lai, ``A novel
  self-adaptive multi-fidelity surrogate-assisted multi-objective evolutionary
  algorithm for simulation-based production optimization,'' \emph{J. Pet. Sci.
  Eng.}, vol. 211, p. 110111, 2022.

\bibitem{WangYTAZL22c}
L.~Wang, Y.~Yao, Z.~Tao, C.~D. Adenutsi, G.~C. Zhao, and F.~Lai, ``A rapid
  intelligent multi-fidelity surrogate-assisted multi-objective optimization
  method for water-flooding reservoir production optimization,'' \emph{Arab. J.
  Geosci.}, vol.~15, 2022.

\bibitem{CardosoD10}
M.~A. Cardoso and L.~J. Durlofsky, ``Use of reduced-order modeling procedures
  for production optimization,'' \emph{SPE Journal}, vol.~15, no.~02, pp.
  426--435, 2010.

\bibitem{WangYWDZL22}
L.~Wang, Y.~Yao, W.~Wang, C.~D. Adenutsi, G.~Zhao, and F.~Lai, ``Integrated
  optimization design for horizontal well spacing and fracture stage placement
  in shale gas reservoir,'' \emph{J. Nat. Gas Eng.}, vol. 105, p. 104706, 2022.

\bibitem{zhaoYWAFW22}
G.~Zhao, Y.~Yao, L.~Wang, C.~D. Adenutsi, D.~Feng, and W.~Z. Wu, ``Optimization
  design of horizontal well fracture stage placement in shale gas reservoirs
  based on an efficient variable-fidelity surrogate model and intelligent
  algorithm,'' \emph{Energy Rep.}, vol.~8, pp. 3589--3599, 2022.

\bibitem{BuYSL22}
H.~Bu, Y.~Yang, L.~Song, and J.~Li, ``Improving the film cooling performance of
  a turbine endwall with multi-fidelity modeling considering conjugate heat
  transfer,'' \emph{J. Turbomach.}, vol. 144, no.~1, p. 011011, 2022.

\bibitem{KimLY18}
Y.~Kim, S.~Lee, and K.~Yee, ``Variable-fidelity optimization of film-cooling
  hole arrangements considering conjugate heat transfer,'' \emph{J. Propuls.
  Power}, vol.~34, no.~5, pp. 1140--1151, 2018.

\bibitem{HartlGMB16}
D.~J. Hartl, E.~Galv{\'a}n, R.~J. Malak, and J.~W. Baur, ``Parameterized design
  optimization of a magnetohydrodynamic liquid metal active cooling concept,''
  \emph{J. Mech. Des.}, vol. 138, p. 031402, 2016.

\bibitem{HartlFB17}
D.~J. Hartl, G.~J. Frank, and J.~W. Baur, ``Embedded magnetohydrodynamic liquid
  metal thermal transport: validated analysis and design optimization,''
  \emph{J. Intell. Mater. Syst. Struct.}, vol.~28, no.~7, pp. 862--877, 2017.

\bibitem{WangLSSZ18}
X.~Wang, Y.~Liu, W.~Sun, X.~Song, and J.~Zhang, ``Multidisciplinary and
  multifidelity design optimization of electric vehicle battery thermal
  management system,'' \emph{J. Mech. Des.}, vol. 140, no.~9, p. 094501, 2018.

\bibitem{TranCMVLBH19}
D.~Tran, S.~Chakraborty, A.~V. Melckebeke, N.~Vu, Y.~Lan, M.~E. Baghdadi, and
  O.~Hegazy, ``Multi-fidelity electro-thermal optimization of multiport
  converter employing sic {MOSFET} and indirect liquid cooling,'' in
  \emph{EVER'19: Proc. of the 14th International Conference on Ecological
  Vehicles and Renewable Energies}, 2019, pp. 1--7.

\bibitem{JacobsKO13}
J.~P. Jacobs, S.~Koziel, and S.~Ogurtsov, ``Computationally efficient
  multi-fidelity bayesian support vector regression modeling of planar antenna
  input characteristics,'' \emph{IEEE Trans. Antennas Propag.}, vol.~61, pp.
  980--984, 2013.

\bibitem{LiuKA17}
B.~Liu, S.~Koziel, and N.~T. Ali, ``{SADEA-II:} {A} generalized method for
  efficient global optimization of antenna design,'' \emph{J. Comput. Des.
  Eng.}, vol.~4, no.~2, pp. 86--97, 2017.

\bibitem{KozielOL11}
S.~Koziel, S.~Ogurtsov, and L.~Leifsson, ``Simulation-driven design of antennas
  using coarse-discretization electromagnetic models,'' in \emph{ICCS'11: Proc.
  of the 2011 International Conference on Computational Science}, ser. Procedia
  Computer Science, vol.~4.\hskip 1em plus 0.5em minus 0.4em\relax Elsevier,
  2011, pp. 1252--1261.

\bibitem{KozielB16}
S.~Koziel and A.~Bekasiewicz, ``Rapid adjoint-based design optimization of
  compact microwave structures using multi-fidelity simulation models,''
  \emph{EuMC'16: Proc. of the 46th European Microwave Conference}, pp.
  174--177, 2016.

\bibitem{KozielO14}
S.~Koziel and S.~Ogurtsov, ``Multilevel microwave design optimization with
  automated model fidelity adjustment,'' \emph{Int. J. RF Microw. Comput.-Aided
  Eng.}, vol.~24, pp. 281--288, 2014.

\bibitem{KimKN0}
H.~S. Kim, M.~Koç, and J.~Ni, ``A hybrid multi-fidelity approach to the
  optimal design of warm forming processes using a knowledge-based artificial
  neural network,'' \emph{Int. J. Mach. Tools Manuf.}, vol.~47, pp. 211--222,
  2007.

\bibitem{LiLHCZHZ23}
M.~Li, Z.~Liu, L.~Huang, Q.~Chen, Q.~Zhai, W.~Han, and P.~Zhu, ``Multi-fidelity
  data-driven optimization design framework for self-piercing riveting process
  parameters,'' \emph{J. Manuf. Process.}, vol.~99, pp. 812--824, 2023.

\bibitem{ZhouYJSCHGW17}
Q.~Zhou, Y.~Yang, P.~Jiang, X.~Shao, L.~Cao, J.~Hu, Z.~Gao, and C.~Wang, ``A
  multi-fidelity information fusion metamodeling assisted laser beam welding
  process parameter optimization approach,'' \emph{Adv. Eng. Softw.}, vol. 110,
  pp. 85--97, 2017.

\bibitem{YangGC18}
Y.~Yang, Z.~Gao, and L.~Cao, ``Identifying optimal process parameters in deep
  penetration laser welding by adopting hierarchical-{Kriging} model,''
  \emph{Infrared Phys. Technol.}, vol.~92, pp. 443--453, 2018.

\bibitem{ZhouHW19}
X.~Zhou, S.-J. Hsieh, and J.-C. Wang, ``Accelerating extrusion-based additive
  manufacturing optimization processes with surrogate-based multi-fidelity
  models,'' \emph{Int. J. Adv. Manuf. Technol.}, vol. 103, pp. 4071--4083,
  2019.

\bibitem{SorourifarCP23}
F.~Sorourifar, N.~A. Choksi, and J.~A. Paulson, ``Computationally efficient
  integrated design and predictive control of flexible energy systems using
  multi-fidelity simulation-based bayesian optimization,'' \emph{Optim. Control
  Appl. Methods.}, vol.~44, no.~2, pp. 549--576, 2023.

\bibitem{ReistZRPB19}
T.~A. Reist, D.~W. Zingg, M.~Rakowitz, G.~Potter, and S.~Banerjee,
  ``Multifidelity optimization of hybrid wing--body aircraft with stability and
  control requirements,'' \emph{J. Aircr.}, vol.~56, no.~2, pp. 442--456, 2019.

\bibitem{AnselmaBRBE19}
P.~G. Anselma, A.~Biswas, J.~Roeleveld, G.~Belingardi, and A.~Emadi,
  ``Multi-fidelity near-optimal on-line control of a parallel hybrid electric
  vehicle powertrain,'' in \emph{ITEC'2019: IEEE transportation electrification
  conference and expo}.\hskip 1em plus 0.5em minus 0.4em\relax IEEE, 2019, pp.
  1--6.

\bibitem{KeskinPKW20}
M.~F. Keskin, B.~Peng, B.~Kulcs{\'a}r, and H.~Wymeersch, ``Altruistic control
  of connected automated vehicles in mixed-autonomy multi-lane highway
  traffic,'' \emph{IFAC-PapersOnLine}, vol.~53, pp. 14\,966--14\,971, 2020.

\bibitem{AlexandrovLGGN00}
N.~Alexandrov, R.~Lewis, C.~Gumbert, L.~Green, and P.~Newman, ``Optimization
  with variable-fidelity models applied to wing design,'' in \emph{Proc. of the
  38th Aerospace Sciences Meeting and Exhibit}, 2000, p. 841.

\bibitem{NagawkarRDLK21}
J.~Nagawkar, J.~Ren, X.~Du, L.~Leifsson, and S.~Koziel, ``Single-and multipoint
  aerodynamic shape optimization using multifidelity models and manifold
  mapping,'' \emph{J. of Aircr.}, vol.~58, no.~3, pp. 591--608, 2021.

\bibitem{GoldfeldVAK05}
Y.~Goldfeld, K.~Vervenne, J.~Arbocz, and F.~V. Keulen, ``Multi-fidelity
  optimization of laminated conical shells for buckling,'' \emph{Struct.
  Multidiscip. Optim.}, vol.~30, pp. 128--141, 2005.

\bibitem{ECCOMAS}
R.~Pellegrini, C.~Leotardi, U.~Iemma, E.~F. Campana, and M.~Diez, ``A
  multi-fidelity adaptive sampling method for metamodel-based uncertainty
  quantification of computer simulations,'' in \emph{ECCOMAS'16: Proc. of the
  VII European congress on computational methods in applied sciences and
  engineering}, 2016.

\bibitem{YajiYF22}
K.~Yaji, S.~Yamasaki, and K.~Fujita, ``Data-driven multifidelity topology
  design using a deep generative model: Application to forced convection heat
  transfer problems,'' \emph{Comput. Methods Appl. Mech. Eng.}, vol. 388, p.
  114284, 2022.

\bibitem{MannarinoM14}
A.~Mannarino and P.~Mantegazza, ``Multifidelity control of aeroelastic systems:
  an immersion and invariance approach,'' \emph{J. Guid. Control Dyn.},
  vol.~37, no.~5, pp. 1568--1582, 2014.

\bibitem{MoPR23}
C.~Mo, P.~Perdikaris, and J.~R. Raney, ``Accelerated design of architected
  materials with multifidelity {Bayesian} optimization,'' \emph{J. Eng. Mech.},
  vol. 149, 2023.

\bibitem{PilaniaGL17}
G.~Pilania, J.~E. Gubernatis, and T.~Lookman, ``Multi-fidelity machine learning
  models for accurate bandgap predictions of solids,'' \emph{Comput. Mater.
  Sci.}, vol. 129, pp. 156--163, 2017.

\bibitem{GhoreishiMSAA18}
S.~F. Ghoreishi, A.~Molkeri, A.~Srivastava, R.~Arr{\'o}yave, and D.~L. Allaire,
  ``Multi-information source fusion and optimization to realize {ICME}:
  Application to dual-phase materials,'' \emph{J. Mech. Des.}, vol. 140, p.
  111409, 2018.

\bibitem{KhatamsazMCJAAS21}
D.~Khatamsaz, A.~Molkeri, R.~Couperthwaite, J.~James, R.~Arr{\'o}yave, D.~L.
  Allaire, and A.~Srivastava, ``Efficiently exploiting
  process-structure-property relationships in material design by
  multi-information source fusion,'' \emph{Acta Mater.}, vol. 206, p. 116619,
  2021.

\bibitem{VianaSBL09}
F.~A.~C. Viana, V.~Steffen, S.~Butkewitsch, and M.~de~Freitas~Leal,
  ``Optimization of aircraft structural components by using nature-inspired
  algorithms and multi-fidelity approximations,'' \emph{J. Glob. Optim.},
  vol.~45, no.~3, pp. 427--449, 2009.

\bibitem{AnYK23}
H.~An, B.~D. Youn, and H.~S. Kim, ``Variable-stiffness composite optimization
  using dynamic and exponential multi-fidelity surrogate models,'' \emph{Int.
  J. Mech. Sci.}, vol. 257, p. 108547, 2023.

\bibitem{PourhabibHWZWD15}
A.~Pourhabib, J.~Z. Huang, K.~Wang, C.~Zhang, B.~Wang, and Y.~Ding, ``Modulus
  prediction of buckypaper based on multi-fidelity analysis involving latent
  variables,'' \emph{IIE Transactions}, vol.~47, pp. 141--152, 2015.

\bibitem{QianCZZZ21}
J.~Qian, Y.~Cheng, A.~Zhang, Q.~Zhou, and J.~Zhang, ``Optimization design of
  metamaterial vibration isolator with honeycomb structure based on
  multi-fidelity surrogate model,'' \emph{Struct. Multidisc. Optim.}, vol.~64,
  pp. 423--439, 2021.

\bibitem{GuoHWHX20}
Q.~Guo, J.~Hang, S.~Wang, W.~Hui, and Z.~H. Xie, ``Design optimization of
  variable stiffness composites by using multi-fidelity surrogate models,''
  \emph{Struct. Multidisc. Optim.}, vol.~63, pp. 439--461, 2020.

\bibitem{KhatamsazMCJASA21}
D.~Khatamsaz, A.~Molkeri, R.~Couperthwaite, J.~James, R.~Arr{\'o}yave,
  A.~Srivastava, and D.~L. Allaire, ``Adaptive active subspace-based efficient
  multifidelity materials design,'' \emph{Mater. Des.}, vol. 209, p. 110001,
  2021.

\bibitem{FareFBVP22}
C.~Fare, P.~Fenner, M.~Benatan, A.~Varsi, and E.~O. Pyzer-Knapp, ``A
  multi-fidelity machine learning approach to high throughput materials
  screening,'' \emph{Npj Comput. Mater.}, vol.~8, pp. 1--9, 2022.

\bibitem{Lewis10}
S.~P. Lewis, ``A how-to guide for predicting properties of materials with {DFT}
  [review of "density functional theory: {A} practical introduction" (sholl,
  {D.S.} and steckel, j.a.; 2009) [book review],'' \emph{Comput. Sci. Eng.},
  vol.~12, no.~6, pp. 5--7, 2010.

\bibitem{Greenman21}
K.~Greenman, W.~H. Green, and R.~G{\'o}mez-Bombarelli, ``Multi-fidelity
  prediction of molecular optical peaks with deep learning,'' \emph{Chem.
  Sci.}, vol.~13, pp. 1152--1162, 2021.

\bibitem{EgorovaHWD20}
O.~A. Egorova, R.~Hafizi, D.~C. Woods, and G.~M. Day, ``Multifidelity
  statistical machine learning for molecular crystal structure prediction,''
  \emph{J. Phys. Chem. A}, vol. 124, no.~39, pp. 8065--8078, 2020.

\bibitem{MayrWWL22}
\BIBentryALTinterwordspacing
F.~Mayr, M.~Wieder, O.~Wieder, and T.~Langer, ``Improving small molecule pk a
  prediction using transfer learning with graph neural networks,'' \emph{Front.
  Chem.}, vol.~10, 2022. [Online]. Available:
  \url{https://api.semanticscholar.org/CorpusID:246178950}
\BIBentrySTDinterwordspacing

\bibitem{WuWWZCHH22}
J.~Wu, Y.~Wan, Z.~Wu, S.~Zhang, D.~Cao, C.-Y. Hsieh, and T.~Hou, ``Mf-sup-pka:
  Multi-fidelity modeling with subgraph pooling mechanism for pka prediction,''
  \emph{Acta Pharm. Sin. B.}, vol.~13, pp. 2572--2584, 2022.

\bibitem{DuanFZLHLLDHHC23}
Y.~Duan, L.~Fu, X.~Zhang, T.~Long, Y.~He, Z.~Liu, A.~Lu, Y.~Deng, C.~Hsieh,
  T.~Hou, and D.~Cao, ``Improved gnns for log d 7.4 prediction by transferring
  knowledge from low-fidelity data,'' \emph{J. Chem. Inf. Model.}, vol.~63,
  no.~8, pp. 2345--2359, 2023.

\bibitem{WangXXZCRNTQZWCLZ23}
Y.~Wang, J.~Xiong, F.~Xiao, W.~Zhang, K.~Cheng, J.~Rao, B.~Niu, X.~Tong, N.~Qu,
  R.~Zhang, D.~Wang, K.~Chen, X.~Li, and M.~Zheng, ``Logd7.4 prediction
  enhanced by transferring knowledge from chromatographic retention time,
  microscopic pka and logp,'' \emph{J. Cheminformatics}, vol.~15, no.~1, p.~76,
  2023.

\bibitem{ButerezJKL23}
D.~Buterez, J.~P. Janet, S.~J. Kiddle, and P.~Li{\`{o}}, ``{MF-PCBA:}
  multifidelity high-throughput screening benchmarks for drug discovery and
  machine learning,'' \emph{J. Chem. Inf. Model.}, vol.~63, no.~9, pp.
  2667--2678, 2023.

\bibitem{HernSJLB23}
A.~Hern{\'{a}}ndez{-}Garc{\'{\i}}a, N.~Saxena, M.~Jain, C.~Liu, and Y.~Bengio,
  ``Multi-fidelity active learning with {GFlowNets},'' \emph{CoRR}, vol.
  abs/2306.11715, 2023.

\bibitem{MadinS23}
\BIBentryALTinterwordspacing
O.~C. Madin and M.~R. Shirts, ``Using physical property surrogate models to
  perform accelerated multi-fidelity optimization of force field parameters,''
  \emph{Digital Discovery}, vol.~2, pp. 828 -- 847, 2023. [Online]. Available:
  \url{https://api.semanticscholar.org/CorpusID:258539519}
\BIBentrySTDinterwordspacing

\bibitem{JagtapMK22}
\BIBentryALTinterwordspacing
A.~D. Jagtap, D.~Mitsotakis, and G.~E. Karniadakis, ``Deep learning of inverse
  water waves problems using multi-fidelity data: Application to
  {Serre-Green-Naghdi} equations,'' \emph{CoRR}, vol. abs/2202.02899, 2022.
  [Online]. Available: \url{https://arxiv.org/abs/2202.02899}
\BIBentrySTDinterwordspacing

\bibitem{BasirS21}
S.~Basir and I.~Senocak, ``Physics and equality constrained artificial neural
  networks: Application to forward and inverse problems with multi-fidelity
  data fusion,'' \emph{J. Comput. Phys.}, vol. 463, p. 111301, 2021.

\bibitem{RegazzoniPCLQ21}
F.~Regazzoni, S.~Pagani, A.~Cosenza, A.~Lombardi, and A.~M. Quarteroni, ``A
  physics-informed multi-fidelity approach for the estimation of differential
  equations parameters in low-data or large-noise regimes,'' \emph{Rendiconti
  Lincei}, vol.~32, no.~3, pp. 437--470, 2021.

\bibitem{RamezankhaniNNV22}
M.~Ramezankhani, A.~Nazemi, A.~Narayan, H.~Voggenreiter, M.~Harandi, R.~J.
  Seethaler, and A.~S. Milani, ``A data-driven multi-fidelity physics-informed
  learning framework for smart manufacturing: {A} composites processing case
  study,'' in \emph{{ICPS}'22: Proc. of the 5th {IEEE} International Conference
  on Industrial Cyber-Physical System}, 2022, pp. 1--7.

\bibitem{IslamTMN21}
M.~Islam, M.~S.~H. Thakur, S.~Mojumder, and M.~N. Hasan, ``Extraction of
  material properties through multi-fidelity deep learning from molecular
  dynamics simulation,'' \emph{Comput. Mater. Sci.}, vol. 188, p. 110187, 2021.

\bibitem{LeiLW22}
J.~Lei, Q.~Liu, and X.~Wang, ``Physics-informed multi-fidelity learning-driven
  imaging method for electrical capacitance tomography,'' \emph{Eng. Appl.
  Artif. Intell.}, vol. 116, p. 105467, 2022.

\bibitem{LeiW23}
J.~Lei and X.~Wang, ``Transfer learning-driven inversion method for the imaging
  problem in electrical capacitance tomography,'' \emph{Expert Syst. Appl.},
  vol. 227, p. 120277, 2023.

\bibitem{ThelenBSB22}
A.~S. Thelen, D.~E. Bryson, B.~K. Stanford, and P.~S. Beran, ``Multi-fidelity
  gradient-based optimization for high-dimensional aeroelastic
  configurations,'' \emph{Algorithms}, vol.~15, no.~4, p. 131, 2022.

\bibitem{LiuCOW19}
H.~Liu, J.~Cai, Y.~Ong, and Y.~Wang, ``Understanding and comparing scalable
  gaussian process regression for big data,'' \emph{Knowl. Based Syst.}, vol.
  164, pp. 324--335, 2019.

\bibitem{PadidarZHGB21}
M.~Padidar, X.~Zhu, L.~Huang, J.~R. Gardner, and D.~Bindel, ``Scaling gaussian
  processes with derivative information using variational inference,'' in
  \emph{NeurIPS'21: Proc. of 34th Annual Conference on Neural Information
  Processing Systems 2021}, 2021, pp. 6442--6453.

\bibitem{LiFK11}
K.~Li, {\'{A}}.~Fialho, and S.~Kwong, ``Multi-objective differential evolution
  with adaptive control of parameters and operators,'' in \emph{LION5: Proc. of
  the 5th International Conference on Learning and Intelligent Optimization},
  2011, pp. 473--487.

\bibitem{LiKWTM13}
K.~Li, S.~Kwong, R.~Wang, K.~Tang, and K.~Man, ``Learning paradigm based on
  jumping genes: {A} general framework for enhancing exploration in
  evolutionary multiobjective optimization,'' \emph{Inf. Sci.}, vol. 226, pp.
  1--22, 2013.

\bibitem{SunL20}
L.~Sun and K.~Li, ``Adaptive operator selection based on dynamic thompson
  sampling for {MOEA/D},'' in \emph{PPSN'20: Proc. of 16th International
  Conference on Parallel Problem Solving from Nature}, ser. Lecture Notes in
  Computer Science, vol. 12270.\hskip 1em plus 0.5em minus 0.4em\relax
  Springer, 2020, pp. 271--284.

\bibitem{LiZLZL09}
K.~Li, J.~Zheng, M.~Li, C.~Zhou, and H.~Lv, ``A novel algorithm for
  non-dominated hypervolume-based multiobjective optimization,'' in
  \emph{SMC'09: Proc. of 2009 the {IEEE} International Conference on Systems,
  Man and Cybernetics}, 2009, pp. 5220--5226.

\bibitem{LiKWCR12}
K.~Li, S.~Kwong, R.~Wang, J.~Cao, and I.~J. Rudas, ``Multi-objective
  differential evolution with self-navigation,'' in \emph{SMC'12: Proc. of the
  2012 {IEEE} International Conference on Systems, Man, and Cybernetics}, 2012,
  pp. 508--513.

\bibitem{LiZKLW14}
K.~Li, Q.~Zhang, S.~Kwong, M.~Li, and R.~Wang, ``Stable matching-based
  selection in evolutionary multiobjective optimization,'' \emph{{IEEE} Trans.
  Evol. Comput.}, vol.~18, no.~6, pp. 909--923, 2014.

\bibitem{LiFKZ14}
K.~Li, {\'{A}}.~Fialho, S.~Kwong, and Q.~Zhang, ``Adaptive operator selection
  with bandits for a multiobjective evolutionary algorithm based on
  decomposition,'' \emph{{IEEE} Trans. Evol. Comput.}, vol.~18, no.~1, pp.
  114--130, 2014.

\bibitem{LiKD15}
K.~Li, S.~Kwong, and K.~Deb, ``A dual-population paradigm for evolutionary
  multiobjective optimization,'' \emph{Inf. Sci.}, vol. 309, pp. 50--72, 2015.

\bibitem{LiDZK15}
K.~Li, K.~Deb, Q.~Zhang, and S.~Kwong, ``An evolutionary many-objective
  optimization algorithm based on dominance and decomposition,'' \emph{{IEEE}
  Trans. Evol. Comput.}, vol.~19, no.~5, pp. 694--716, 2015.

\bibitem{LiKZD15}
K.~Li, S.~Kwong, Q.~Zhang, and K.~Deb, ``Interrelationship-based selection for
  decomposition multiobjective optimization,'' \emph{{IEEE} Trans. Cybern.},
  vol.~45, no.~10, pp. 2076--2088, 2015.

\bibitem{WuKZLWL15}
M.~Wu, S.~Kwong, Q.~Zhang, K.~Li, R.~Wang, and B.~Liu, ``Two-level stable
  matching-based selection in {MOEA/D},'' in \emph{SMC'15: Proc. of the 2015
  {IEEE} International Conference on Systems, Man, and Cybernetics}, 2015, pp.
  1720--1725.

\bibitem{WuKJLZ17}
M.~Wu, S.~Kwong, Y.~Jia, K.~Li, and Q.~Zhang, ``Adaptive weights generation for
  decomposition-based multi-objective optimization using gaussian process
  regression,'' in \emph{GECCO'17: Proc. of the 2017 Genetic and Evolutionary
  Computation Conference}.\hskip 1em plus 0.5em minus 0.4em\relax {ACM}, 2017,
  pp. 641--648.

\bibitem{WuLKZZ17}
M.~Wu, K.~Li, S.~Kwong, Y.~Zhou, and Q.~Zhang, ``Matching-based selection with
  incomplete lists for decomposition multiobjective optimization,''
  \emph{{IEEE} Trans. Evol. Comput.}, vol.~21, no.~4, pp. 554--568, 2017.

\bibitem{LiCMY18}
K.~Li, R.~Chen, G.~Min, and X.~Yao, ``Integration of preferences in
  decomposition multiobjective optimization,'' \emph{{IEEE} Trans. Cybern.},
  vol.~48, no.~12, pp. 3359--3370, 2018.

\bibitem{WuLKZZ19}
\BIBentryALTinterwordspacing
M.~Wu, K.~Li, S.~Kwong, Q.~Zhang, and J.~Zhang, ``Learning to decompose: {A}
  paradigm for decomposition-based multiobjective optimization,'' \emph{{IEEE}
  Trans. Evol. Comput.}, vol.~23, no.~3, pp. 376--390, 2019. [Online].
  Available: \url{https://doi.org/10.1109/TEVC.2018.2865931}
\BIBentrySTDinterwordspacing

\bibitem{WuLKZ20}
M.~Wu, K.~Li, S.~Kwong, and Q.~Zhang, ``Evolutionary many-objective
  optimization based on adversarial decomposition,'' \emph{{IEEE} Trans.
  Cybern.}, vol.~50, no.~2, pp. 753--764, 2020.

\bibitem{LyuYWHL23}
B.~Lyu, Y.~Yang, S.~Wen, T.~Huang, and K.~Li, ``Neural architecture search for
  portrait parsing,'' \emph{{IEEE} Trans. Neural Networks Learn. Syst.},
  vol.~34, no.~3, pp. 1112--1121, 2023.

\bibitem{LyuHYCYLWH23}
B.~Lyu, M.~Hamdi, Y.~Yang, Y.~Cao, Z.~Yan, K.~Li, S.~Wen, and T.~Huang,
  ``Efficient spectral graph convolutional network deployment on memristive
  crossbars,'' \emph{{IEEE} Trans. Emerg. Top. Comput. Intell.}, vol.~7, no.~2,
  pp. 415--425, 2023.

\bibitem{LyuLHWYL23}
B.~Lyu, L.~Lu, M.~Hamdi, S.~Wen, Y.~Yang, and K.~Li, ``{MTLP-JR:} multi-task
  learning-based prediction for joint ranking in neural architecture search,''
  \emph{Comput. Electr. Eng.}, vol. 105, p. 108474, 2023.

\bibitem{TangLDWZF20}
R.~Tang, K.~Li, W.~Ding, Y.~Wang, H.~Zhou, and G.~Fu, ``Reference point based
  multi-objective optimization of reservoir operation: a comparison of three
  algorithms,'' \emph{Water Resources Management}, vol.~34, no.~3, pp.
  1005--1020, 2020.

\bibitem{ZhouLM22a}
S.~Zhou, K.~Li, and G.~Min, ``Attention-based genetic algorithm for adversarial
  attack in natural language processing,'' in \emph{PPSN'22: Proc. of 17th
  International Conference on Parallel Problem Solving from Nature}, ser.
  Lecture Notes in Computer Science, vol. 13398.\hskip 1em plus 0.5em minus
  0.4em\relax Springer, 2022, pp. 341--355.

\bibitem{ZhouLM22b}
------, ``Adversarial example generation via genetic algorithm: a preliminary
  result,'' in \emph{{GECCO}'22: Companion of 2022 Genetic and Evolutionary
  Computation Conference}.\hskip 1em plus 0.5em minus 0.4em\relax {ACM}, 2022,
  pp. 469--470.

\bibitem{WilliamsLM23a}
P.~N. Williams, K.~Li, and G.~Min, ``A surrogate assisted evolutionary strategy
  for image approximation by density-ratio estimation,'' in \emph{CEC'23: Proc.
  of 2023 {IEEE} Congress on Evolutionary Computation}.\hskip 1em plus 0.5em
  minus 0.4em\relax {IEEE}, 2023, pp. 1--8.

\bibitem{WilliamsLM23b}
\BIBentryALTinterwordspacing
------, ``Sparse adversarial attack via bi-objective optimization,'' in
  \emph{EMO'23: Proc. of the 12th International Conference on Evolutionary
  Multi-Criterion Optimization}, ser. Lecture Notes in Computer Science, vol.
  13970.\hskip 1em plus 0.5em minus 0.4em\relax Springer, 2023, pp. 118--133.
  [Online]. Available: \url{https://doi.org/10.1007/978-3-031-27250-9\_9}
\BIBentrySTDinterwordspacing

\bibitem{WilliamsLM22}
------, ``Black-box adversarial attack via overlapped shapes,'' in
  \emph{{GECCO} '22: Genetic and Evolutionary Computation Conference, Companion
  Volume, Boston, Massachusetts, USA, July 9 - 13, 2022}.\hskip 1em plus 0.5em
  minus 0.4em\relax {ACM}, 2022, pp. 467--468.

\bibitem{WilliamsL23c}
P.~N. Williams and K.~Li, ``Black-box sparse adversarial attack via
  multi-objective optimisation {CVPR} proceedings,'' in \emph{CVPR'23: Proc. of
  2023 {IEEE/CVF} Conference on Computer Vision and Pattern Recognition}.\hskip
  1em plus 0.5em minus 0.4em\relax {IEEE}, 2023, pp. 12\,291--12\,301.

\bibitem{LiXT19}
K.~Li, Z.~Xiang, and K.~C. Tan, ``Which surrogate works for empirical
  performance modelling? {A} case study with differential evolution,'' in
  \emph{CEC'19: Proc. of the 2019 {IEEE} Congress on Evolutionary Computation},
  2019, pp. 1988--1995.

\bibitem{LiXCWT20}
K.~Li, Z.~Xiang, T.~Chen, S.~Wang, and K.~C. Tan, ``Understanding the automated
  parameter optimization on transfer learning for cross-project defect
  prediction: an empirical study,'' in \emph{{ICSE}'20: Proc. of the 42nd
  International Conference on Software Engineering}.\hskip 1em plus 0.5em minus
  0.4em\relax {ACM}, 2020, pp. 566--577.

\bibitem{LiuLC20}
M.~Liu, K.~Li, and T.~Chen, ``{DeepSQLi}: deep semantic learning for testing
  {SQL} injection,'' in \emph{{ISSTA}'20: Proc. of the 29th {ACM} {SIGSOFT}
  International Symposium on Software Testing and Analysis}.\hskip 1em plus
  0.5em minus 0.4em\relax {ACM}, 2020, pp. 286--297.

\bibitem{LiXCT20}
K.~Li, Z.~Xiang, T.~Chen, and K.~C. Tan, ``{BiLO-CPDP}: Bi-level programming
  for automated model discovery in cross-project defect prediction,'' in
  \emph{ASE'20: Proc. of the 35th {IEEE/ACM} International Conference on
  Automated Software Engineering}.\hskip 1em plus 0.5em minus 0.4em\relax
  {IEEE}, 2020, pp. 573--584.

\bibitem{LiYV23}
K.~Li, H.~Yang, and W.~Visser, ``Danuoyi: Evolutionary multi-task injection
  testing on web application firewalls,'' \emph{IEEE Trans. Softw. Eng.}, 2023,
  accepted for publication.

\bibitem{XuLAZ21}
J.~Xu, K.~Li, M.~Abusara, and Y.~Zhang, ``Admm-based {OPF} problem against
  cyber attacks in smart grid,'' in \emph{SMC'21: Proc. of 2021 {IEEE}
  International Conference on Systems, Man, and Cybernetics}.\hskip 1em plus
  0.5em minus 0.4em\relax {IEEE}, 2021, pp. 1418--1423.

\bibitem{XuLA21}
J.~Xu, K.~Li, and M.~Abusara, ``Multi-objective reinforcement learning based
  multi-microgrid system optimisation problem,'' in \emph{EMO'21: Proc. of 11th
  International Conference on Evolutionary Multi-Criterion Optimization}, vol.
  12654.\hskip 1em plus 0.5em minus 0.4em\relax Springer, 2021, pp. 684--696.

\bibitem{XuLA22}
------, ``Preference based multi-objective reinforcement learning for
  multi-microgrid system optimization problem in smart grid,'' \emph{Memetic
  Comput.}, vol.~14, no.~2, pp. 225--235, 2022.

\bibitem{BillingsleyLMMG19}
J.~Billingsley, K.~Li, W.~Miao, G.~Min, and N.~Georgalas, ``A formal model for
  multi-objective optimisation of network function virtualisation placement,''
  in \emph{EMO'19: Proc. of 10th International Conference on Evolutionary
  Multi-Criterion Optimization}, ser. Lecture Notes in Computer Science, vol.
  11411.\hskip 1em plus 0.5em minus 0.4em\relax Springer, 2019, pp. 529--540.

\bibitem{BillingsleyLMMG20}
------, ``Routing-led placement of vnfs in arbitrary networks,'' in
  \emph{CEC'20: Proc. of 2020 {IEEE} Congress on Evolutionary
  Computation}.\hskip 1em plus 0.5em minus 0.4em\relax {IEEE}, 2020, pp. 1--8.

\bibitem{BillingsleyMLMG20}
J.~Billingsley, W.~Miao, K.~Li, G.~Min, and N.~Georgalas, ``Performance
  analysis of {SDN} and {NFV} enabled mobile cloud computing,'' in
  \emph{GLOBECOM'20: Proc. of 2020 {IEEE} Global Communications
  Conference}.\hskip 1em plus 0.5em minus 0.4em\relax {IEEE}, 2020, pp. 1--6.

\bibitem{BillingsleyLMMG21}
J.~Billingsley, K.~Li, W.~Miao, G.~Min, and N.~Georgalas, ``Parallel algorithms
  for the multiobjective virtual network function placement problem,'' in
  \emph{EMO'21: Proc. of 11th International Conference on Evolutionary
  Multi-Criterion Optimization}, ser. Lecture Notes in Computer Science, vol.
  12654.\hskip 1em plus 0.5em minus 0.4em\relax Springer, 2021, pp. 708--720.

\bibitem{CaoKWL12}
J.~Cao, S.~Kwong, R.~Wang, and K.~Li, ``A weighted voting method using minimum
  square error based on extreme learning machine,'' in \emph{ICMLC'12: Proc. of
  the 2012 International Conference on Machine Learning and Cybernetics}, 2012,
  pp. 411--414.

\bibitem{LiWKC13}
K.~Li, R.~Wang, S.~Kwong, and J.~Cao, ``Evolving extreme learning machine
  paradigm with adaptive operator selection and parameter control,''
  \emph{International Journal of Uncertainty, Fuzziness and Knowledge-Based
  Systems}, vol. supp02, pp. 143--154, 2013.

\bibitem{LiK14}
K.~Li and S.~Kwong, ``A general framework for evolutionary multiobjective
  optimization via manifold learning,'' \emph{Neurocomputing}, vol. 146, pp.
  65--74, 2014.

\bibitem{CaoKWLLK15}
J.~Cao, S.~Kwong, R.~Wang, X.~Li, K.~Li, and X.~Kong, ``Class-specific soft
  voting based multiple extreme learning machines ensemble,''
  \emph{Neurocomputing}, vol. 149, pp. 275--284, 2015.

\bibitem{WangYLK21}
R.~Wang, S.~Ye, K.~Li, and S.~Kwong, ``Bayesian network based label correlation
  analysis for multi-label classifier chain,'' \emph{Inf. Sci.}, vol. 554, pp.
  256--275, 2021.

\bibitem{GaoNL19}
H.~Gao, H.~Nie, and K.~Li, ``Visualisation of pareto front approximation: {A}
  short survey and empirical comparisons,'' in \emph{CEC'19: Proc. of the 2019
  {IEEE} Congress on Evolutionary Computation}, 2019, pp. 1750--1757.

\bibitem{LiDY18}
K.~Li, K.~Deb, and X.~Yao, ``R-metric: {Evaluating} the performance of
  preference-based evolutionary multiobjective optimization using reference
  points,'' \emph{{IEEE} Trans. Evol. Comput.}, vol.~22, no.~6, pp. 821--835,
  2018.

\bibitem{LiLDMY20}
K.~Li, M.~Liao, K.~Deb, G.~Min, and X.~Yao, ``Does preference always help? {A}
  holistic study on preference-based evolutionary multiobjective optimization
  using reference points,'' \emph{{IEEE} Trans. Evol. Comput.}, vol.~24, no.~6,
  pp. 1078--1096, 2020.

\bibitem{LiNGY22}
K.~Li, H.~Nie, H.~Gao, and X.~Yao, ``Posterior decision making based on
  decomposition-driven knee point identification,'' \emph{{IEEE} Trans. Evol.
  Comput.}, vol.~26, no.~6, pp. 1409--1423, 2022.

\bibitem{TanabeL23}
R.~Tanabe and K.~Li, ``Quality indicators for preference-based evolutionary
  multi-objective optimization using a reference point: A review and
  analysis,'' \emph{IEEE Trans. Evol. Comput.}, 2023, accepted for publication.

\bibitem{ChenLY18}
R.~Chen, K.~Li, and X.~Yao, ``Dynamic multiobjectives optimization with a
  changing number of objectives,'' \emph{{IEEE} Trans. Evol. Comput.}, vol.~22,
  no.~1, pp. 157--171, 2018.

\bibitem{Li19}
K.~Li, ``Progressive preference learning: Proof-of-principle results in
  {MOEA/D},'' in \emph{EMO'19: Proc. of the 10th International Conference
  Evolutionary Multi-Criterion Optimization}, 2019, pp. 631--643.

\bibitem{LiCSY19}
K.~Li, R.~Chen, D.~A. Savic, and X.~Yao, ``Interactive decomposition
  multiobjective optimization via progressively learned value functions,''
  \emph{{IEEE} Trans. Fuzzy Syst.}, vol.~27, no.~5, pp. 849--860, 2019.

\bibitem{FanLT20}
X.~Fan, K.~Li, and K.~C. Tan, ``Surrogate assisted evolutionary algorithm based
  on transfer learning for dynamic expensive multi-objective optimisation
  problems,'' in \emph{CEC'20: Proc. of the 2020 {IEEE} Congress on
  Evolutionary Computation}.\hskip 1em plus 0.5em minus 0.4em\relax {IEEE},
  2020, pp. 1--8.

\bibitem{ChenL21a}
R.~Chen and K.~Li, ``Transfer bayesian optimization for expensive black-box
  optimization in dynamic environment,'' in \emph{SMC'21: Proc. of 2021 {IEEE}
  International Conference on Systems, Man, and Cybernetics}.\hskip 1em plus
  0.5em minus 0.4em\relax {IEEE}, 2021, pp. 1374--1379.

\bibitem{LiLL23}
S.~Li, K.~Li, and W.~Li, ``"why not looking backward?" a robust two-step
  method to automatically terminate bayesian optimization,'' in
  \emph{NeurIPS'23: Proc. of 37th Conference on Neural Information Processing
  Systems}, 2023, pp. 1--12.

\bibitem{LiCFY19}
K.~Li, R.~Chen, G.~Fu, and X.~Yao, ``Two-archive evolutionary algorithm for
  constrained multiobjective optimization,'' \emph{{IEEE} Trans. Evol.
  Comput.}, vol.~23, no.~2, pp. 303--315, 2019.

\bibitem{ShanL21}
X.~Shan and K.~Li, ``An improved two-archive evolutionary algorithm for
  constrained multi-objective optimization,'' in \emph{EMO'21: Proc. of the
  11th International Conference on Evolutionary Multicriteria Optimization},
  ser. Lecture Notes in Computer Science, vol. 12654.\hskip 1em plus 0.5em
  minus 0.4em\relax Springer, 2021, pp. 235--247.

\bibitem{LiLL22}
S.~Li, K.~Li, and W.~Li, ``Do we really need to use constraint violation in
  constrained evolutionary multi-objective optimization?'' in \emph{PPSN'22:
  Proc. of the 17th International Conference on Parallel Problem Solving from
  Nature}, ser. Lecture Notes in Computer Science, vol. 13399.\hskip 1em plus
  0.5em minus 0.4em\relax Springer, 2022, pp. 124--137.

\bibitem{WangL24}
\BIBentryALTinterwordspacing
S.~Wang and K.~Li, ``Constrained {Bayesian} optimization under partial
  observations: Balanced improvements and provable convergence,'' in
  \emph{AAAI'24: Proc. of the Thirty-Eighth AAAI Conference on Artificial
  Intelligence}, 2024, accepted for publication. [Online]. Available:
  \url{https://arxiv.org/abs/2312.03212}
\BIBentrySTDinterwordspacing

\bibitem{IshibuchiSMN17}
H.~Ishibuchi, Y.~Setoguchi, H.~Masuda, and Y.~Nojima, ``Performance of
  decomposition-based many-objective algorithms strongly depends on pareto
  front shapes,'' \emph{{IEEE} Trans. Evol. Comput.}, vol.~21, no.~2, pp.
  169--190, 2017.

\bibitem{YingKCR0H19}
C.~Ying, A.~Klein, E.~Christiansen, E.~Real, K.~Murphy, and F.~Hutter,
  ``{NAS-Bench-101}: {Towards} reproducible neural architecture search,'' in
  \emph{ICML'19: Proc. of the 36th International Conference on Machine
  Learning}, vol.~97.\hskip 1em plus 0.5em minus 0.4em\relax {PMLR}, 2019, pp.
  7105--7114.

\bibitem{DongY20}
X.~Dong and Y.~Yang, ``{NAS-Bench-201}: {Extending} the scope of reproducible
  neural architecture search,'' in \emph{ICLR'20: Proc. of the 8th
  International Conference on Learning Representations}.\hskip 1em plus 0.5em
  minus 0.4em\relax OpenReview.net, 2020.

\bibitem{EggenspergerMMF21}
K.~Eggensperger, P.~M{\"{u}}ller, N.~Mallik, M.~Feurer, R.~Sass, A.~Klein,
  N.~H. Awad, M.~Lindauer, and F.~Hutter, ``{HPOBench}: {A} collection of
  reproducible multi-fidelity benchmark problems for {HPO},'' in
  \emph{NeurIPS'21: Proc. of the Annual Conference on Neural Information
  Processing Systems 2021}, 2021.

\bibitem{PfistererSMBB22}
F.~Pfisterer, L.~Schneider, J.~Moosbauer, M.~Binder, and B.~Bischl, ``{YAHPO}
  gym - an efficient multi-objective multi-fidelity benchmark for
  hyperparameter optimization,'' in \emph{AutoML'22: Proc. of 2022
  International Conference on Automated Machine Learning}, vol. 188, 2022, pp.
  3/1--39.

\end{thebibliography}


\end{document}